\documentclass[aps,footinbib, notitlepage,superscriptaddress, longbibliography,eqsecnum]{revtex4-2}

\usepackage[english]{babel}
\usepackage[utf8]{inputenc}
\usepackage{amsmath}
\usepackage{amsthm}
\usepackage{braket}
\usepackage{amssymb}
\usepackage{graphicx}
\usepackage{hyperref}
\usepackage[normalsize]{subfigure}
\usepackage{dsfont}
\usepackage{xcolor}
\usepackage{geometry}
\usepackage{mathtools}
\usepackage{bbold}
\usepackage{bbm}
\usepackage[capitalize]{cleveref}

\DeclareFontFamily{OT1}{pzc}{}
\DeclareFontShape{OT1}{pzc}{m}{it}{<-> s * [1.10] pzcmi7t}{}
\DeclareMathAlphabet{\mathpzc}{OT1}{pzc}{m}{it}

\definecolor{deepchampagne}{rgb}{0.98, 0.84, 0.65}

\providecommand{\st}[1]{_{\text{#1}}}
\providecommand{\sfrac}[2]{#1/#2}

\def\const{\mathrm{const}}

\def\Imat{\mathbbm{I}}
\def\im{\mathrm{i}}

\def\kv{\bv{k}}

\def\qv{\bv{q}}

\def\gv{\bv{g}}
\def\Gv{\bv{G}}

\def\Fv{\bv{F}}
\def\fv{\bv{f}}

\def\yv{\bv{y}}

\def\xv{\bv{x}}

\def\Rv{\bv{R}}

\def\hv{\bv{h}}

\def\b0{\bv{0}}

\def\Dcal{\mathcal{D}}

\def\Ocal{\mathcal{O}}
\def\Pcal{\mathcal{P}}
\def\Qcal{\mathcal{Q}}

\def\Scal{\mathcal{S}}

\def\psiv{\boldsymbol{\psi}}
\def\Psiv{\boldsymbol{\Psi}}

\def\Dt{{\Delta t}}

\def\reals{\mathbb{R}}

\def\E{\mathbb{E}}

\def\fch{\st{fch}}

\newcommand{\beq}{\begin{equation}}
\newcommand{\eeq}{\end{equation}}
\newcommand{\beqn}{\begin{equation*}}
\newcommand{\eeqn}{\end{equation*}}
\newcommand{\bv}[1]{\mathbf{#1}}

\begin{document}

\title{Flow map learning in nonlinear vector autoregressive models:\\ influence of the feature-library structure on the training error}

\author{Markus Gross}

\affiliation{Institute for AI Safety and Security, German Aerospace Center (DLR),\\ Sankt Augustin and Ulm, Germany}

\date{\today}

\begin{abstract}
Time series forecasting often requires learning nonlinear and time-delayed dependencies. A paradigmatic class of forecasting models are nonlinear vector autoregressive processes (NVAR), also known as next-generation reservoir computers (NG-RCs). These models approximate the Koopman operator on the space spanned by their explicit feature library. We consider the identifiability problem for learning Markovian nonlinear dynamical systems and show that the training error as a function of time resolution follows characteristic (pre-)asymptotic scaling laws. These laws depend on whether the feature library can represent the early Lie-series coefficients of the flow map (propagator) exactly or merely approximately. For dynamical systems governed by polynomial vector fields, we demonstrate the mechanism for NVAR/NG-RC models with monomial and Fourier feature libraries. We determine the dependence of the training error on the temporal resolution, the involved nonlinear degree, and the number of delay terms. While delay terms reduce the optimal one-step training error, they improve long-horizon forecasts only when the library provides sufficient nonlinearity. Thus, small training error coexists with weak generalization as the model class is mismatched to the true data-generating process. Numerical experiments on various chaotic dynamical systems confirm the theoretical predictions.
\end{abstract}

\maketitle

\section{Introduction}
\label{sec_intro}

Reservoir computing (RC) has seen broad success in the domain of time series forecasting \cite{lukosevicius_practical_2012,gilpin_model_2023,yan_emerging_2024}. 
Part of its appeal stems from the fact that the classical echo-state-network (ESN) architecture is realizable in various physical systems \cite{tanaka_recent_2019}.
Next-generation reservoir computing (NG-RC) \cite{gauthier_next_2021} has been introduced as an alternative to ESN-type RC: it can be more efficiently trained, entails better control over hyperparameters, and is easier to interpret due to an explicit feature library.
Recent applications include, in particular, complex spatio-temporal and non-stationary chaotic systems \cite{shahi_prediction_2022,barbosa_learning_2022,flynn_exploring_2022,koglmayr_extrapolating_2024,brucke_benchmarking_2024,schotz_machine_2025}.
Historically, NG-RC has been obtained by approximating ESN-type RCs as a nonlinear vector autoregressive process (NVAR) \cite{bollt_explaining_2021}. NVARs have a long history in time series forecasting \cite{leontaritis_inputoutput_1985,kekatos_sparse_2011,billings_nonlinear_2013}.
In the context of static data processing, NG-RCs are also known as extreme learning machines  \cite{huang_extreme_2006,vanheeswijk_adaptive_2009,huang_extreme_2011,butcher_reservoir_2013}, which are closely related to random feature methods \cite{rahimi_random_2007}, i.e., single-layer feed-forward networks with random weights.
Evolving a hand-chosen dictionary (e.g., instantaneous and time-delay coordinates, and their low-order products) without necessarily projecting back to the physical state is practiced in the Extended Dynamic Mode Decomposition (EDMD) method \cite{williams_data_2015,brunton_datadriven_2019}.
Since the training objective enforces that the dictionary evolves within itself, EDMD renders a finite-dimensional approximation to the Koopman operator \cite{tu_dynamic_2014,chen_variants_2012,kutz_dynamic_2016,korda_convergence_2018,mezic_koopman_2020,surana_koopman_2020,brunton_modern_2022}.
The Koopman operator approach constitutes an overarching theoretical framework for modeling complex dynamical systems and has found numerous applications \cite{mauroy_koopmanbased_2019,bevanda_koopman_2021,otto_koopman_2021,ghosh_koopman_2024,shi_koopman_2026}.
Choosing a monomial basis for the Koopman operator leads to the Carleman linearization method \cite{kowalski_nonlinear_1991}.

Characterizing the representations that a model learns from its training data is a central issue of system identification and inverse problem theory \cite{overschee_subspace_1996,ljung_system_1999,billings_nonlinear_2013,brunton_discovering_2016,iten_discovering_2020,lai_structural_2021,fronk_interpretable_2023,churchill_flow_2023,chen_deeposg_2023,yu_learning_2024,koltai_koopman_2024,zhang_quantitative_2024,wang_identifiability_2024,schonlieb_datadriven_2025}.
Integrable dynamical systems can in principle be described by many equivalent surrogate models as a consequence of the underlying symmetries \cite{shumaylov_when_2025,parikh_why_2026}.
In teacher–student setups with well-specified (bias-free) models, the student can learn the data-generating process up to a similarity transform \cite{roeder_linear_2021}, resulting in near perfect generalization \cite{gyorgyi_firstorder_1990,seung_statistical_1992,watkin_statistical_1993,hess_generalized_2023}. 
Physics-informed learning supports generalization by incorporating additional structure into a model \cite{karniadakis_physicsinformed_2021,adler_physicsinformed_2024}.

A straightforward approach to system identification is provided by the SINDy method \cite{brunton_discovering_2016} and its refinements \cite{messenger_weak_2021,russo_convergence_2024,pecile_datadriven_2025}. 
However, these methods typically require approximating the time derivative of the state variables, which can limit robustness for noisy data \cite{messenger_weak_2021}.
We instead focus on the discrete-time setting, where the dynamics is specified by the flow map.
Moreover, the presence of delays can obscure the direct relationship between the readout coefficients and the flow map \cite{gauthier_next_2021,zhang_catch22s_2023,zhang_how_2025}.
Various studies have discussed this in the context of numerical instability \cite{santos_emergence_2025}, noise robustness \cite{liu_noise_2023}, and empirical time-stepping formulations \cite{chen_next_2022,zhang_how_2025}.
For stationary signals, viable surrogate models are harmonic time series \cite{kay_spectrum_1981,rath_revisiting_2012,chen_variants_2012}. These can be exactly represented by linear delay models (linear VARs) \cite{so_linear_2005,vaseghi_advanced_2008,pan_structure_2020}. 
If the underlying system is linear or follows a limit cycle (periodic motion), the approximation error is mainly limited by the number of frequencies included in the model. 
Nonlinearities in the underlying system are reflected in coupled phases and nontrivial polyspectra (i.e., Fourier-transformed higher-order cumulants of the process) of the harmonic surrogate. 

The discrepancy between one-step training error and autonomous forecast accuracy is well known in system identification as the distinction between prediction-error and simulation-error fitting \cite{piroddi_simulation_2008,aguirre_prediction_2010,schar_surrogate_2025}, and in sequence modeling as the teacher-forcing or exposure-bias problem \cite{bengio_scheduled_2015,huszar_how_2015,schmidt_generalization_2019}. One-step-ahead minimization can select models with poor autonomous rollout behavior, especially for highly sampled data where adjacent lags are strongly correlated \cite{zhang_how_2025,somalwar_learning_2025}. 
In Koopman-based approaches such as EDMD, prediction accuracy is known to depend on how close the dictionary is to being Koopman-invariant, and that residual error alone does not encode the quality of the feature space \cite{haseli_temporal_2022,haseli_generalizing_2023,haseli_invariance_2025}.

In the present work, we consider the problem of learning flow maps of fully observed Markovian dynamical systems using NVAR/NG-RC models with instantaneous and delay-augmented feature libraries. 
We formulate a training error scaling theory based on the Lie-series expansion of the flow map and specialize it to polynomial and Fourier feature libraries.
Closely related to the present work are also error analyses of the Koopman and Carleman linearization approaches \cite{forets_explicit_2017,korda_convergence_2018,klus_datadriven_2020,kostic_learning_2022,amini_carleman_2022,nuske_finitedata_2022,haseli_generalizing_2023,philipp_variance_2024,zhang_quantitative_2024,haseli_invariance_2025}, including polynomial and delay-augmented embeddings \cite{kamb_timedelay_2020,iacob_finite_2023}, and of learned multi-step methods \cite{raissi_multistep_2018,du_discovery_2022}.
The present work hopes to contribute a useful addition to these studies in the context of NVAR/NG-RC modeling.

\section{NVAR/NG-RC Model}
\label{sec_ngrc_modeling}

In the Koopman/EDMD formulation \cite{brunton_datadriven_2019,mezic_koopman_2020,brunton_modern_2022}, the next-step prediction problem is given by
\beq \psiv(t+\Delta t) = K \psiv(t),
\label{eq_nextstep_task}\eeq 
where the feature vector $\psiv(t)=\Gv(\xv(t),\xv(t-\tau),\ldots, \xv(t)^2,\ldots )\in \reals^n$ is a dictionary of nonlinear or delayed observables constructed from some time series $\xv(t)\in \reals^d$.
The matrix $K\in \reals^{n\times n}$ approximates the Koopman operator on the space defined by the observable set $\psiv$. It is estimated from $P$ observations via linear regression, 
\beq \hat K=\arg\min_{K} \sum_t \|\psiv(t+\Delta t)- K\psiv(t)\|^2 = \arg\min_K \| \Psiv_{t+\Delta t} - K \Psiv_t \|^2,
\label{eq_nvar_optprob}\eeq 
where 
\beq \Psiv_t=[\psiv(t_1), \psiv(t_2),\ldots, \psiv(t_P)]\in \reals^{n\times P}
\label{eq_design_matrix}\eeq 
is a design matrix with columns given by the time-sampled observables $\psiv(t_i)$.
If $P>n$ and $\Psiv_t$ has full row rank, the optimal solution is given in terms of the pseudoinverse by 
\beq \hat K = \Psiv_{t+\Delta t} \Psiv_t^+ = \Psiv_{t+\Delta t}\Psiv_t^T (\Psiv_t \Psiv_t^T)^{-1}.
\label{eq_nvar_optsol}\eeq 
In order to improve conditioning when features are highly correlated, the pseudoinverse can be replaced by a Tikhonov-regularized form $\Psiv_t^T (\Psiv_t \Psiv_t^T + \lambda \Imat)^{-1}$ with regularization parameter $\lambda$. 
In particular, when $\psiv$ consists of delay observables with small lag $\tau\simeq \Dt$, scaling $\lambda \propto P$ mitigates the growth of the condition number \cite{zhang_how_2025}.
\Cref{eq_nvar_optsol} can be compactly expressed in terms of empirical correlation matrices $\hat C(\Delta t) = \frac{1}{P}\sum_{p=1}^P \psiv(t_p+\Delta t)\psiv(t_p)^T$:
\beq \hat K = \hat C(\Delta t) \hat C(0)^{-1} \simeq \Imat + \Delta t \hat C'(0) \hat C(0)^{-1},
\label{eq_linreg_correl}\eeq 
where the last approximation is valid for small $\Delta t$.

In the NG-RC method, instead of \cref{eq_nextstep_task}, one uses only the state-part of $\psiv$ as the target and thus considers the regression problem
\beq \xv(t+\Delta t)=Q \psiv(t+\Delta t) \simeq H \psiv(t),\qquad Q=\begin{bmatrix}I_d & \mathbf{0}_{d\times(n-d)}\end{bmatrix}\in\mathbb{R}^{d\times n}
\label{eq_ngrc_reduct}\eeq 
where $H\in \reals^{d\times n}$ denotes the NG-RC next-step operator and the projector $Q$ selects $\xv$ from the observable set $\psiv$.
Thus, the minimization problem of \cref{eq_nvar_optprob} is replaced by $\hat H = \arg\min_H \|X_{t+\Dt} - H \Psiv_t\|^2$. Using $X_{t+\Dt} = Q \Psiv_{t+\Dt}$ gives
\beq \hat H = X_{t+\Dt} \Psiv_t^+ = Q \hat K,
\eeq 
showing that \cref{eq_nextstep_task} contains the solution of \cref{eq_ngrc_reduct}.
Predicting beyond the next time step involves recursively lifting the state to the latent space and applying the learned next-step operator: 
\beq \hat \xv(t+\Dt) = \hat H \Gv(\hat \xv(t),\ldots),\qquad \hat\xv(0)=\xv_0.
\label{eq_ngrc_autonom}\eeq 
The result differs, in general, from $Q\hat K^j \Gv(\xv(0),\ldots)$, because the selected observables $\psiv$ typically do not form a Koopman-invariant subspace on which $\Gv(\xv(t+\Dt),\ldots)=\hat K\Gv(\xv(t),\ldots)$ would hold.

\section{Learning Markovian dynamical systems}
\label{sec_markov_flow}
We now discuss learning of flow maps of fully observed Markovian dynamical systems using instantaneous and delay-augmented feature libraries.
Consider a Markovian (and possibly nonlinear) dynamical system described by 
\beq \dot\xv(t) = \fv(\xv(t))
\label{eq_dynsys_ODE}\eeq 
with a state vector $\xv$ and a vector field $\fv:\Omega\subset \reals^d \to \reals^d$.
The discrete time evolution is described by the associated flow map (propagator) $\Phi_\Dt(\xv(t))$, which, for small $\Dt$, can be expressed as a Taylor/Lie series: 
\beq\begin{split} \xv(t+\Delta t) = \Phi_\Dt(\xv(t))  
&=  \xv(t) + \fv(\xv(t)) \Delta t + \sum_{l=1}^d \frac{\partial \fv}{\partial x_l} f_l(\xv(t)) \frac{(\Delta t)^2}{2!} + \sum_{q=3}^{\infty} \frac{(\Delta t)^q}{q!} \frac{d^q}{d t^q} \xv(t) \\
&= \xv(t) + \sum_{q=1}^\infty \frac{(\Delta t)^q}{q!}\Fv_q(\xv(t)) ,
\end{split}\label{eq_flowmap}\eeq 
where $\Fv_q$ denotes the $q$-th Lie-series coefficient, defined recursively by $\Fv_1 = \fv$ and
$\Fv_{q+1} = L_\fv \Fv_q = (\nabla \Fv_q)\fv$, where $L_\fv \hv := \fv \cdot \nabla \hv$ denotes the Lie derivative along the vector field $\fv$.
Equivalently, along trajectories one has $\Fv_q(\xv(t)) = \frac{d^q}{dt^q}\xv(t)$.

We focus henceforth on the case where the vector field $\fv$ has \emph{polynomial} dependence on $x_j$ with a \emph{maximum degree} $p$.
The degree $M'$ of monomials occurring in the expanded flow map $\Phi_\Dt$, truncated at the $r$-th time derivative, can be estimated by noting that $\frac{d^r}{dt^r} x(t)$ has a maximum total polynomial degree of 
\beq M'= r(p-1) + 1,
\label{eq_flowmap_polydeg}\eeq 
leaving an error of $\Ocal((\Delta t)^{r+1})$ in the Taylor series \footnote{The equality for $M'$ holds for the generic case, disregarding possible cancellations or highly structured flow maps. Moreover, the flow map of a polynomial vector field is in general not polynomial, but analytic. For example, for $\dot x = x^2$, the exact flow map is $\Phi_{\Delta t}(x)=\frac{x}{1-\Delta t x} = x + \Delta t x^2 + \Delta t^2 x^3 + \cdots$.}.

\subsection{Learning with instantaneous feature libraries}
\label{sec_inst_learn}

We consider now the specific NG-RC regression problem, where one learns an empirical flow map by fitting a given feature library $\gv(\xv)\in\reals^n$ 
\beq \xv(t+\Delta t) \simeq H \gv(\xv(t)),
\label{eq_NGRC}\eeq
where $H\in \reals^{d\times n}$ is a matrix of trainable weights.
The associated (mean-square) \emph{training error} is given by
\beq E\st{train} \equiv \varepsilon\st{train}^2 = \frac{1}{P}\sum_{j=1}^P \|\xv(t_j+\Dt) - \hat H \gv(\xv(t_j)) \|^2 ,
\label{eq_ngrc_trainerr}\eeq 
with $\varepsilon$ denoting the root-mean-square error (RMSE).
Analogously, a \emph{test} error $E\st{test} = \E \|\xv(t+\Dt) - \hat H \gv(\xv(t)) \|^2$ can be defined by replacing the sample average by an average over the statistical distribution of $\xv$. 
In the large sample number limit ($P\to\infty$), and assuming ergodicity/mixing, the training error $E\st{train}$ converges to the test error $E\st{test}$.

We assume in the following that the state $\xv$ can be exactly represented by $\gv$. 
Using \cref{eq_flowmap} and expanding the learned flow map as $\hat H \gv(\xv) = \xv + \Dt \hat \fv(\xv) + \Ocal(\Dt^2)$, the training error becomes $E\st{train} \simeq \Dt ^2 \E \| \fv(\xv) - \hat \fv(\xv) \|^2 + \Ocal(\Dt^3)$ (large $P$ limit).
This motivates defining the \emph{truncation error} \cite{langtangen_finite_2017}
\beq E\st{trunc}\equiv \varepsilon^2\st{trunc}\equiv \frac{E\st{train}}{\Dt^2}
\label{eq_ngrc_truncerr}\eeq
as a measure indicating how well the flow map is learned.
Accordingly, $E\st{trunc}=\const$ as $\Dt\to 0$ indicates that already the vector field $\fv$ is not exactly representable by the feature library $\gv$.
This heuristic is made precise in \cref{sec_feature_library_structure}. 

Focusing now on the case of \emph{polynomial} vector fields, we can distinguish two cases:
\begin{enumerate}
    \item \emph{Polynomial feature library.} If $\gv$ consists of \emph{instantaneous monomials} up to total degree $M$ in $x_j(t)$, the largest representable Lie order $r^\ast$ in \cref{eq_flowmap} follows from $M\ge M'$ as 
    \beq r^\ast=\left\lfloor \frac{M-1}{p-1} \right\rfloor,
    \label{eq_dtexp_poly}\eeq 
    where $\lfloor \cdot \rfloor$ is the largest integer less than or equal to the argument. The trivial linear case $p=1$ can be treated separately.
    Accordingly, the training and truncation errors behave as [see \cref{eq:poly_scaling_generic}] 
    \beq E\st{train} \propto \Dt^{2(r^\ast+1)},\qquad E\st{trunc} \propto \Dt^{2 r^\ast}.
    \label{eq_trainerr_poly}\eeq
    For $M<p$, the model is misspecified and $E\st{trunc}>0$ as $\Dt\to 0$.
    \item \emph{Non-polynomial feature library.} 
    \begin{enumerate}
        \item \emph{Generic case}. If the vector field is not exactly representable by the feature span, the trivial asymptotic scaling [\cref{eq:asymptotic_scaling_norep}]
        \beq E\st{train} \propto \Dt^2,\qquad E\st{trunc}=\const.,\qquad \Dt\ll \Dt_\times
        \label{eq_Etrunc_norep}\eeq 
        applies. An estimate for the crossover time scale $\Dt_\times$ is given in \cref{eq:crossover_scale}.
        For larger $\Dt$, a pre-asymptotic scaling law of the form 
        \beq E\st{trunc} \propto \Dt^{2r_*},\qquad \Dt\gg \Dt_\times
        \label{eq_Etrunc_preasymp}\eeq 
        can emerge provided the approximation error of Lie orders earlier than $r_*+1$ is sufficiently small [see \cref{eq:preasymptotic_scaling_generic}]. 
        \item \emph{Fourier feature library}. A particular relevant case is a Fourier feature library, 
        \beq
        \gv(\xv)=\left\{\xv, \sin\!\bigl(\omega_0 \, \kv\cdot \xv\bigr),\cos\!\bigl(\omega_0 \, \kv\cdot \xv\bigr):\kv \in \{0,\dots,N\}^d\right\},
        \label{eq_Fourier_library}
        \eeq
        with $N$ modes (per dimension) and base frequency $\omega_0$ chosen such that $\omega_0 R\ll 1$ on the data domain $\Omega = [-R,R]^d$ (see \cref{sec_Fourier_lib} for details; the state $\xv$ is included here to transfer those results to the full flow-map prediction). Spectral models are widely used in forecasting \cite{lange_fourier_2020}.
        While also here the asymptotic scaling \eqref{eq_Etrunc_norep} holds for small $\Dt$, a pre-asymptotic regime described by \cref{eq_Etrunc_preasymp} with an exponent
        \beq
        r_* \gtrsim \left\lfloor \frac{N-1}{p-1} \right\rfloor
        \label{eq_Fourier_scal_exp}\eeq
        is expected [see \cref{eq:Fourier_preasy_exp}].

    \end{enumerate}
    
\end{enumerate}

The above statements apply also to non-polynomial dynamical systems and generic feature libraries (see \cref{sec_feature_library_structure}), except for the specific predictions involving the polynomial degree $p$ [\cref{eq_dtexp_poly,eq_Fourier_scal_exp}].
Effects of ill-conditioning \cite{zhang_how_2025,santos_emergence_2025} and noise give rise to a lower bound $E\st{floor}(M,\lambda)$ to $E\st{train}$.
When the feature library can reproduce the flow map to some nontrivial order, the training error also controls the forecast horizon (see \cref{app_train_err}).

As will be discussed in detail in \cref{sec_delay_markov}, delay terms do not by themselves remove a structural mismatch to the nonlinear Markovian vector field, but they can change the one-step training-error scaling through multistep extrapolation. Thus the truncation-error diagnostic is cleanest for instantaneous libraries, or for delay models only after separating genuine flow-map representation from delay-induced interpolation.

The scaling laws derived here describe the \emph{approximation-bias} component of the one-step prediction risk (see \cref{app_bias_variance}). A feature library is called \emph{well specified} only if the flow map belongs to the span of the library. 
However, this is usually not fulfilled for finite libraries, since even for polynomial vector fields the exact flow map is typically of non-polynomial form.
We therefore use the weaker notion of being well specified to Lie order $r$, meaning that the first $r$ Lie-series coefficients of the flow map are contained in the feature span. In this sense, a polynomial library with $M\geq p$ is \emph{bias-free} at the vector-field level, whereas polynomial libraries with $M<p$, generic finite Fourier libraries, and linear delay-only models for nonlinear Markovian systems are \emph{misspecified}.

\subsection{Learning Markovian systems with delay-feature libraries}
\label{sec_delay_markov}

We now extend the previous setting by including time-lagged states $\xv(t-k\tau)$, $k=0,\dots,\Gamma-1$ in the feature library (where $\Gamma$ denotes the total number of taps), and analyze how the resulting \emph{multistep predictor} represents the Markovian flow map \eqref{eq_flowmap}. We assume $\tau=\Delta t$ for simplicity. 
The learning problem is then given by
\beq
\xv(t+\Delta t) = K \Gv\big(\{\xv(t-k\Delta t)\}_{k=0}^{\Gamma-1}\big),
\label{eq_delay_pred}
\eeq
where the delay feature library $\Gv \in\mathbb{R}^{D_D}$ is constructed from all monomials in the delay coordinates $\{\xv(t-k\Delta t)\}$ up to some maximal total degree $M$. 

According to \eqref{eq_flowmap}, one has
\beq
\xv(t-k\Delta t) = \Phi_{-k\Delta t}(\xv(t)) = \sum_{l=0}^{\infty} \frac{(-k\Delta t)^l}{l!} \frac{\mathrm{d}^l}{\mathrm{d}t^l} \xv(t) =  \mathbf{R}_k(\xv(t)) + \Ocal\big((k\Delta t)^{r+1}\big),
\label{eq_delay_taylor}
\eeq
where, by truncating the expansion at order $l=r$, we can express each delay state as a vector $\mathbf{R}_k$ of polynomials (of degree at most $M' = r(p-1)+1$) in $\xv(t)$.
Let the feature vector $\boldsymbol{\psi}(\xv) \in \mathbb{R}^{D_I}$ consist of all instantaneous monomials in $\xv$ up to some maximal degree $M_I\ge M'$, such that all components of $\mathbf{R}_k(\xv)$ can in principle be written as linear combinations of the entries of $\boldsymbol{\psi}(\xv)$. 

By \cref{eq_delay_taylor}, every component of the library $\Gv$ is itself a polynomial in $\xv(t)$ of degree at most $M M'$. 
Hence, provided $M_I \ge M M'$, we can express $\Gv$ as a linear combination of the entries of $\boldsymbol{\psi}$:
\beq
\Gv\big(\{\xv(t-k\Delta t)\}_{k=0}^{\Gamma-1}\big)
\approx A \boldsymbol{\psi}(\xv(t)),
\label{eq_delay_change_basis}
\eeq
where $A\in\mathbb{R}^{D_D\times D_I}$ follows from the expansion in \eqref{eq_delay_taylor}. 
Assume that the truncated delay library spans the chosen instantaneous polynomial basis used by the truncated flow map, so that $A$ has full column rank and thus $A^+ A = \Imat_{D_I}$.
As in \cref{eq_NGRC}, we express the Markovian flow map in the instantaneous basis, $\xv(t+\Delta t) = \Phi_{\Delta t}(\xv(t)) \approx H \boldsymbol{\psi}(\xv(t))$ where the entries of $H\in\mathbb{R}^{d\times D_I}$ follow from the Taylor expansion \eqref{eq_flowmap}.
We then invert \eqref{eq_delay_change_basis} to express $\boldsymbol{\psi}$ in terms of the delay library: 
\beq
\xv(t+\Delta t) \approx H A^+ \Gv\big(\{\xv(t-k\Delta t)\}_{k=0}^{\Gamma-1}\big) + \Ocal(\|H A^+\|\Dt^{r+1}).
\label{eq_delay_flow_representation}
\eeq
This shows that a library of delay coordinates can be linearly transformed into a next-step predictor of the form $\hat{K} = H A^+$.
Note, however, that for small $\Delta t$ the columns of $A$ [\cref{eq_delay_change_basis}] become nearly linearly dependent because the different delays share the same time derivatives in \eqref{eq_delay_taylor}, leading to an ill-conditioned pseudoinverse $A^+$ and to the strong correlation and regularization effects observed when using dense delay libraries in NG-RC \cite{zhang_how_2025}. This can also cause the error in \cref{eq_delay_flow_representation} to diverge for $\Dt\to 0$.

Depending on the choice of the library features, the regressor can realize several predictor types via fitting $\hat{K}$. 
If $\Gv$ contains only \emph{linear} delay coordinates, the learned update is linear in the delay state. It may nevertheless achieve small one-step error by exploiting temporal interpolation/extrapolation along the observed trajectory. However, it does not represent a state-only nonlinear Markovian flow map and generally does not yield a consistent autonomous rollout for nonlinear systems (see also \cref{app_delay_extrap}).
This is consistent with the need for lifted observables in Koopman/EDMD approaches \cite{brunton_datadriven_2019,mezic_koopman_2020,brunton_modern_2022}.
Nevertheless, one can use such a linear multistep predictor to understand the general impact of delay features on the training error.
A canonical linear predictor is the extrapolation stencil
\beq
\tilde{\xv}(t+\Delta t)= \sum_{k=0}^{\Gamma-1} \zeta_k \xv(t-k\Delta t) + \Ocal(\Delta t^\Gamma),\qquad \zeta_k=(-1)^k \binom{\Gamma}{k+1}.
\label{eq_delay_extrap_predictor}
\eeq 
The coefficients $\zeta_k$ follow straightforwardly by inserting the Taylor expansion of $\xv(t-j\Delta t)$ into the left and right hand sides and equating the coefficients at equal powers of $\Delta t$.
For generic data and at fixed $M$, this canonical predictor achieves a training error of \footnote{If the data-generating process obeys recurrence relations, other forms of linear predictors may yield lower training errors. For example, for $x(t)=\sin(\omega t)$, a vanishing training error is achieved for $\tilde x(t+\Delta t) = 2\cos(\omega\Delta t) x(t)-x(t-\Delta t)$.}
\begin{align}
E_{\mathrm{train}}(\Gamma) &\le \mathbb{E}\big\|\xv(t+\Delta t)-\tilde{\xv}(t+\Delta t)\big\|^2 \simeq \Dt^{2 \Gamma} \mathbb{E}\|\xv^{(\Gamma)}(t)\|^2 \sim \Ocal(\Delta t^{2 \Gamma}).
\label{eq_delay_trainerr_asy}
\end{align}
Since the predictor \eqref{eq_delay_extrap_predictor} is contained in the feature class for $M\ge 1$, the optimal training error admits the combined bound following from \cref{eq_delay_trainerr_asy,eq_trainerr_poly}:
\beq
E_\mathrm{train}(\Gamma) \sim \Ocal\bigl(\Delta t^{2\max\{r^\ast(M)+1,\Gamma\}}\bigr).
\label{eq_delay_trainerr_asy_maxform}
\eeq
Accordingly, delay coordinates generally decrease the (unregularized) training error monotonically with the number of taps $\Gamma$. Note, however, that these scaling laws can be affected by $\Gamma$-dependent prefactors.
By allowing the $\zeta_k$ to pick up frequency information, the predictor \eqref{eq_delay_extrap_predictor} can represent a harmonic signal, as shown in \cref{app_delay_extrap}.
In numerical experiments, predictor forms specific to the dataset are often observed \cite{chen_next_2022,zhang_how_2025}. 


\section{Numerical experiments}
\label{sec_ngrc_tests}

We now discuss numerical experiments on flow map reconstruction of nonlinear Markovian dynamical systems, using polynomial as well as Fourier feature libraries.
To improve numerical conditioning and stability, we typically train on increments $\xv(t+\Dt)-\xv(t)$ instead of $\xv(t+\Dt)$ \cite{gauthier_next_2021,chen_next_2022}.

\paragraph*{Metrics.} To assess the quality of the flow map reconstruction, we consider, besides the training error [\cref{eq_ngrc_trainerr}], the forecast horizon $T\st{fch}$, defined as the smallest time where the autonomously predicted trajectory $\hat \yv(t)\equiv \hat \xv(t+\Dt)$ [\cref{eq_ngrc_autonom}] deviates more than a certain threshold from the true trajectory $\yv(t)\equiv \xv(t+\Dt)$: 
\beq T\fch = \inf \left\{ t : \| \yv(t) - \hat{\yv}(t) \| > \kappa \sigma \right\},
\label{eq_fc_hor}\eeq 
Here, $\sigma^2= \frac{1}{P}\sum_{i=1}^P \| \bar \yv - \yv(t_i)\|^2$ is the estimated variance, $\bar \yv = \frac{1}{P}\sum_{i=1}^P \yv(t_i)$, and we typically take $\kappa=1$. 
In the noiseless, overdetermined regime ($P\gg n$) considered here, the train-test gap is typically small, so we focus on training error.
We use low solver tolerances $\sim 10^{-13}$ \cite{schotz_machineprecision_2025}, for which the training error can reach $\varepsilon\sim 10^{-13}$.

\paragraph*{Model systems.} As primary benchmarks, we use the Halvorsen and Lorenz-63 models \cite{sprott_elegant_2010}, which have polynomial vector fields of order 2.
The \emph{Halvorsen} model is given by:
\beq 
\begin{aligned}
    \dot x &= -a x - b y - b z - y^2, \\
    \dot y &= -a y - b z - b x - z^2, \\
    \dot z &= -a z - b x - b y - x^2,
\end{aligned}
\label{eq_Halvorsen}\eeq 
with standard parameters $a=1.4$ and $b=4$. Reported values for the largest Lyapunov exponent range between $0.72$ \cite{ma_efficient_2023} and $0.81$ \cite{vaidyanathan_adaptive_2016}, depending slightly on the method used.
The \emph{Lorenz-63} model is given by:
\beq 
\begin{aligned}
\dot x &= \sigma (y - x) ,\\
\dot y &= x (\rho - z) - y ,\\
\dot z &= xy - \beta z ,
\end{aligned}
\label{eq_Lorenz63}\eeq
with standard parameters $\sigma = 10$, $\rho = 28$, $\beta = \sfrac{8}{3}$.
The largest Lyapunov exponent is reported between $0.86$ \cite{ma_efficient_2023} and $0.91$ \cite{ratas_application_2024}.
We also consider two other chaotic systems with cubic nonlinearities:
the \emph{Sprott cubic jerk} system \cite{sprott_simple_1997,vaidyanathan_chaotic_2019}, defined by 
\beq 
\begin{aligned}
\dot{x} &= y, \\
\dot{y} &= z, \\
\dot{z} &= -a z + x y^2 - x^3
\end{aligned}
\label{eq_sprott_sys}\eeq 
with $a=3.6$ and largest Lyapunov exponent $\lambda_1\approx 0.14$, and the \emph{Rabinovich-Fabrikant} system \cite{sprott_chaos_2003}, defined by 
\beq
\begin{aligned}
\dot{x} &= y(z - 1 + x^2) + \gamma x, \\
\dot{y} &= x(3z + 1 - x^2) + \gamma y, \\
\dot{z} &= -2z(\alpha + xy),
\end{aligned}
\label{eq_RabFab_sys}\eeq
with $\alpha=1.1$, $\gamma=0.87$ and largest Lyapunov exponent $\lambda_1\approx 0.12$.

\subsection{Polynomial feature library}

Here, the NG-RC feature library consists of all monomials of the state variables and their delay terms up to depth $\Gamma$ ($\Gamma=1$ corresponding to no delay terms) and nonlinear degree $M$.
We typically whiten the features by subtracting the empirical mean and normalize by the empirical standard deviation.
Specifically, given the design matrix $\Psi \in \mathbb{R}^{n \times P}$ (\cref{eq_design_matrix}, excluding the bias row if present), we use in the NG-RC fit the standardized quantities
\beq
\tilde{\Psi}_{jp} = \frac{\Psi_{jp} - \mu_j}{\sigma_j},\qquad \text{where}\quad \mu_j = \frac{1}{P} \sum_{p=1}^{P} \Psi_{jp}, \quad \sigma_j^2 = \frac{1}{P} \sum_{p=1}^{P} (\Psi_{jp} - \mu_j)^2.
\label{eq_whitened_design_matrix}\eeq

\begin{figure}[t]
    \centering
    \subfigure[]{\includegraphics[width = 0.47\linewidth]{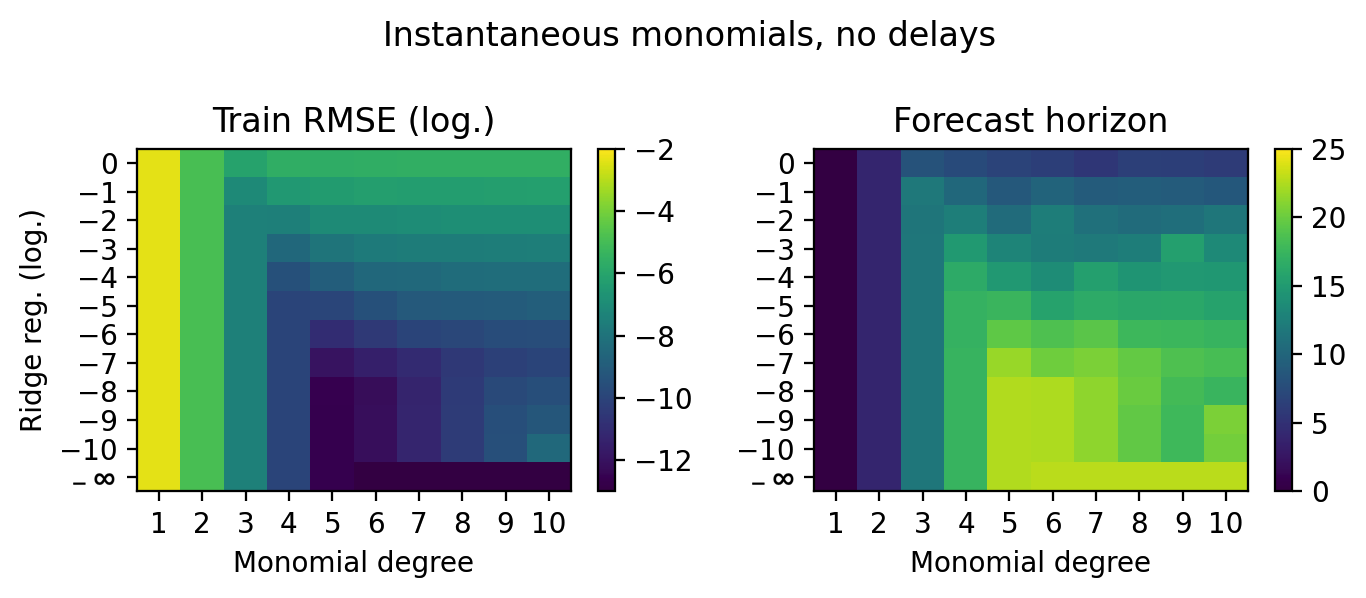}}\qquad
    \subfigure[]{\includegraphics[width = 0.47\linewidth]{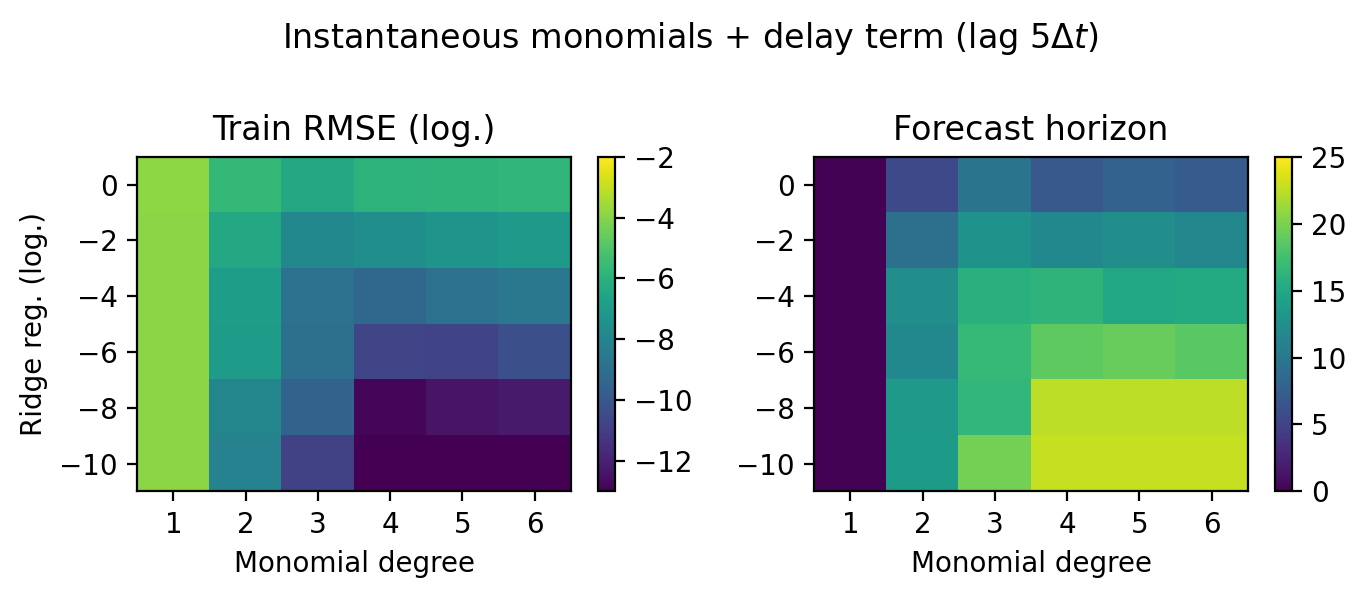}}
    \caption{Effect of increasing the monomial degree $M$ on the training error and forecast horizon for the Halvorsen model, without any delay terms (a) and with a single delay with lag $\tau=5\Dt$ (b). The results in (b) remain similar for other values of the lag $\tau$. The training data is generated using $\Dt=0.001$. Note that the last plotted row in (a) corresponds to Ridge regularization $\lambda=0$.} 
    \label{fig_ngrc_monomial_eff}
\end{figure}

\begin{figure}[t]
    \centering
    \subfigure[]{\includegraphics[width = 0.77\linewidth]{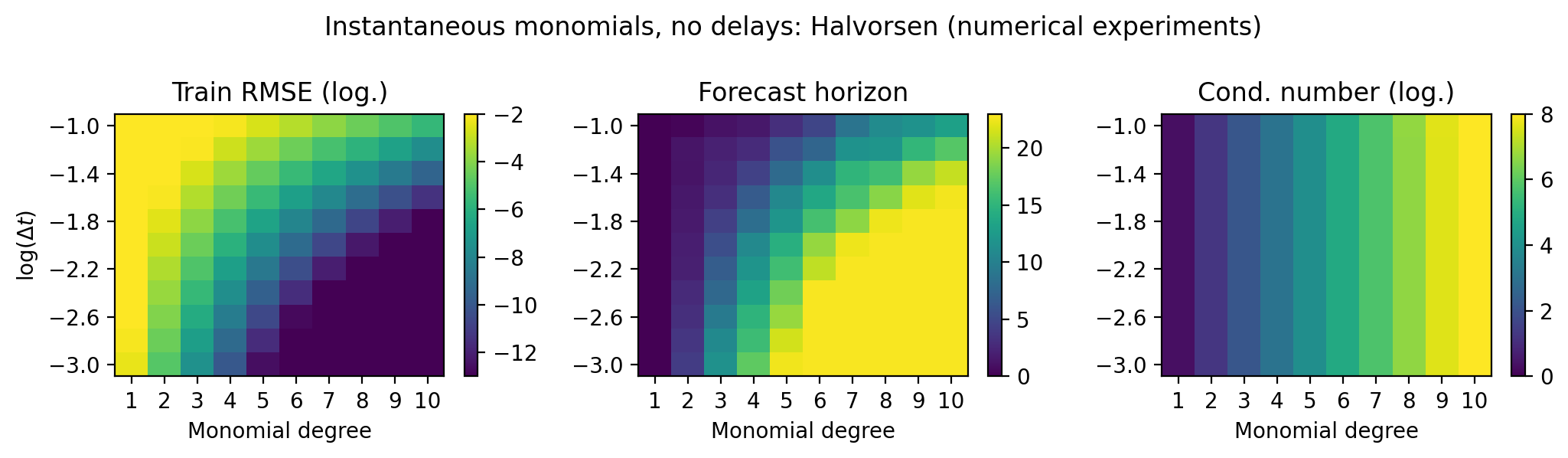}}
    \subfigure[]{\includegraphics[width = 0.34\linewidth]{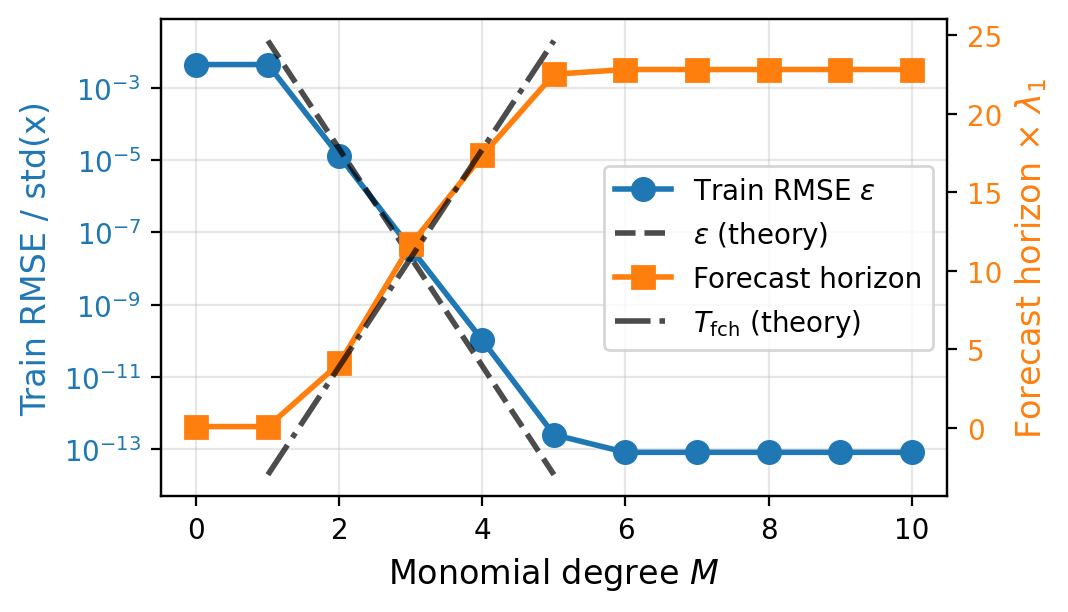}}
    \subfigure[]{\includegraphics[width = 0.32\linewidth]{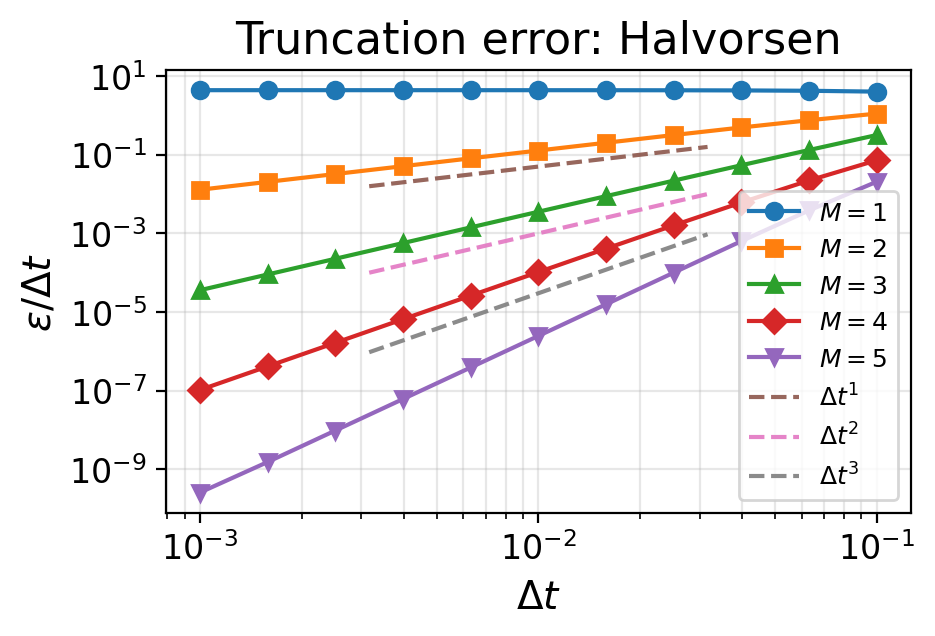}}
    \subfigure[]{\includegraphics[width = 0.32\linewidth]{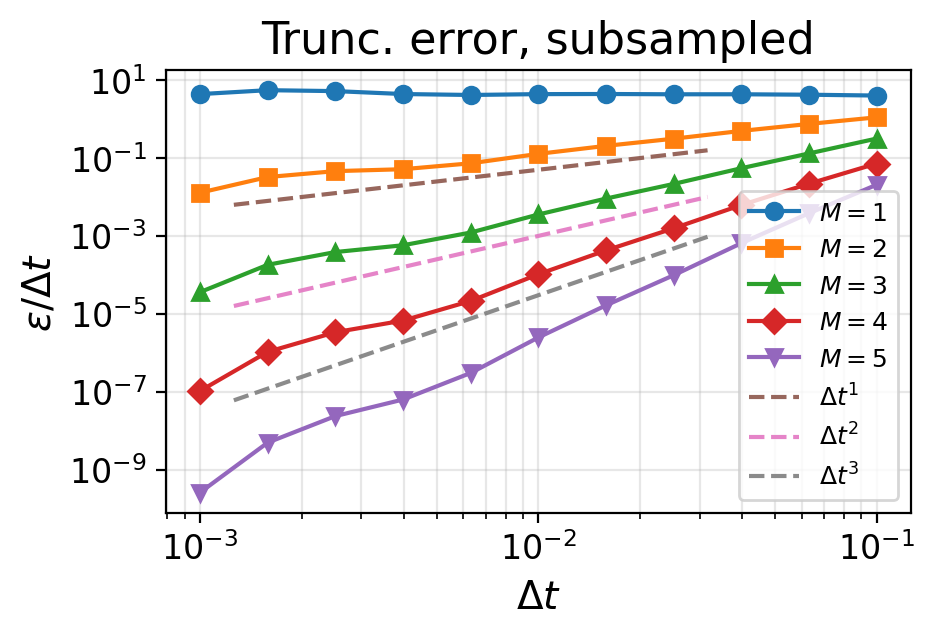}}
    \caption{(a) Dependence of the training error, forecast horizon, and the condition number (of the training design matrix $\Psiv$, \cref{eq_design_matrix}) on the time resolution $\Dt$ of the training data and of the monomial degree $M$ of the feature map for the Halvorsen model. (b) RMS training error $\varepsilon$ (normalized by the standard deviation of the training data) and forecast horizon $T\st{fch}$ vs.\ monomial degree $M$ (for time resolution $\Dt=0.001$). The dashed line gives the theoretical prediction $\varepsilon \simeq a \Dt^{\lfloor(M-1)/(p-1)\rfloor+1}$ [\cref{eq_trainerr_poly}], where $p=2$ is the maximum degree of the monomials in the ODE and $a\approx 20$ is obtained from a fit. The dash-dotted line represents $T\st{fch} = \lambda_1^{-1}\ln(C/\varepsilon)$ [\cref{eq_forec_hor}], where the constant $C\approx 10^{-3}$ subsumes numerical parameters. (c) Truncation error $\varepsilon/\Dt$ [\cref{eq_ngrc_truncerr}] for various monomial degrees $M$ of the feature map. (d) Truncation error obtained by subsampling trajectories generated originally for $\Delta t_0=10^{-3}$. In all plots, we set the Ridge regularization parameter $\lambda=0$.} 
    \label{fig_ngrc_monomial_dt_hal}
\end{figure}

\begin{figure}[t]
    \centering
    (a)\includegraphics[width = 0.55\linewidth]{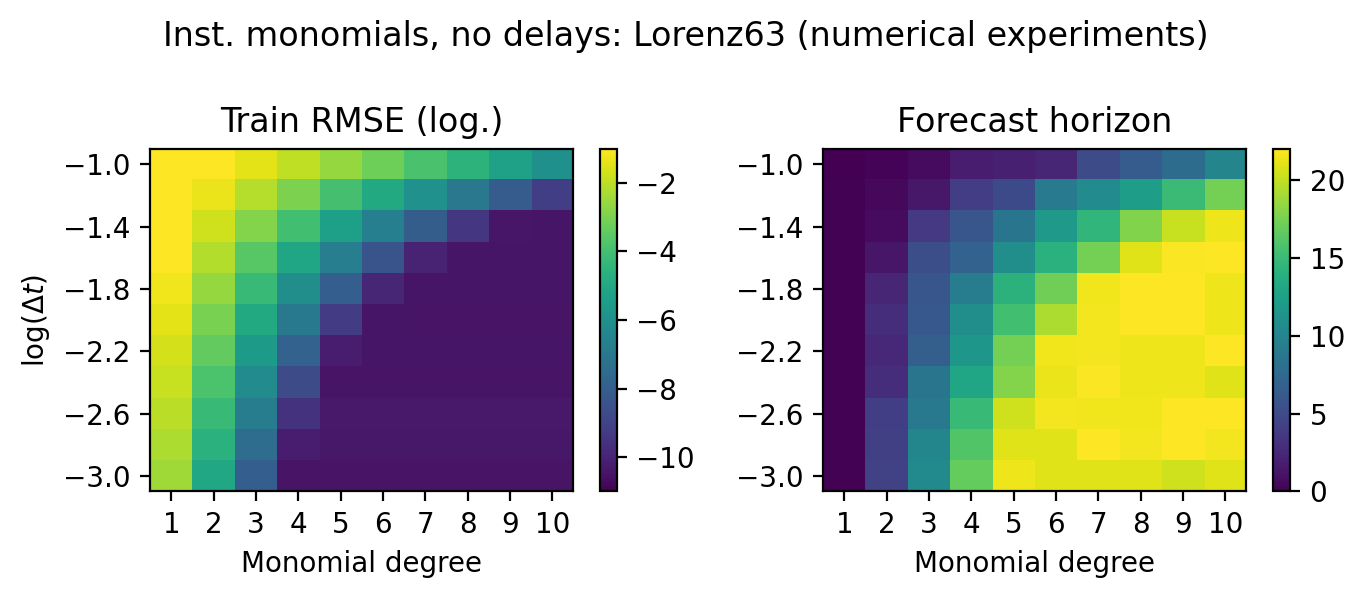}\qquad
    (b)\includegraphics[width = 0.32\linewidth]{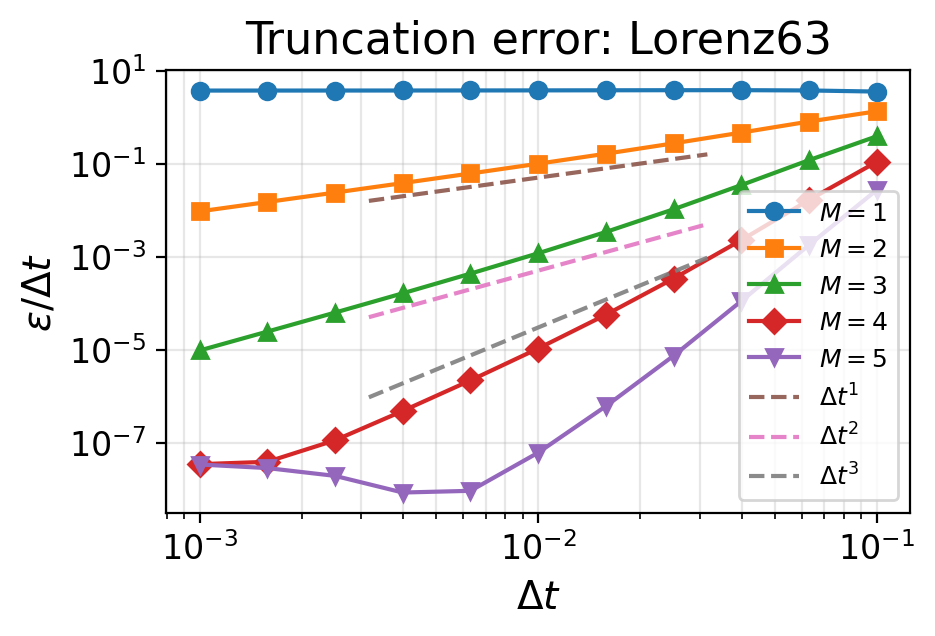}
    (c)\includegraphics[width = 0.55\linewidth]{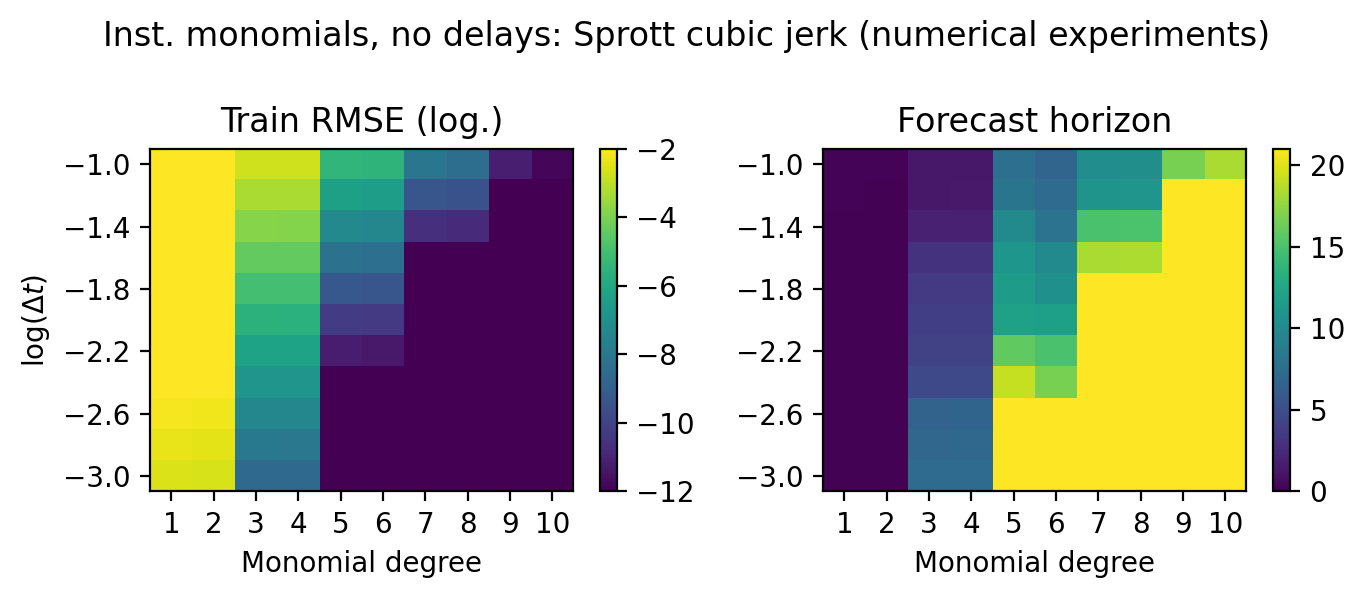}
    (d)\includegraphics[width = 0.32\linewidth]{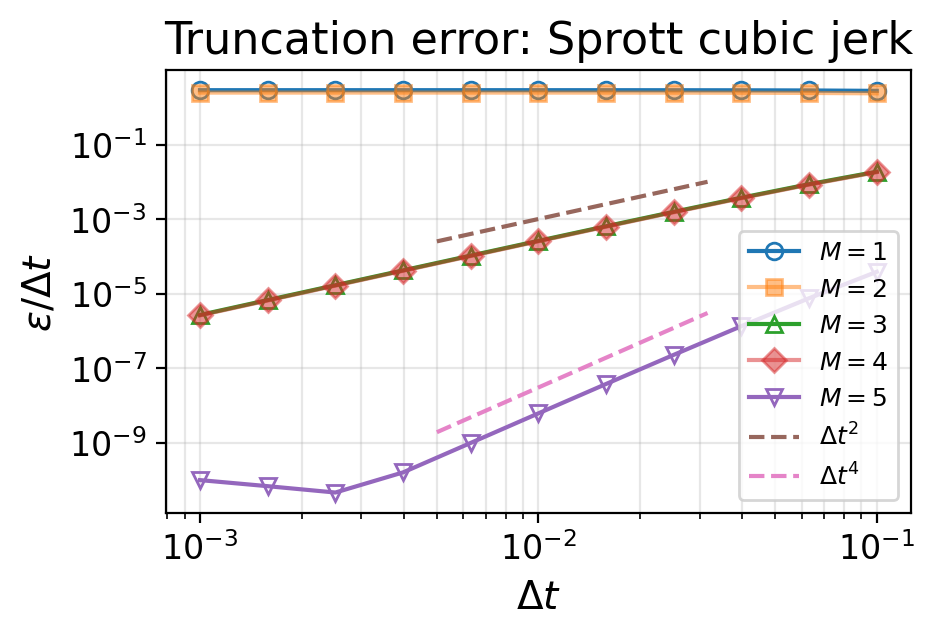}
    (e)\includegraphics[width = 0.55\linewidth]{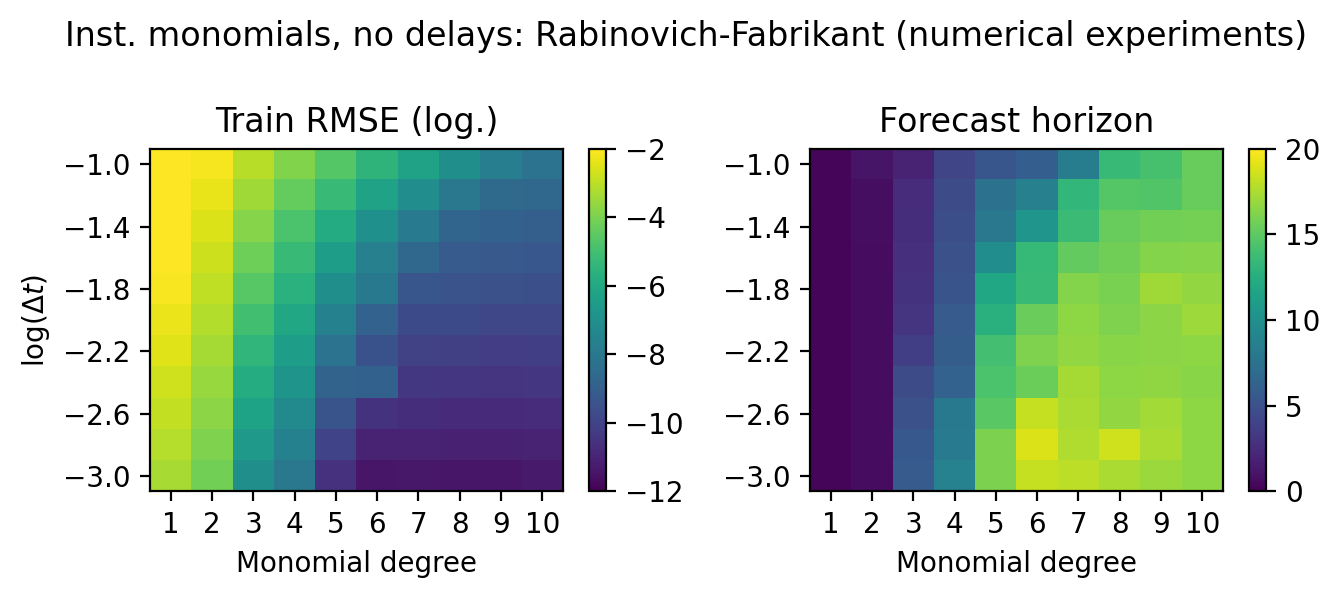}
    (f)\includegraphics[width = 0.32\linewidth]{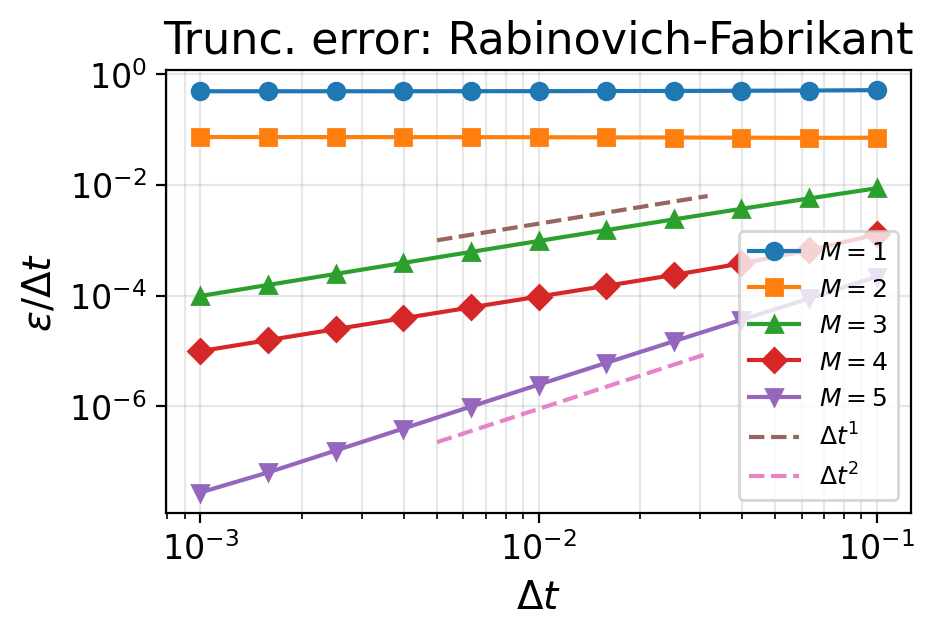}
    \caption{(a,c,e) Dependence of the training error and forecast horizon (in units of the Lyapunov time) on the time resolution $\Dt$ of the training data and of the monomial degree $M$ of the feature map for the Lorenz-63, the Sprott cubic jerk, and the Rabinovich-Fabrikant system, where the latter two have cubic nonlinearities. (b,d,f) Truncation error $\varepsilon/\Dt$ for various monomial degrees $M$ of the feature map. In all plots, we set the Ridge regularization parameter $\lambda=0$.} 
    \label{fig_ngrc_monomial_dt_other}
\end{figure}

\begin{figure}[tb]
    \centering
    \includegraphics[width = 0.75\linewidth]{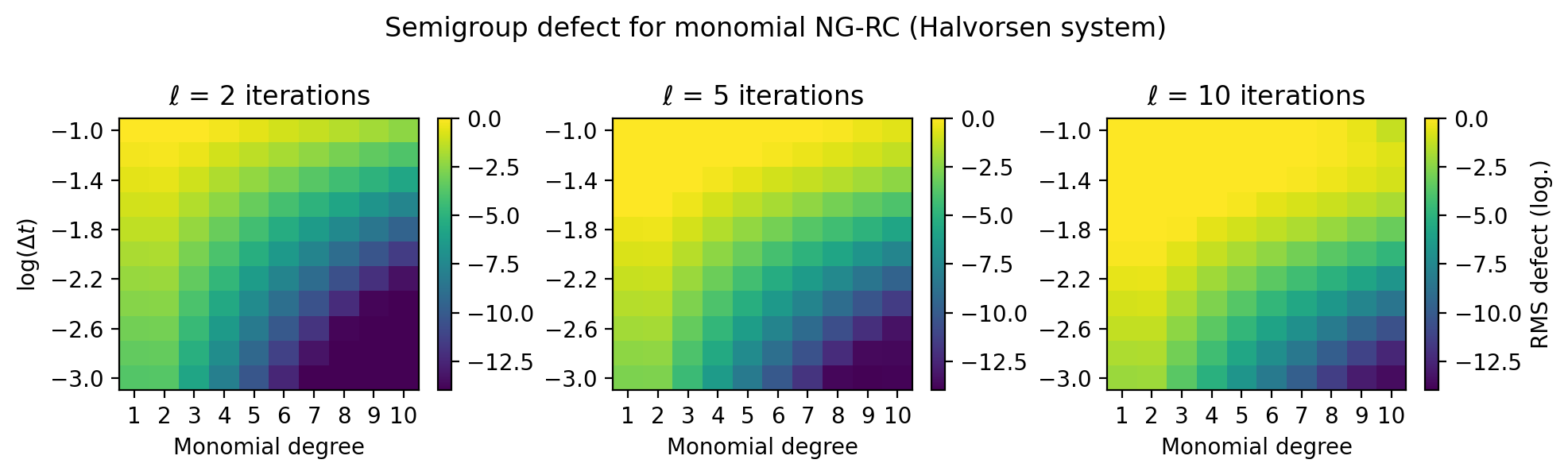}
    \caption{Semigroup consistency test. The semigroup defect $\Scal_\ell$ [\cref{eq_semigroup_consistency}] is shown for various iteration lengths $\ell$ as function of temporal resolution $\Dt$ of the trajectories and monomial degree of the feature map trained on the Halvorsen system. Since the feature map can reproduce the early Lie derivatives, the semigroup defect closely follows the training error, see \cref{fig_ngrc_monomial_dt_hal}.}
    \label{fig_ngrc_semigroup}
\end{figure}

\begin{figure}[t]
    \centering
    \subfigure[]{\includegraphics[width = 0.47\linewidth]{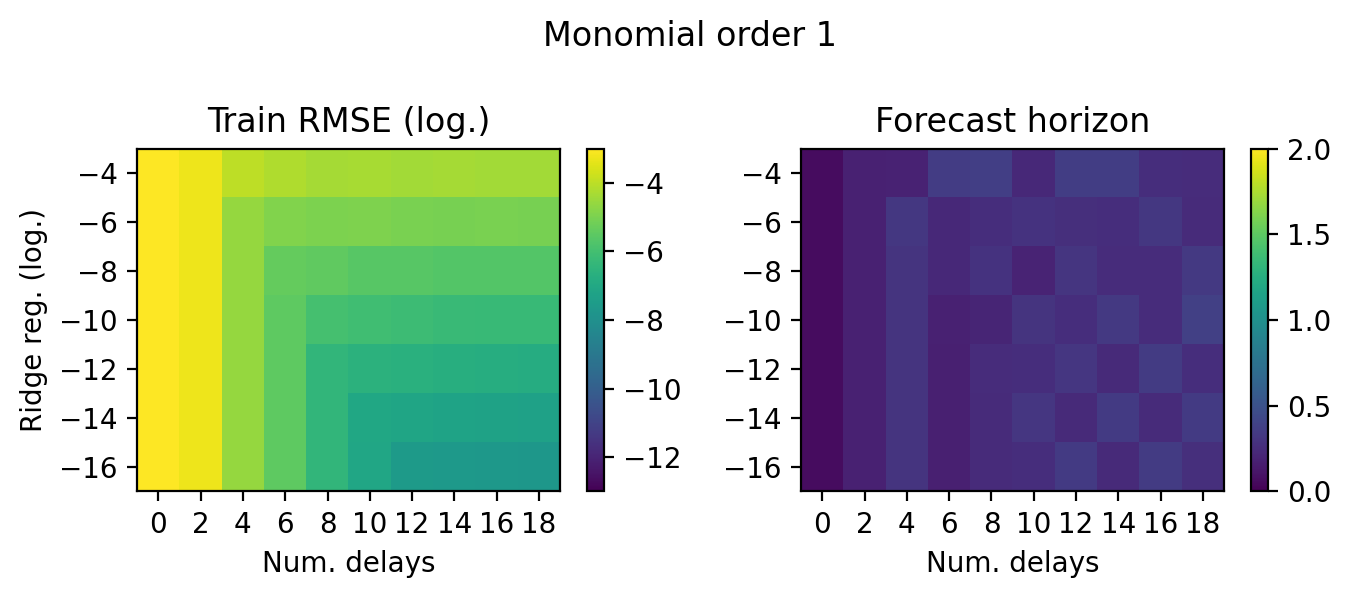}}\qquad
    \subfigure[]{\includegraphics[width = 0.47\linewidth]{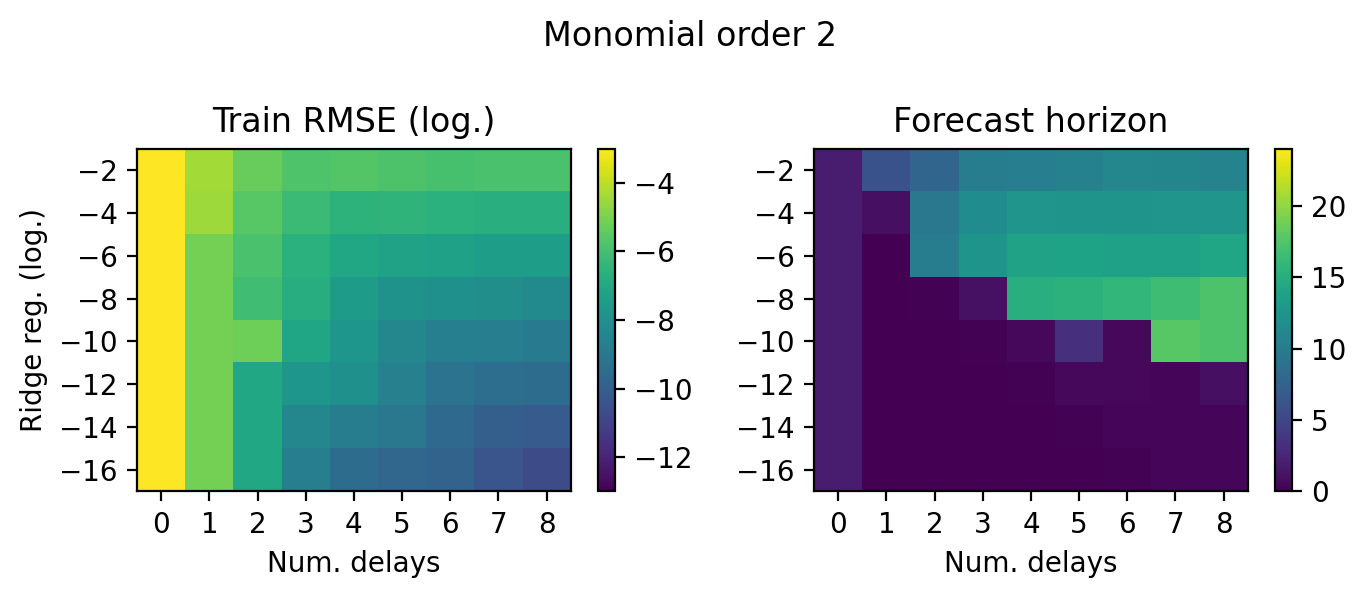}}
    \subfigure[]{\includegraphics[width = 0.47\linewidth]{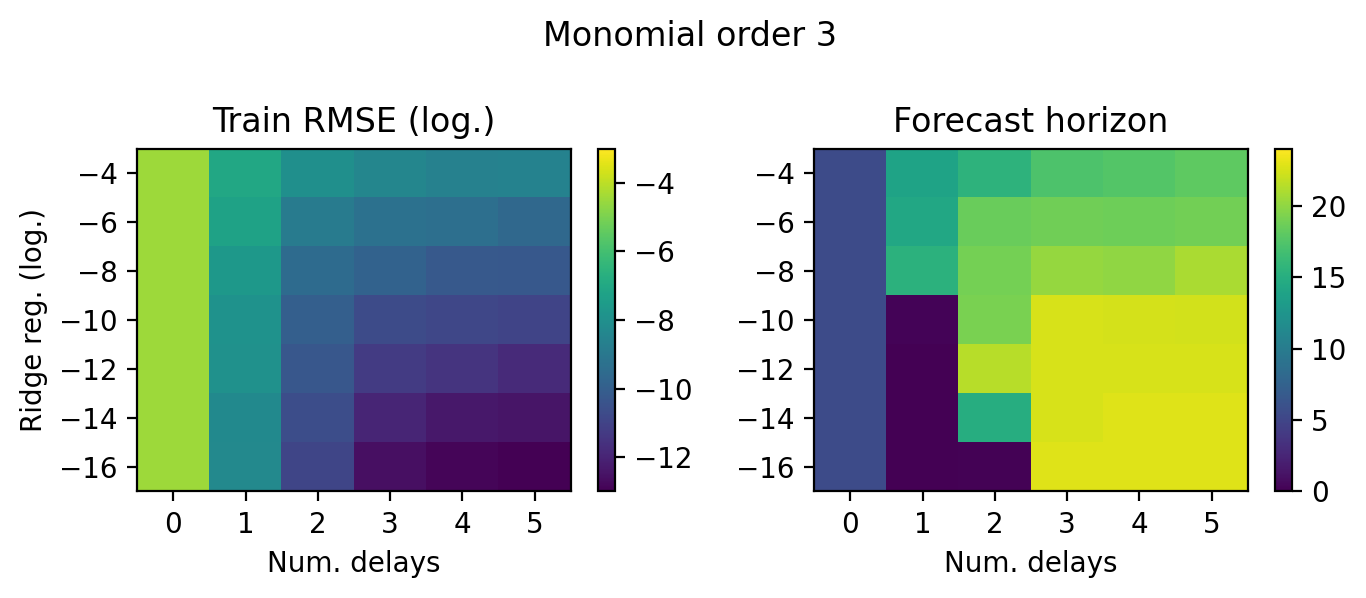}}\qquad
    \subfigure[]{\includegraphics[width = 0.47\linewidth]{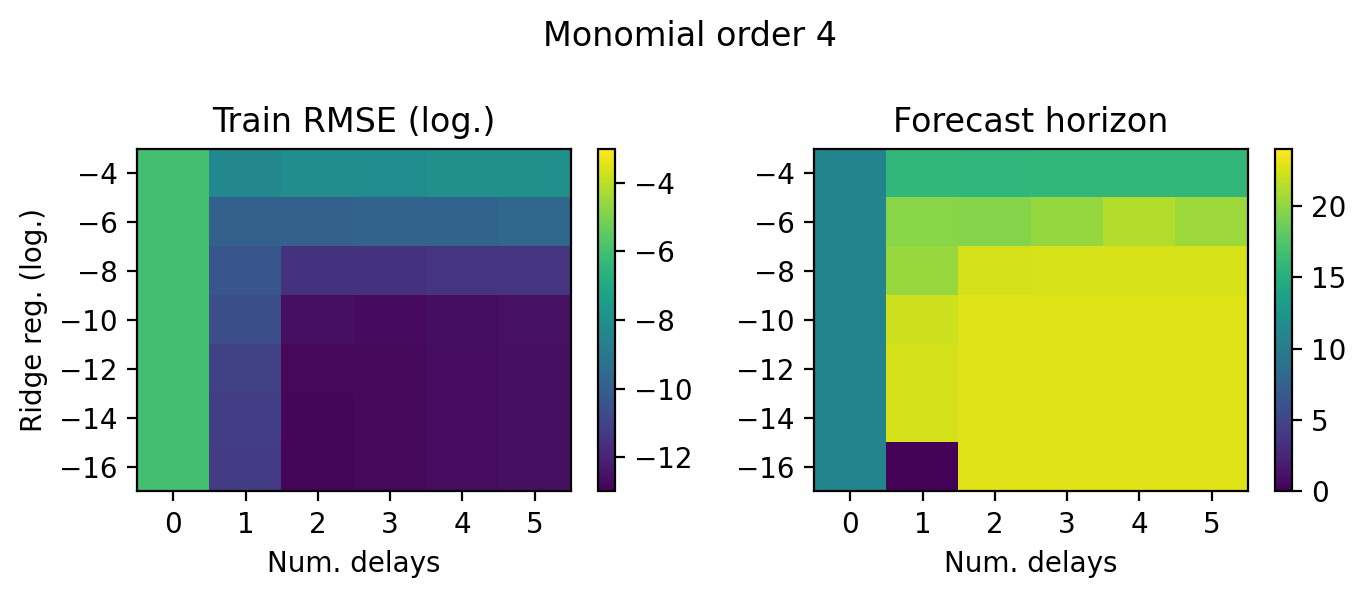}}
    \caption{Effect of delay coordinates on the training error and forecast horizon for the Halvorsen model. The feature library contains only linear delay states in $x(t),y(t),z(t)$ in (a), all monomials up to 2nd order in (b), 3rd order in (c), and 4th order in (d). We use $\Dt=0.01$ to generate the training data, and $\tau=\Dt$ as the delay lag. Forecast horizon is given in units of the Lyapunov time $\lambda_1^{-1}$.} 
    \label{fig_ngrc_delay_eff}
\end{figure}

\begin{figure}
    \centering
    \includegraphics[width = 0.35\linewidth]{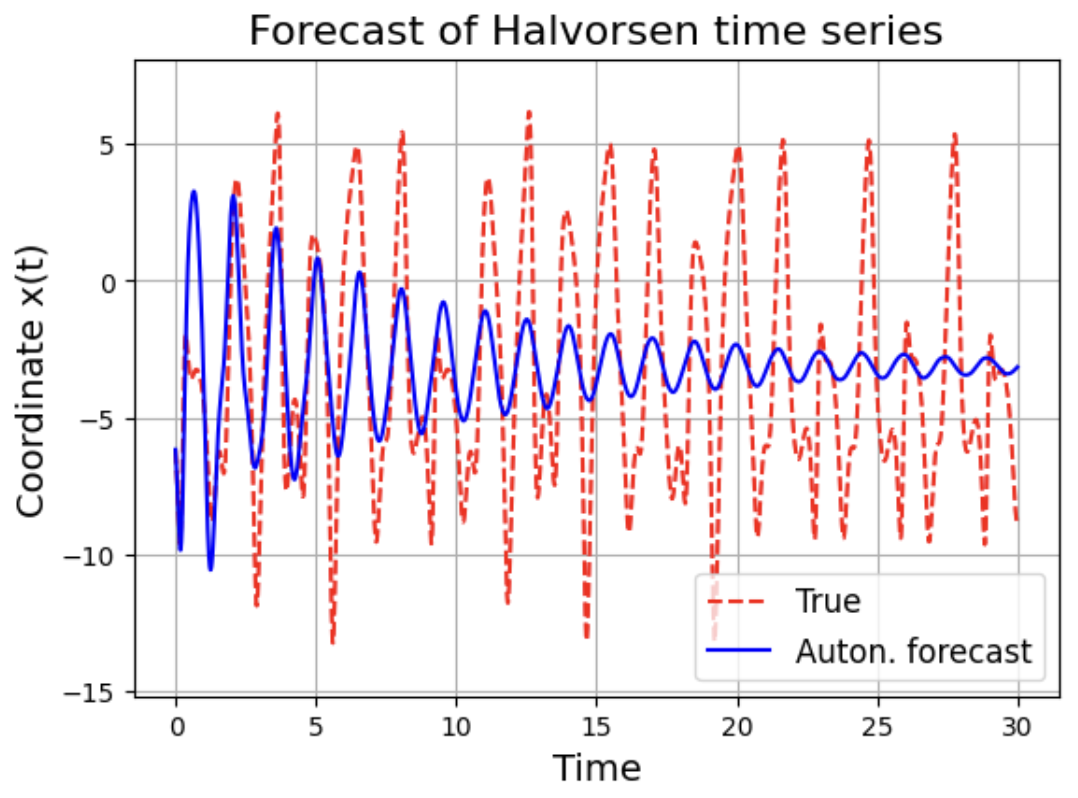}
    \caption{A linear delay predictor learns a harmonic approximation (solid curve) of a (nonlinear) time series (Halvorsen model, dashed curve). We consider the same setting as in \cref{fig_ngrc_delay_eff} and use $\lambda=0$, $\Dt=0.01$, and a delay lag $\tau=\Dt$ with $\Gamma=150$ delay taps; the training error is $\varepsilon\approx 10^{-7}$, while the forecast horizon is close to 0.}
    \label{fig_ngrc_harmonic_learn}
\end{figure}

\begin{figure}[t]
    \centering
    \subfigure[]{\includegraphics[width = 0.495\linewidth]{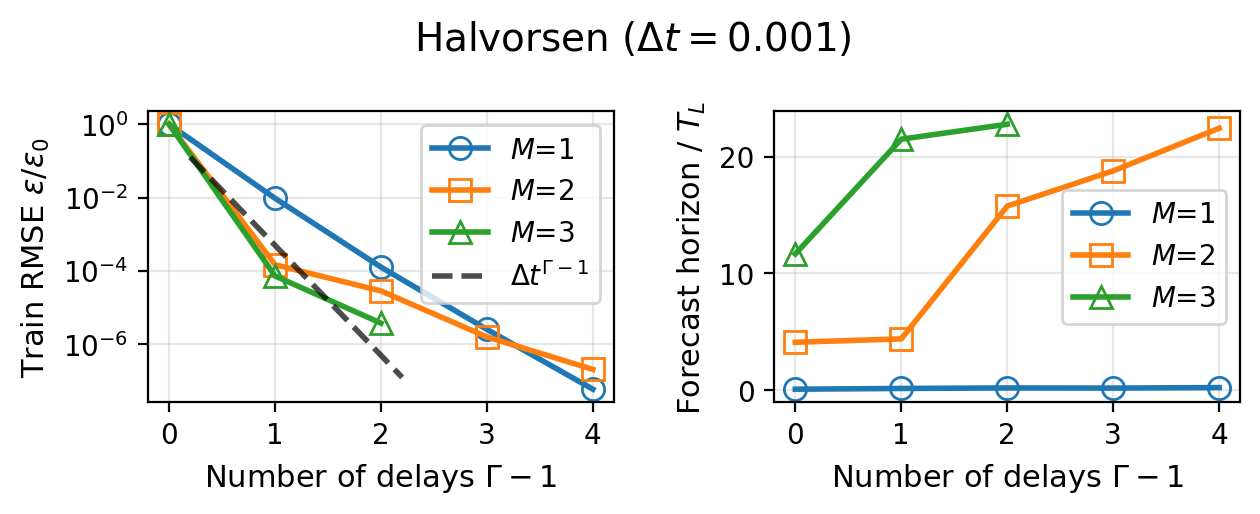}}
    \subfigure[]{\includegraphics[width = 0.495\linewidth]{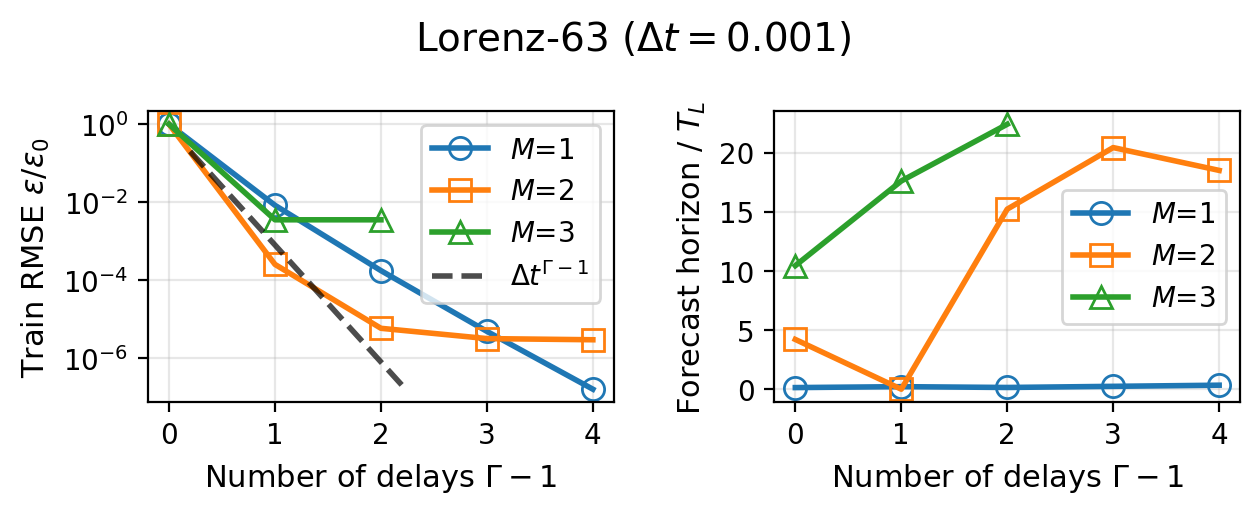}}
    \subfigure[]{\includegraphics[width = 0.495\linewidth]{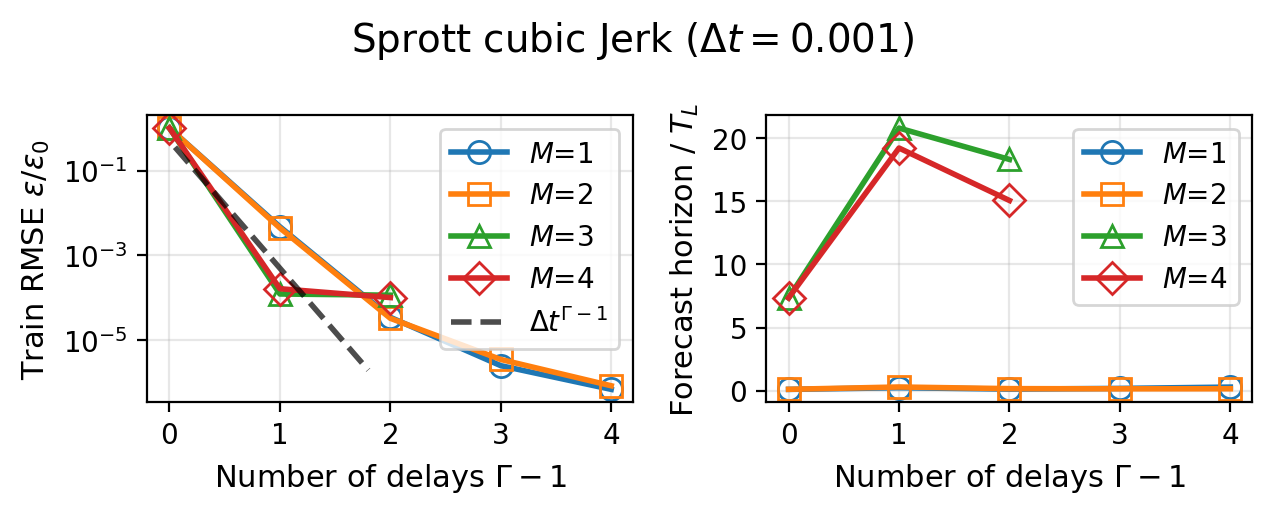}}
    \subfigure[]{\includegraphics[width = 0.495\linewidth]{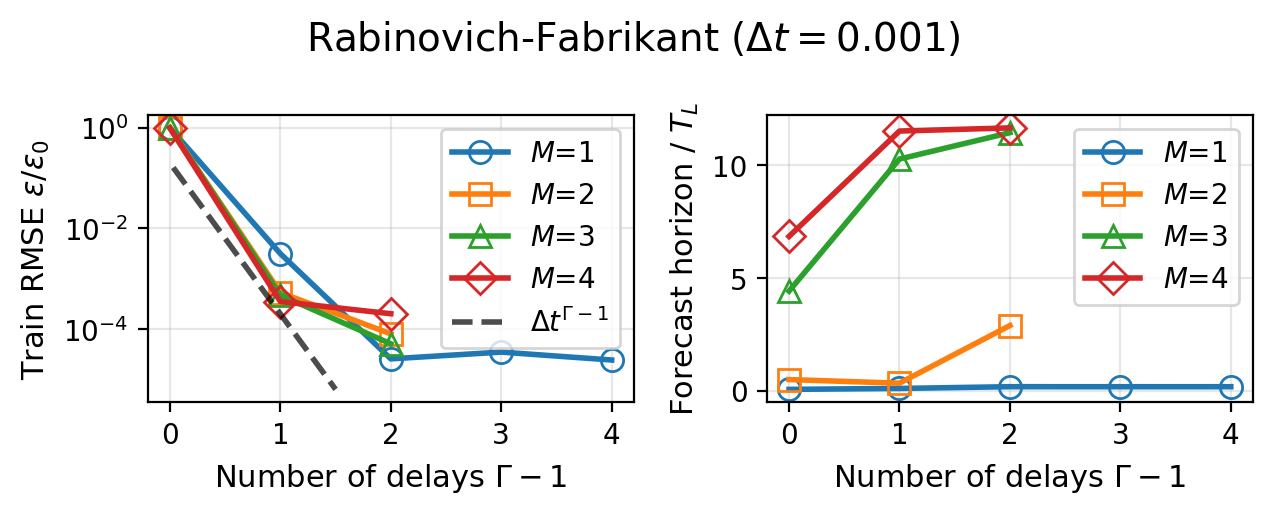}} 
    \caption{Dependence of the training error $\varepsilon/\varepsilon_0$ and the forecast horizon $T\st{fch}$ (in units of the Lyapunov time $T_L$) on the number of time taps $\Gamma$ and the monomial degree $M$ of the feature map for various dynamical systems. The theoretical prediction \eqref{eq_delay_trainerr_asy_maxform} (dashed line) gives an asymptotic error scaling for infinite resolution $\Dt\to 0$, neglecting $\Gamma$-dependent prefactors. In our convention, $\Gamma=1$ corresponds to purely instantaneous features and $\varepsilon_0$ denotes the corresponding training error. Adding delay taps always improves the training error, but systematically enhances the forecast horizon only when the order of nonlinearities $M\geq p$ is at least the maximum degree $p$ of the vector field of the dynamical system ($p=2$ for Halvorsen and Lorenz-63, $p=3$ for Sprott cubic jerk and Rabinovich-Fabrikant).} 
    \label{fig_trainerr_delay}
\end{figure}

\begin{figure}[t]
    \centering
    \subfigure[]{\includegraphics[width = 0.49\linewidth]{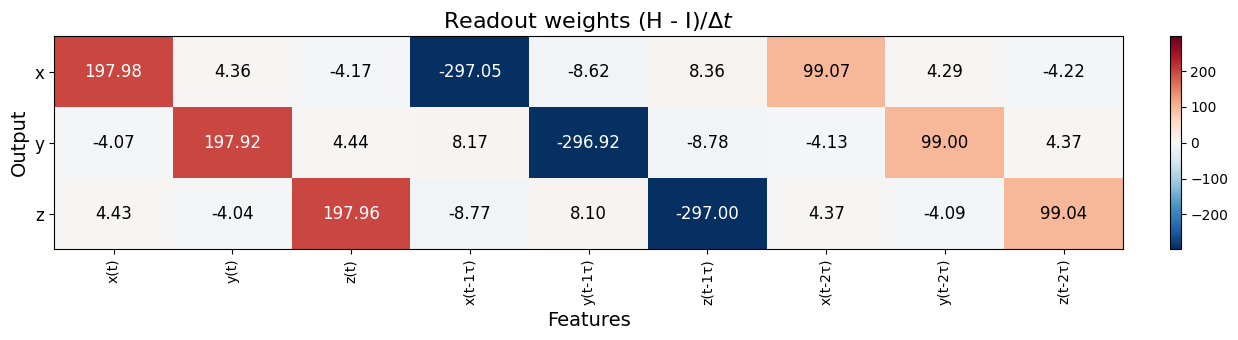}}\quad 
    \subfigure[]{\includegraphics[width = 0.48\linewidth]{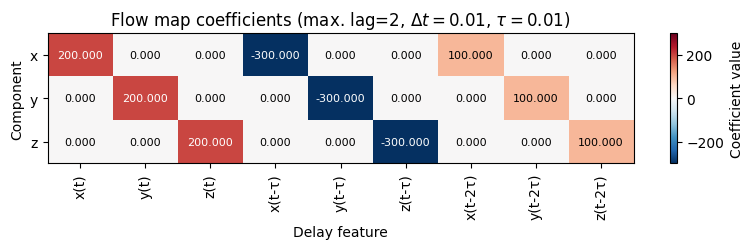}}\\
    \subfigure[]{\includegraphics[width = 0.48\linewidth]{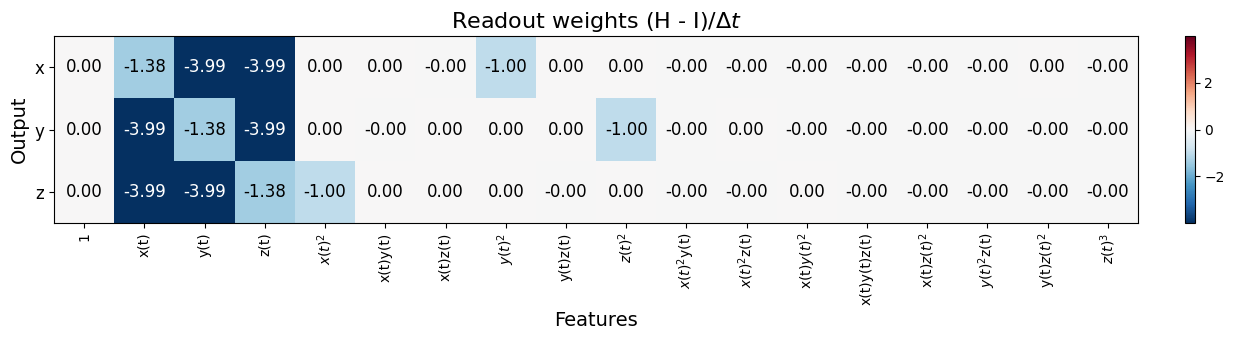}}\quad 
    \subfigure[]{\includegraphics[width = 0.49\linewidth]{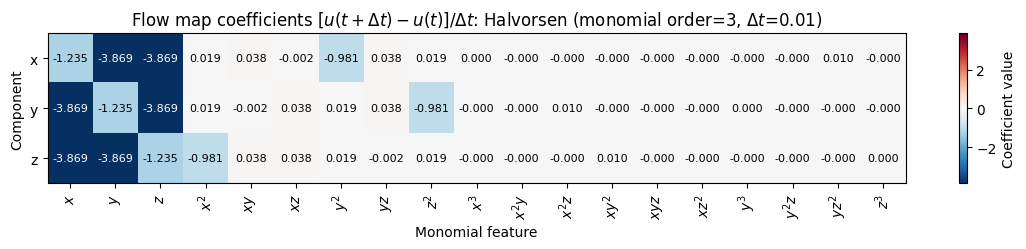}}\\
    \subfigure[]{\includegraphics[width = 0.48\linewidth]{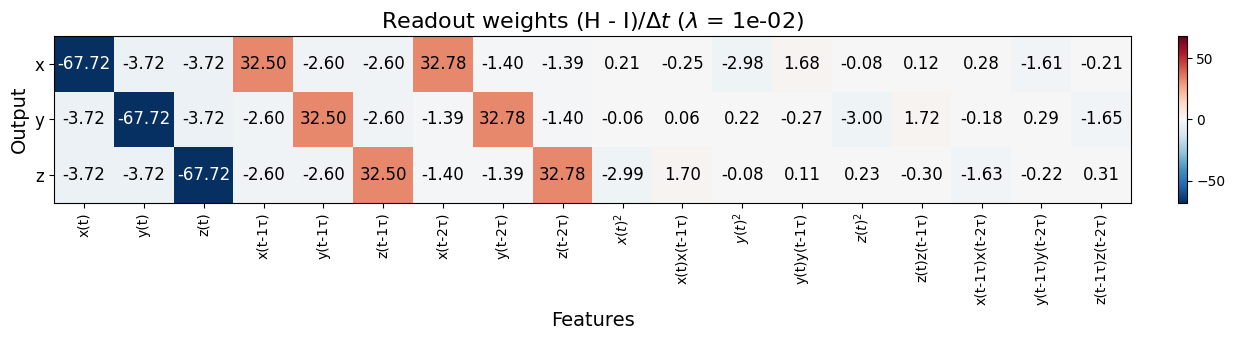}}\quad
    \subfigure[]{\includegraphics[width = 0.49\linewidth]{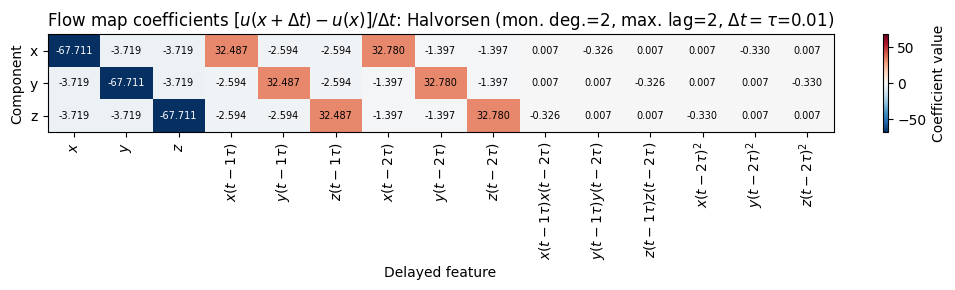}}
    \caption{Comparison between readout weights (a,c,e) of a NG-RC model trained on the Halvorsen system [\cref{eq_Halvorsen}], and the theoretically expected flow map coefficients (b,d,f). When using only linear delays (a), the weights are close to the generic finite-difference stencil \eqref{eq_delay_extrap_predictor} (b). For a feature library consisting of only instantaneous monomials (c), the polynomial flow map approximation [\cref{eq_flowmap}] is learned (up to discretization error, (d)). When using both instantaneous and delay features, the weights learned by the regressor (e) follow closely the theoretical prediction of \cref{eq_delay_flow_representation} (f). The regularization parameter is $\lambda\approx 0$ in (a,c), while similar results as in (e,f) are obtained for $\lambda\in[10^{-3},10^3]$.} 
    \label{fig_readout_delay}
\end{figure}

\subsubsection{Effect of monomial degree}

\Cref{fig_ngrc_monomial_eff} illustrates the training error and forecast horizon (obtained for the Halvorsen model) in dependence of the monomial degree $M$ of the feature library and the Ridge regularization $\lambda$.
For raw monomial libraries, the RMS training error behaves approximately as $\varepsilon\sim a (\Dt)^{b M}+ \varepsilon\st{floor}(M,\lambda)$, with constants $a,b$. The first term is the Taylor-remainder term, which monotonically shrinks as $M$ increases [see \cref{eq_trainerr_poly}]. The second term can dominate for ill-conditioned features, which arises when features have strong disparities in their magnitude or are nearly collinear.
By contrast, for whitened features, the term $\varepsilon\st{floor}(M,\lambda)$ vanishes as demonstrated by the bottom row plotted in \cref{fig_ngrc_monomial_eff}(a). Hence, in this case it is advantageous to train without Ridge regularization ($\lambda=0$).
Including delays can improve the training error to a certain extent at fixed monomial order, as shown in \cref{fig_ngrc_monomial_eff}(b). We will return to a more detailed analysis of delays below.

\subsubsection{Effect of temporal resolution}

\Cref{fig_ngrc_monomial_dt_hal}(a,b) shows in more detail the behavior of the training error, following closely \cref{eq_trainerr_poly} until a floor is reached. 
The forecast horizon $T\st{fch}$ grows according to \cref{eq_forec_hor} with decreasing $\varepsilon$.
The saturation of the training error and forecast horizon is due to the integrator accuracy used to generate the trajectories and can be further increased by using a higher floating point precision \cite{schotz_machine_2025} \footnote{We used the \texttt{solve\_ivp} function from the Python \texttt{scipy} library with accuracy $10^{-13}$.}.

\Cref{fig_ngrc_monomial_dt_hal}(c) demonstrates that the scaling behavior with $\Dt$ of the truncation error $\varepsilon\st{trunc}=\varepsilon/\Dt$ [\cref{eq_ngrc_truncerr}] can distinguish a misspecified model ($\varepsilon\st{trunc}\sim\const$) from a model that is able to learn the vector field ($\varepsilon\st{trunc}\sim\Dt^{r^\ast}$ with a $r^\ast>0$).
The former case corresponds to $M<p$, where $p$ is the nonlinear degree of the vector field.
The latter case corresponds to $M\geq p$ and leads to a consistent autoregressive evolution with a nonzero $T\st{fch}$.

Notably, this characteristic scaling behavior also occurs for an effective $\Dt = k\Dt_0$ (with $k\in\mathbb{N}$) obtained via subsampling the time series of a higher temporal resolution $\Dt_0$ (see \cref{fig_ngrc_monomial_dt_hal}(d)).
This is relevant for practical situations, where one may not be able to change the temporal resolution $\Dt$ of the data.

In \cref{fig_ngrc_monomial_dt_other}, we repeat the same analysis for the Lorenz-63 model, the Sprott cubic jerk, and the Rabinovich-Fabrikant system [\cref{eq_sprott_sys,eq_RabFab_sys}]. The latter two systems have cubic nonlinearities, such that the truncation error vanishes with $\Dt$ only for feature maps $M\geq 3$, as confirmed by the numerics (panels c--f). 

\subsubsection{Semigroup consistency}
The semigroup consistency of a learned flow map $\widehat\Phi_\Dt$ requires 
\beq \widehat\Phi_{\ell\Dt}\approx \widehat\Phi_{\Dt}^\ell
\label{eq_semigroup_flow}\eeq
where $\widehat\Phi_{\Dt}^\ell = \widehat\Phi_{\Dt} \circ \widehat\Phi_{\Dt} \circ \cdots \circ \widehat\Phi_{\Dt}$ ($\ell$ times).
When the model can reproduce the early Lie derivatives of the true flow map, the error in the approximation in \cref{eq_semigroup_flow} is proportional to the training error [see \cref{eq:semigroup_training_relation}], 
\beq \Scal_m:= \|\widehat\Phi_{\ell\Dt}-\widehat \Phi_{\Dt}^\ell\|\sim \varepsilon\st{train}.
\label{eq_semigroup_consistency}\eeq 
which is illustrated in \cref{fig_ngrc_semigroup}.
Note that, in general, the semigroup consistency only ensures that the flow map generates autonomous dynamics, not that the model has learned the ground truth.

\subsubsection{Effect of delay terms}

\Cref{fig_ngrc_delay_eff} demonstrates the influence of the number of time taps $\Gamma$ on training error and forecast horizon at fixed monomial degree $M$ (Halvorsen model with $\Dt=0.01$).
When using only linear delays ($M=1$), the NG-RC model becomes a linear autoregressive process and learns a harmonic approximation, often with a slowly decaying amplitude envelope (see \cref{fig_ngrc_harmonic_learn}) \cite{so_linear_2005,vaseghi_advanced_2008,pan_structure_2020}. 
While arbitrarily low training errors can be achieved up to noise and conditioning effects [see \cref{eq_delay_trainerr_asy_maxform}], the model is unable to produce accurate forecasts for nonlinear systems. 

As shown in \cref{fig_trainerr_delay} in more detail, the empirical training error follows the theoretical scaling prediction \eqref{eq_delay_trainerr_asy_maxform} only for small delay numbers $\Gamma$ and sufficiently large monomial orders $M$. For larger $\Gamma$, the training error eventually reaches the noise floor controlled by accuracy of the generated trajectories. (This happens earlier for large $M$ since training error $\varepsilon_0$ without delays is already small.) For small monomial orders $M$, the observed scaling of $\varepsilon$ with $\Gamma$ differs significantly from the asymptotic prediction \eqref{eq_delay_trainerr_asy_maxform}. Besides a possible $\Gamma$-dependence of prefactors, we attribute this discrepancy to the finite temporal resolution. 

\Cref{fig_trainerr_delay} shows moreover that adding delay terms does not automatically improve the forecast horizon: this in general requires the model to learn some finite-$\Delta t$ approximation of the flow map, which is only possible when the monomial degree $M$ is at least as large as the maximum degree $p$ of the vector field of the system ($p=2$ for the Halvorsen and Lorenz-63 systems, and $p=3$ for the Sprott and Rabinovich-Fabrikant systems). 
However, delay terms, especially when combined with low-order nonlinearities, can support a harmonic approximation on top of a rough flow map representation. This can result in nonzero forecast horizons even for $M<p$, as observed for the Rabinovich-Fabrikant system in \cref{fig_trainerr_delay}(d).

\subsubsection{Readout weights}

In \cref{fig_readout_delay}, we compare the learned readout weights of a NG-RC trained on the Halvorsen system to the theoretical predictions in \cref{sec_delay_markov}. A feature map consisting only of linear delay terms results in the generic weights of \cref{eq_delay_extrap_predictor} (panels a,b). Conversely, if the feature map consists only of instantaneous monomials (panels c,d), the true flow map \eqref{eq_flowmap} is recovered to that order (see also \cite{chen_next_2022,zhang_catch22s_2023,zhang_how_2025}). 
A purely Markovian flow map can also be approximated by a linear combination of instantaneous and delay features. In this case, the learned weights (panel e) are approximated by the expression $H A^+$ [\cref{eq_delay_flow_representation}, panel f].

\subsection{Fourier feature library}

\begin{figure}[tb]
    \centering
    (a)\includegraphics[width = 0.55\linewidth]{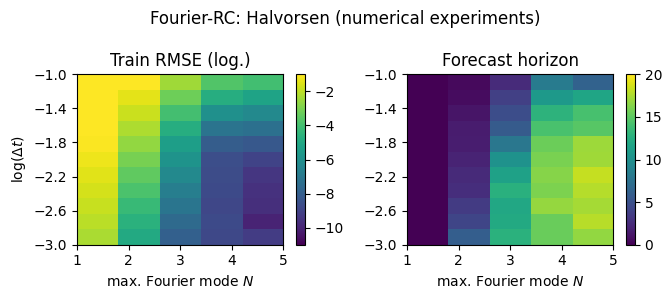}\qquad
    \includegraphics[width = 0.32\linewidth]{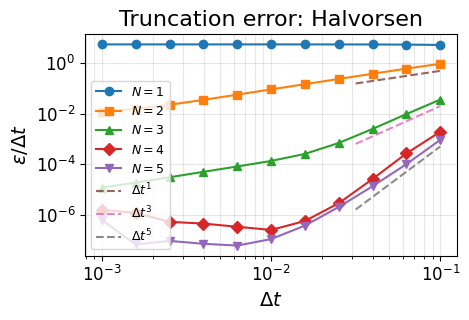}
    (b)\includegraphics[width = 0.55\linewidth]{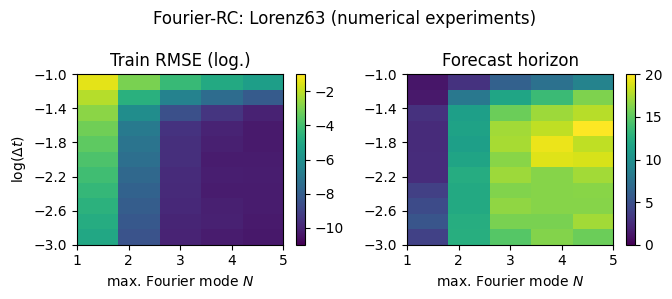}\qquad
    \includegraphics[width = 0.32\linewidth]{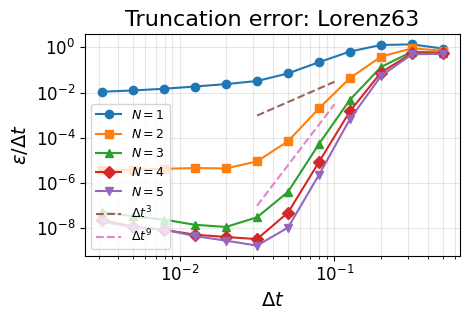}
    (c)\includegraphics[width = 0.55\linewidth]{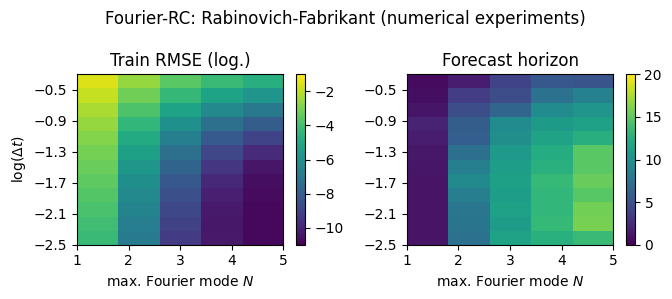}\qquad
    \includegraphics[width = 0.32\linewidth]{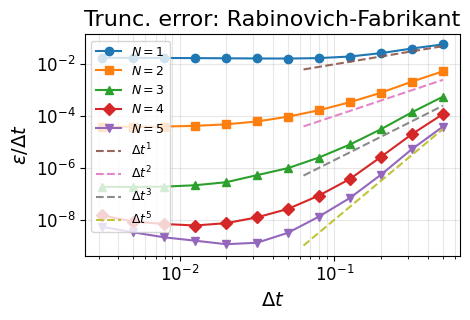}
    (d)\includegraphics[width = 0.55\linewidth]{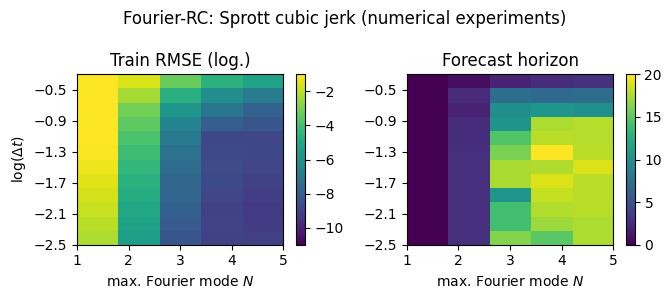}\qquad
    \includegraphics[width = 0.32\linewidth]{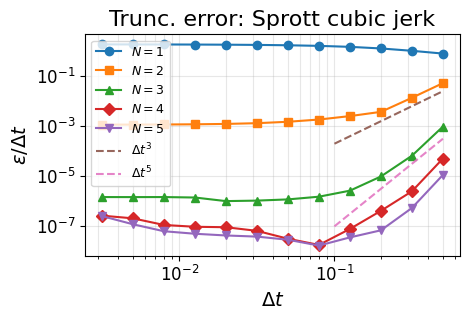}
    \caption{NG-RC with a Fourier feature library trained on various polynomial dynamical systems. The left panels show the dependence of the training error and forecast horizon (in units of the Lyapunov time) on the time resolution $\Dt$ of the training data and of the number of Fourier modes $N$ in the feature library. The right panels show the truncation error $\varepsilon/\Dt$ vs.\ $\Dt$ for various $N$. The base frequency $\omega_0$ is set to $0.01$ in (a,b), $0.3$ in (c), and $0.05$ in (d). The Ridge regularization parameter is generally set to $\lambda=0$, which is found to be optimal for all considered systems.} 
    \label{fig_ngrc_four_dt_kmax}
\end{figure}

\begin{figure}[tb]
    \centering
    \includegraphics[width = 0.75\linewidth]{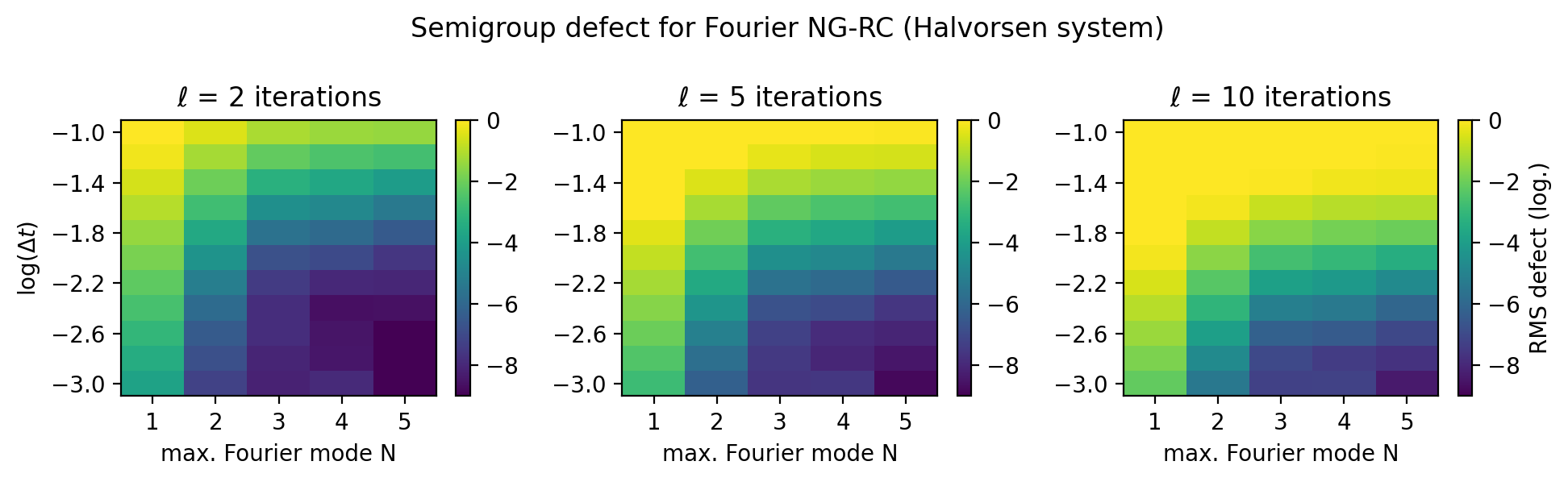}
    \caption{Semigroup consistency test for a NG-RC with Fourier-features. The semigroup defect $\Scal_\ell$ [\cref{eq_semigroup_consistency}] is shown for various iteration lengths $\ell$ as function of temporal resolution $\Dt$ of the trajectories and number of Fourier modes $N$ of the feature map trained on the Halvorsen system. Except for small $\Dt$, the Fourier features can approximate the early Lie derivatives with small error, hence the semigroup defect closely follows the training error, see \cref{fig_ngrc_four_dt_kmax}(a).}
    \label{fig_ngrc_semigroup_four}
\end{figure}

We now consider a NG-RC model with Fourier features [\cref{eq_Fourier_library}] as an example of reconstructing a truncated flow map with a non-polynomial feature library.
As illustrated in \cref{fig_ngrc_four_dt_kmax}, while the overall prediction accuracy is comparable to the polynomial case (depending slightly on the considered dynamical system), the most distinct difference is the presence of a plateau of the truncation error $\varepsilon/\Dt$ for small $\Dt$.
This is due to the functional mismatch between the Fourier features and the low-order Lie coefficients generated by a polynomial vector field [see \cref{eq_Etrunc_norep}]. 
Since these coefficients are of polynomial form, a finite Fourier feature library can only approximate them, generically leading to nonzero first defect [see \cref{eq:defect_profile} for details].

Actual power law behavior of the truncation error can appear only in a pre-asymptotic regime.
This is approximately described by \cref{eq_Etrunc_preasymp} with an empirical exponent $r_*\propto N$ for small $N$, which is consistent with the conservative lower bound in \cref{eq_Fourier_scal_exp}.
For larger $N\gtrsim 4$, we find the exponent to saturate.
Due to the highly nonlinear nature of the Fourier feature library, the truncation error does not show a clear threshold behavior as in the polynomial case that would indicate exact flow map representation.

\begin{figure}[tb]
    \centering
    \subfigure[]{\includegraphics[width = 0.52\linewidth]{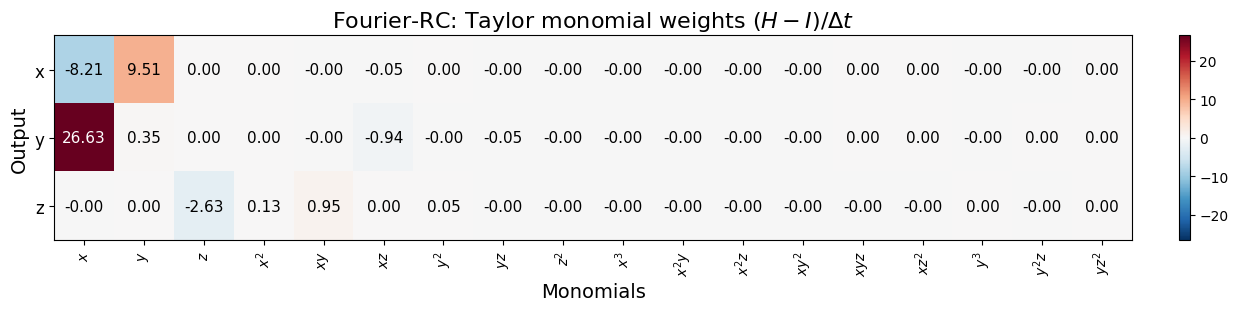}}\quad 
    \subfigure[]{\includegraphics[width = 0.45\linewidth]{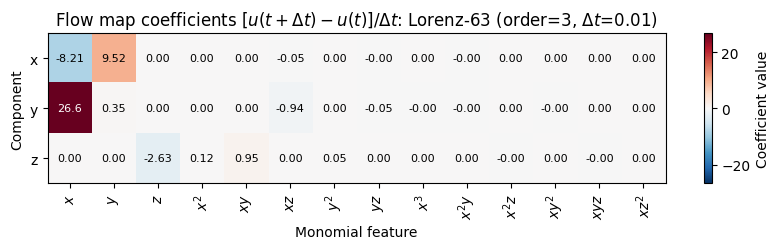}}\\
    \subfigure[]{\includegraphics[width = 0.50\linewidth]{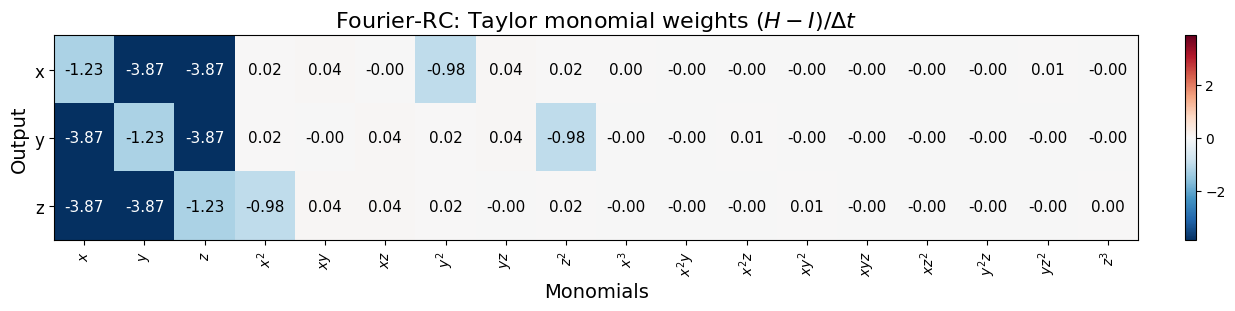}}
    \subfigure[]{\includegraphics[width = 0.49\linewidth]{flowmap_Halvor_inst3.png}}\\
    \caption{Comparison of the learned effective flow map of a NG-RC with Fourier features (a,c) with the true flow map of the Lorenz-63 system (b) and the Halvorsen system (d). In (a,c), the Fourier features are Taylor-expanded up to order 3 to obtain a polynomial form. We use Fourier modes up to $N=3$, a base frequency $\omega_0=0.01$, and no Ridge regularization ($\lambda=0$).} 
    \label{fig_four_readout}
\end{figure}

\Cref{fig_ngrc_semigroup_four} illustrates the semigroup defect for the learned Fourier-feature flow map. Although the Fourier features cannot exactly reproduce the Lie derivatives, the deviations are still small enough for the semigroup defect to closely follow the training error.
The learned representation of the flow map can be understood by Taylor expanding the Fourier feature map, thereby bringing it into a polynomial form. 
\Cref{fig_four_readout} reveals close agreement with the theoretical expectations, demonstrating that the model linearly combines the Fourier features in order to represent the monomial terms occurring in the true flow map.



\section{Summary}
\label{sec_summary}

We analyzed here NVAR/NG-RC models trained on time series generated by Markovian nonlinear dynamical systems and showed that a structural mismatch between the vector field and the feature map results in characteristic scaling behaviors of the (RMS) training error $\varepsilon\st{train}\sim\Dt^{1+r^\ast}$ with temporal resolution $\Dt$.
The exact asymptotic exponent $r^\ast$ is determined by the first order in the Taylor/Lie expansion of the flow map that can not be represented exactly within the feature space. 
Libraries with exact reproduction result in a non-zero asymptotic exponent $r^\ast$ because they annihilate the first few orders of the Lie derivatives exactly. 
By contrast, libraries that can merely approximate the Lie coefficients generally lead to a ``trivial'' asymptotic law $\varepsilon\st{train}\sim \Dt$ ($r^\ast=0$) for $\Dt\to 0$.
However, they may exhibit steeper pre-asymptotic scaling when the low-order Lie coefficients are approximated with a sufficiently small error, resulting in an effective exponent $r_*>0$ at larger time resolutions $\Dt$.

In order to illustrate this generic mechanism, we focused on dynamical systems with polynomial vector fields. 
Since the Lie coefficients are themselves polynomials in this case, a NVAR/NG-RC equipped with a delay-free monomial library can exactly reproduce the flow map to a given order in $\Dt$.
Accordingly, this is reflected by a superlinear scaling of the RMS training error, $\varepsilon\st{train}\sim\Dt^{1+r^\ast}$ with an $r^\ast>0$ that directly depends on the nonlinear order of the vector field [see \cref{eq_trainerr_poly} and \cref{fig_ngrc_monomial_dt_hal,fig_ngrc_monomial_dt_other}].
By contrast, a finite Fourier feature library can only approximately reproduce the polynomial Lie coefficients, leading to an asymptotic plateau of the truncation error (see \cref{fig_ngrc_four_dt_kmax}).
Note that an investigation of the forecast horizon alone would not immediately reveal these qualitative differences of the feature maps: although the forecast horizon is primarily controlled by the training error (see \cref{app_train_err}), it can be additionally affected by model instabilities and the intrinsic Lyapunov divergence.

Under Markovian dynamics, delay features are deterministic functions of the current state and can thus in principle be represented as linear combinations of instantaneous features [see \cref{eq_delay_flow_representation}].
They are thus not crucial for learning Markovian systems, but still generally reduce the one-step training error, since they allow the regressor to form better interpolants [see, e.g., \cite{zhang_how_2025} and \cref{eq_delay_trainerr_asy}].
However, they do not generically lead to better generalization behavior, i.e., longer forecast horizons, unless also sufficiently high-order nonlinearities are present in the feature map (see \cref{fig_ngrc_delay_eff,fig_trainerr_delay}).
This constitutes an example of a model failing to generalize despite having a small training error.
Delay terms are essential for learning non-Markovian systems, which will be interesting to analyze in future work.

In summary, one can identify three distinct learning regimes of an NVAR/NG-RC model when trained on time series of nonlinear dynamical systems: (i) a linear delay model learns a harmonic approximation (see \cref{app_delay_extrap}), which can render a small training error but inconsistent forecasts \cite{so_linear_2005,vaseghi_advanced_2008,pan_structure_2020}; (ii) a nonlinear model with a structural mismatch to the underlying vector field can achieve long forecast horizons, but the RMS training error will scale suboptimally (linearly) as $\Dt\to 0$; (iii) a nonlinear model that structurally matches the vector field achieves long forecast horizons and an optimal (superlinear) scaling of the RMS training error with $\Dt$. 
In situations where the underlying data-generating process is unknown, the truncation error scaling can be used via subsampling to assess the structural properties of the underlying system and to guide the design of the feature library. It will be useful to analyze this further in the future.
The semigroup property of the flow map is a standard diagnostic for flow map learning \cite{brunton_modern_2022,chen_deeposg_2023,zhang_quantitative_2024}. For a well-specified model, the semigroup defect closely follows the training error (see \cref{fig_ngrc_semigroup,fig_ngrc_semigroup_four}).
We finally remark that the present results can also help to construct optimal encodings in quantum machine learning models \cite{schuld_machine_2021}, including quantum RCs \cite{fujii_harnessing_2017}, for which the feature library can often be explicitly stated \cite{gross_theory_2026}. In particular, common quantum encodings lead to a finite Fourier feature library \cite{schuld_effect_2021}.

\acknowledgments
This project was made possible by the DLR Quantum Computing Initiative and the Federal Ministry for Research, Technology and Space; \url{https://qci.dlr.de/NeMoQC}.


\appendix


\begin{table}[h]
\centering
\renewcommand{\arraystretch}{1.2}
\begin{tabular}{lll}
\hline\hline
Symbol & Description & Definition \\
\hline
$\Omega \subset \reals^d$ & Domain of the dynamical system & \eqref{eq_dynsys_ODE} \\
$\xv(t) \in \Omega$ & State vector or observable at time $t$ & \eqref{eq_dynsys_ODE} \\
$\fv(\xv)$ & Vector field of the continuous dynamical system ($\dot\xv = \fv(\xv)$) & \eqref{eq_dynsys_ODE} \\
$\Phi_{\Delta t}$ & Discrete-time flow map ($\xv(t+\Delta t) = \Phi_{\Delta t}(\xv(t))$) & \eqref{eq_flowmap} \\
$\Delta t$ & Sampling time step & \eqref{eq_flowmap} \\
$\Fv_q(\xv)$ & $q$-th Lie-series coefficient, $\Fv_1=\fv$, $\Fv_{q+1}=L_\fv \Fv_q$ & \eqref{eq_flowmap} \\
$\psiv(t) \in \reals^n$ & Feature vector (reservoir state) at time $t$ & \eqref{eq_nextstep_task} \\
$\Psiv \in \reals^{n\times P}$ & Design matrix & \eqref{eq_nvar_optprob} \\
$\gv(\cdot), \Gv(\cdot)$ & Feature maps ($\gv$ instantaneous; $\Gv$ possibly delay-augmented) & \eqref{eq_NGRC}, \eqref{eq_delay_pred} \\
$K\in \reals^{n\times n}$ & Koopman operator approximation & \eqref{eq_nextstep_task} \\
$H \in \reals^{d \times n}$ & Next-step prediction operator & \eqref{eq_ngrc_reduct} \\
$P$ & Number of training samples & \eqref{eq_nvar_optprob} \\
$\Gamma$ & Number of time taps in the delay vector & \eqref{eq_delay_pred} \\
$\tau$ & Time lag in delay embedding & \eqref{eq_delay_pred} \\
$M$ & Maximum degree of instantaneous polynomial features in $\gv$ & \eqref{eq_NGRC} \\
$M'$ & Polynomial degree of the truncated flow map & \eqref{eq_flowmap_polydeg} \\
$N$ & Number of (non-zero) Fourier modes (per dim.) in $\gv$ & \eqref{eq_Fourier_library} \\
$\omega_0$ & Base frequency & \eqref{eq_Fourier_library} \\
$E\st{train}$ & Mean-squared training error & \eqref{eq_ngrc_trainerr} \\
$\varepsilon\st{train}$ & RMS training error & \eqref{eq_ngrc_trainerr} \\
$T\st{fch}$ & Forecast horizon & \eqref{eq_fc_hor} \\
$\lambda$ & Tikhonov regularization parameter & \eqref{eq_nvar_optsol} \\
$\lambda_1$ & Largest Lyapunov exponent & \\
\hline\hline
\end{tabular}
\caption{Summary of notation used in this paper. We use $\hat A$ to specifically indicate a fitted operator $A$, and $A^+$ for the pseudo-inverse.}
\label{tab:notation}
\end{table}

\clearpage


\section{Feature-library structure and asymptotic error scaling}
\label{sec_feature_library_structure}

As in \cref{sec_markov_flow}, we consider a dynamical system
\beq
\dot\xv(t) = \fv(\xv(t)), \qquad \xv(t) \in \reals^d,
\eeq
with flow map $\Phi_\Dt$.
We study predicting the increment
\beq
\Fv_\Dt(\xv) := \Phi_\Dt(\xv)-\xv
\eeq
from a prescribed feature library $\gv(\xv) = (g_1(\xv),\dots,g_n(\xv)) \in \reals^n$ by a linear readout. 
A central question is how the structure of the feature library influences the small-$\Dt$ behavior of the mean-square error and gives rise to possible pre-asymptotic scaling laws.
The results apply to the full state prediction problem in the main text since we assume that $\xv$ is included in $\gv$ if necessary.

We assume that the data are distributed according to a probability measure $\mu$ on a domain $\Omega \subset \reals^d$, and we identify the idealized mean-square training error with the projection error in $L^2(\mu;\reals^d)$. This setting captures the regime in which the empirical training error is dominated by approximation rather than by finite-sample, optimization, or numerical effects.

\subsection{A library-agnostic projection framework}
\label{sec_library_projection}

Given the feature map $\gv$, the associated model class consists of all vector-valued functions that can be represented by a linear readout,
\beq
V := \left\{ H \gv(\cdot) : H \in \reals^{d \times n} \right\} \subset L^2(\mu;\reals^d).
\eeq
Here $H$ is only a generic coefficient matrix parameterizing the elements of $V$. For a fixed target function $\hv \in L^2(\mu;\reals^d)$, the projection $\Pcal \hv$ is the best approximation to $\hv$ among all functions in $V$. We denote by $H_{\hv}$ a coefficient matrix that represents this projected function,
\beq
(\Pcal \hv)(\xv) = H_{\hv}\gv(\xv).
\eeq
This target-dependent matrix is determined by the normal equations for the least-squares projection: defining the population feature covariance and cross-covariance
\beq
G := \int_\Omega \gv(\xv)\gv(\xv)^T \, d\mu(\xv),
\qquad
C_{\hv} := \int_\Omega \hv(\xv)\gv(\xv)^T \, d\mu(\xv),
\eeq
one obtains
\beq
H_{\hv} G = C_{\hv},
\eeq
so that $H_{\hv}=C_{\hv}G^{-1}$ when $G$ is invertible. If $G$ is singular, the representation of the projected function $\Pcal\hv$ by coefficients need not be unique. In this case one may either remove redundant features or choose the Moore--Penrose solution $H_{\hv}=C_{\hv}G^+$. Equivalently, the projector is obtained by Gram--Schmidt orthonormalization of $g_1,\dots,g_n$ in $L^2(\mu)$. 

Let $\Qcal := I-\Pcal$, so that $\Qcal\hv$ is the component of $\hv$ orthogonal to $V$, i.e., the part of $\hv$ that is not exactly representable by the feature library.
The population mean-square error for increment prediction is then given by
\beq
\mathcal{E}(\Dt)=\inf_{\hv \in V} \|\Fv_\Dt-\hv\|_{L^2(\mu)}^2
=\|\Qcal \Fv_\Dt\|_{L^2(\mu)}^2.
\label{eq:population_projection_error}
\eeq
This expression depends only on the span of $V$. Therefore any two feature libraries with the same span, including monomial and orthogonal polynomial bases of the same polynomial subspace, have identical projection error for every $\Dt$ and thus identical asymptotic scaling.

The increment map admits the usual Taylor/Lie-series expansion, using the same Lie-series coefficients $\Fv_m$ as in \cref{eq_flowmap}:
\beq
\Fv_\Dt(\xv) =\sum_{m=1}^\infty \frac{\Dt^m}{m!} \Fv_m(\xv),
\qquad
\Fv_1 = \fv,
\qquad
\Fv_{m+1} = L_\fv \Fv_m,
\label{eq:Lie_series_increment}
\eeq
where $L_\fv \hv := \fv \cdot \nabla \hv$ is the Lie derivative along the vector field. Defining the $m$-th library \emph{defect} by
\beq
\qv_m := \Qcal \Fv_m,
\qquad
\varepsilon_m := \|\qv_m\|_{L^2(\mu)},
\label{eq:defect_profile}
\eeq
we obtain the asymptotic expansion
\beq
\mathcal{E}(\Dt) = \left\| \sum_{m=1}^\infty \frac{\Dt^m}{m!} \qv_m \right\|_{L^2(\mu)}^2 
= \sum_{m,n \ge 1} \frac{\Dt^{m+n}}{m! n!}\langle \qv_m, \qv_n \rangle_{L^2(\mu)}.
\label{eq:exact_error_expansion}
\eeq
It is useful to introduce $r^\ast$ as the largest order to which the library represents the successive Lie derivatives exactly, i.e.,
\beq
r^\ast := \max \left\{ r \ge 0 : \Fv_1,\dots,\Fv_r \in V \right\},
\eeq
with the convention $r^\ast=0$ if already $\Fv_1=\fv \notin V$. 
The first non-vanishing defect order is then 
\beq
m_0 := \min \{ m \ge 1 : \varepsilon_m > 0 \} = r^\ast+1,
\label{eq:defect_order_nonv}\eeq
where the last equality holds provided $r^\ast<+\infty$.
This defines the leading term of the error as
\beq
\mathcal{E}(\Dt)=\frac{\varepsilon_{m_0}^2}{(m_0!)^2} \Dt^{2m_0}+\Ocal\!\left(\Dt^{2m_0+1}\right),
\label{eq:true_asymptotic_scaling}
\eeq
implying the exact asymptotic exponent $2m_0$. The asymptotic exponent for the (mean-square) truncation error [\cref{eq_ngrc_truncerr}] is, correspondingly, $2r^\ast = 2(m_0-1)$.
This shows that the small-$\Dt$ behavior of the error is determined by the structural alignment between the feature map and the successive Lie derivatives $\fv, L_\fv \fv, L_\fv^2 \fv,\dots$.

\subsection{Asymptotic versus pre-asymptotic regimes}

If the vector field is not exactly representable by the feature library, $\fv \notin V$, then $\varepsilon_1 > 0$ and the true asymptotic law is necessarily
\beq
\mathcal{E}(\Dt) = \varepsilon_1^2 \Dt^2 + \Ocal(\Dt^3),
\label{eq:asymptotic_scaling_norep}\eeq
If $\varepsilon_1$ is much smaller than $\varepsilon_m$ for some $m>1$, different pre-asymptotic scalings are possible.
Then, over a window of $\Dt$ in which the $m$-th term dominates in \eqref{eq:exact_error_expansion}, the error behaves effectively as
\beq
\mathcal{E}(\Dt)\approx \frac{\varepsilon_m^2}{(m!)^2} \Dt^{2m}.
\label{eq:preasymptotic_scaling_generic}
\eeq
Accordingly, one should distinguish between the exact asymptotic exponent $2m_0$ [\cref{eq:true_asymptotic_scaling}], which is determined by the first nonzero defect, and the effective pre-asymptotic exponent $2 m$.
This is the main difference between libraries that reproduce the relevant Lie derivatives exactly and libraries that only approximate them well.

The crossover from an intermediate $\Dt^{2m}$ regime back to the asymptotic $\Dt^2$ regime is determined by a balance of defect amplitudes. If $m>1$ is the dominant pre-asymptotic order, then the crossover scale is obtained by equating the first- and $m$-th defect contributions,
$\varepsilon_1 \Dt\sim (\sfrac{\varepsilon_m}{m!}) \Dt^m$, which gives
\beq
\Dt_{\times}^{(1,m)}
\asymp
\left(
\frac{m! \, \varepsilon_1}{\varepsilon_m}
\right)^{1/(m-1)}.
\label{eq:crossover_scale}
\eeq

\subsection{Polynomial vector fields}

In order to make the previous discussion more concrete, consider the case when $\fv$ is polynomial of total degree $p$. In that case each Lie coefficient $\Fv_m = L_\fv^{m-1} \fv$ is again polynomial, with the degree bound
\beq
\mathrm{deg}\,\Fv_m \le 1 + m(p-1),
\qquad p>1.
\label{eq:degree_growth}
\eeq
Hence the question reduces to how well the library reproduces polynomials of the degrees that are produced by the successive Lie derivatives.

\subsubsection{Libraries with exact polynomial reproduction}

Let $\Pi_M^d$ denote the space of $\reals^d$-valued polynomials of total degree at most $M$. 
We define the polynomial reproduction order of the library as the largest integer $M_\ast$ such that
\beq
\Pi_{M_\ast}^d \subset V.
\eeq
Any library whose span $V$ contains $\Pi_{M_\ast}^d$ thus yields the same asymptotic scaling for polynomial dynamics, since all Lie coefficients satisfying $1+m(p-1) \le M_\ast$ are represented exactly. 
Specifically, one has $\Fv_1,\dots,\Fv_{r^\ast}\in V$ with 
\beq r^\ast = \left\lfloor \frac{M_\ast-1}{p-1} \right\rfloor.
\label{eq:poly_scaling_generic}
\eeq 
For a generic polynomial vector field of degree $p$, the first unresolved Lie coefficient thus appears at order $r^\ast+1= m_0$, which provides the asymptotic exponent in \cref{eq:true_asymptotic_scaling}.

Thus polynomial libraries change the true asymptotic exponent because they annihilate the first defects exactly. This is in sharp contrast with the generic behavior of finite non-polynomial libraries, for which the same early defects are typically small but not exactly zero.

\subsubsection{Libraries without exact polynomial reproduction}

For a general non-polynomial feature library the most informative quantity is not a single degree cutoff, but rather the defect profile against polynomial test functions. 
Consider the (worst-case) error in reconstructing a degree-$D$ polynomial by the feature library,
\beq
\delta(D)
:=
\sup_{\substack{\hv \in \Pi_D^d \\ \|\hv\| = 1}}
\|\Qcal \hv\|_{L^2(\mu)}.
\label{eq:polynomial_defect_profile}
\eeq
If $\fv$ is polynomial of degree $p$, then the Lie coefficient $\Fv_m \in \Pi_{1+m(p-1)}^d$, so \eqref{eq:polynomial_defect_profile} yields
\beq
\varepsilon_m
=
\|\Qcal \Fv_m\|_{L^2(\mu)}
\le
\delta(1+m(p-1)) \, \|\Fv_m\|_{L^2(\mu)}.
\label{eq:defect_bound_via_delta}
\eeq
If $\delta(D)=0$ up to a certain degree, then the corresponding Lie coefficients are represented exactly, yielding an asymptotic exponent larger than 2. If $\delta(D)$ is merely small, then the true asymptotic exponent remains $2$ (unless the first defects vanish exactly), but pre-asymptotic windows with larger effective exponents may occur.

\subsection{Fourier feature libraries}
\label{sec_Fourier_lib}

We now consider flow map approximation with a $N$-frequency Fourier feature library:
\beq
\gv^{(N)}(\xv)=\left\{\sin\!\bigl(\omega_0 \, \kv\cdot \xv\bigr),\cos\!\bigl(\omega_0 \, \kv\cdot \xv\bigr):\kv \in \{0,\dots,N\}^d\right\},
\label{eq:Fourier_library}
\eeq
with base frequency $\omega_0$. 
Such a library can show extended pre-asymptotic error scaling before crossing over to the asymptotic law $\propto \Dt^2$.
We assume the data reside in a box $\Omega \subset [-R,R]^d$, and define
\beq
\rho := \omega_0 R.
\eeq
The relevant regime for approximating polynomial dynamics by low-frequency Fourier features is given by $\rho \ll 1$.

Monomials are generated by frequency derivatives:
\beq
\partial_{\kv}^\alpha e^{\im \omega_0 \kv\cdot \xv}
= (\im \omega_0)^{|\alpha|} \xv^\alpha e^{\im \omega_0 \kv\cdot \xv},
\qquad
\xv^\alpha = x_1^{\alpha_1} \cdots x_d^{\alpha_d},
\label{eq:monomial_from_frequency_derivative}
\eeq
which gives at $\kv=0$:
\beq
\xv^\alpha = (\im \omega_0)^{-|\alpha|} \partial_{\kv}^\alpha e^{\im \omega_0 \kv\cdot \xv}\big|_{\kv=0}.\eeq
A finite Fourier library may thus be viewed as a finite-difference approximation to these derivatives in frequency space.
More explicitly, let $a_{\alpha_j,\nu_j}$, $\nu_j=0,\dots,N$, be the one-dimensional finite-difference weights for the $\alpha_j$-th derivative at $0$ using the nodes $0,\dots,N$, so that
\beq
\sum_{\nu_j=0}^N a_{\alpha_j,\nu_j} \nu_j^m = \alpha_j! \, \delta_{m,\alpha_j},\qquad
m=0,\dots,N.
\eeq
Setting
\beq
c_{\alpha,\nu} := \prod_{j=1}^d a_{\alpha_j,\nu_j},\qquad
\nu=(\nu_1,\dots,\nu_d) \in \{0,\dots,N\}^d,
\eeq
one obtains the Fourier approximant for the monomial $\xv^\alpha$,
\beq
Q_{\alpha}^{\mathrm{Four}}(\xv):=
\frac{1}{(\im \omega_0)^{|\alpha|}} \sum_{\nu \in \{0,\dots,N\}^d} c_{\alpha,\nu} e^{\im \omega_0 \nu \cdot \xv}.
\label{eq:Fourier_monomial_approximant}
\eeq
A Taylor expansion of the exponential then yields the uniform bound
\beq
\|Q_{\alpha}^{\mathrm{Four}} - \xv^\alpha\|_{L^\infty(\Omega)}
\le
C_{\alpha,d,N} R^{|\alpha|} \rho^{\,N+1-\|\alpha\|_\infty},
\qquad \|\alpha\|_\infty \le N,
\label{eq:universal_Fourier_bound}
\eeq
for a constant $C_{\alpha,d,N}$ depending only on the indicated parameters. Thus every mixed monomial $\xv^\alpha$ whose largest coordinate exponent does not exceed $N$ is approximated with an error suppressed by a power of $\rho$. In particular, the approximation quality improves rapidly as either $N$ increases or $\omega_0 R$ decreases.

This mechanism is readily illustrated by the following low-order examples:
\beq
x_j=\frac{\sin(\omega_0 x_j)}{\omega_0}+\Ocal(\omega_0^2 x_j^3),\qquad
x_j^2=\frac{2(1-\cos(\omega_0 x_j))}{\omega_0^2}+\Ocal(\omega_0^2 x_j^4),
\label{eq:low_order_1d_fourier_examples}
\eeq
as well as,
\beq 
x_i x_j=\frac{\cos(\omega_0 x_i) + \cos(\omega_0 x_j) - 1 - \cos\!\bigl(\omega_0(x_i+x_j)\bigr)}{\omega_0^2}+\Ocal(\omega_0^2 R^4).
\label{eq:bilinear_fourier_example}
\eeq
Moreover, for even $m$ and $N \ge m/2$, there exist coefficients $a_n^{(m)}$ such that
\beq
x_j^m=\omega_0^{-m}\sum_{n=0}^N a_n^{(m)} \cos(n \omega_0 x_j)+\Ocal\!\left( R^m \rho^{\,2N+2-m} \right),
\eeq
whereas, for odd $m$ and $N \ge (m+1)/2$, there exist coefficients $b_n^{(m)}$ such that
\beq
x_j^m=\omega_0^{-m}\sum_{n=1}^N b_n^{(m)} \sin(n \omega_0 x_j)+\Ocal\!\left( R^m \rho^{\,2N+1-m} \right).
\eeq
These formulas show that pure powers may be easier to approximate than general mixed monomials.

For a polynomial vector field, the Lie coefficients $\Fv_m$ are finite sums of monomials. Let 
\beq \eta_m := \max_{\alpha \in \mathrm{supp}(\Fv_m)} \|\alpha\|_\infty
\label{eq:max_coord_exponent}
\eeq
denote the maximum coordinate exponent of the monomials in $\Fv_m$.
Then \eqref{eq:universal_Fourier_bound} implies that, whenever $\eta_m \le N$,
\beq
\varepsilon_{m} = \|\Qcal \Fv_m\|_{L^2(\mu)} \le C_{m,N} \rho^{\,N+1-\eta_m},\qquad (\eta_m \le N)
\label{eq:Fourier_defect_bound}
\eeq
for a suitable constant $C_{m,N}$. 
This shows that low-order Lie coefficients are not represented exactly, but their defects are strongly suppressed by a positive power of $\rho$. Consequently, for fixed $N$, a finite Fourier library generically satisfies $\varepsilon_{1}>0$, so the true asymptotic law remains 
\beq\mathcal{E}(\Dt) \sim \varepsilon_1^2 \Dt^2.\qquad \text{(asymptotically $\Dt \to 0$)}
\eeq

The \emph{pre-asymptotic} regime is determined by the first Lie coefficient whose degree leads an unsuppressed defect (or, at least, where \cref{eq:Fourier_defect_bound} does not guarantee suppression).
This corresponds to a non-positive power of $\rho$ in \eqref{eq:Fourier_defect_bound}, implying a conservative bound  
\beq
m_*(N) := \min \{ m \ge 1 : \eta_m > N \}.
\label{eq:Fourier_effective_mstar}
\eeq
Over the $\Dt$-range in which the defects of orders $m < m_*(N)$ remain suppressed relative to the $m_*(N)$-th defect, one thus expects
\beq
\mathcal{E}_N(\Dt) \approx A_N \Dt^{2m_*(N)},
\label{eq:Fourier_preasymptotic}
\eeq
with a constant $A_N\sim \varepsilon_{m_*(N)}^2/(m_*(N)!)^2$.
For generic degree-$p$ polynomial dynamics one has $\eta_m \le 1+m(p-1)$, which yields the conservative estimate
\beq
m_*(N) \geq 1+ r_*,\qquad r_* = \left\lfloor \frac{N-1}{p-1} \right\rfloor.
\label{eq:Fourier_preasy_exp}\eeq
Because $\varepsilon_1$ is not exactly zero, this pre-asymptotic regime must ultimately cross over to the true $\Dt^2$ asymptotics at a scale 
\beq
\Dt_{\times} \asymp \left( \frac{m_*(N)! \, \varepsilon_{1}}{\varepsilon_{m_*(N)}} \right)^{1/(m_*(N)-1)}.\label{eq:Fourier_crossover}
\eeq
Note that in practice, empirically observed values of $m_*(N)$ can be larger since bounds like \eqref{eq:universal_Fourier_bound} are not necessarily tight.

\subsection{Connection to bias--variance decomposition}
\label{app_bias_variance}

The projection error \cref{eq:population_projection_error} is the approximation-bias component of the usual bias--variance decomposition. In the increment-prediction setting of \cref{sec_library_projection}, the target is $\Fv_\Dt \in L^2(\mu;\reals^d)$ and the model class is
\beq
V = \{ H \gv(\cdot) : H \in \reals^{d \times n} \}.
\eeq
Let $\hv_\star := \Pcal \Fv_\Dt$ denote the $L^2(\mu)$-orthogonal projection of $\Fv_\Dt$ onto $V$. Then, for any learned predictor $\hat \hv \in V$,
\[
\|\Fv_\Dt-\hat \hv\|_{L^2(\mu)}^2
=
\underbrace{\|\Fv_\Dt-\hv_\star\|_{L^2(\mu)}^2}_{=\mathcal{E}(\Dt)=\text{model bias}^2}
+
\underbrace{\|\hat \hv-\hv_\star\|_{L^2(\mu)}^2}_{\text{estimation error}},
\]
because $\Fv_\Dt-\hv_\star = \Qcal \Fv_\Dt \perp V$.
Averaging over training sets decomposes the second term into the usual finite-sample estimator bias and variance.
For the noiseless, overdetermined, fixed-library large-sample regime considered here, this estimation term vanishes, leaving only the approximation bias.
The Lie-defect expansion in \cref{eq:exact_error_expansion,eq:true_asymptotic_scaling} is therefore an asymptotic expansion of this model-bias term.

\subsection{Truncated flow-map error}
\label{sec_truncated_flowmap_error}

Since the exact expression for the flow map $\Fv_\Dt$ is usually not available, consider its approximation to order $r$ 
\begin{equation}
\Fv_{\Delta t}^{[r]}(\xv):=\Phi_{\Delta t}^{[r]}(\xv)-\xv = \sum_{m=1}^{r}\frac{\Delta t^m}{m!}\Fv_m(\xv),
\end{equation}
and define the difference to the true flow map as $\Rv_{\Delta t}^{[r]}:=\Fv_{\Delta t}-\Fv_{\Delta t}^{[r]}$.
Consider now the difference between $\Fv_{\Delta t}^{[r]}$ and the approximation of the true flow map by the feature library, $\Pcal\Fv_{\Delta t} \approx \hat H \gv$: 
in the same approximation-dominated regime as in \cref{eq:population_projection_error}, one has
\begin{align}
\Dcal_{\Phi,r}(\Delta t)
&= \left\| \Pcal\Fv_{\Delta t} - \Fv_{\Delta t}^{[r]}\right\|_{L^2(\mu)}^2 \\
&= \left\|-\Qcal\Fv_{\Delta t}^{[r]}+\Pcal\Rv_{\Delta t}^{[r]} \right\|_{L^2(\mu)}^2 \\
&= \left\|\sum_{m=1}^{r}\frac{\Delta t^m}{m!}\qv_m\right\|_{L^2(\mu)}^2+\left\| \sum_{m=r+1}^{\infty} \frac{\Delta t^m}{m!}\Pcal\Fv_m \right\|_{L^2(\mu)}^2 ,
\label{eq:truncated_flowmap_decomp}
\end{align}
where the last equality uses the orthogonality of $V$ and $V^\perp$.
Thus the flow-map approximation error consists of an unrepresented part of the retained Lie coefficients and the represented part of the omitted Taylor remainder.

Let $m_0$ be the first non-vanishing defect order defined in
\cref{eq:defect_order_nonv}, and set
\begin{equation}
s_r:=\min\left\{m>r: \|\Pcal\Fv_m\|_{L^2(\mu)}>0 \right\}.
\end{equation}
Then, generically,
\begin{equation}
\Dcal_{\Phi,r}(\Delta t) \sim
\begin{cases}
\dfrac{\varepsilon_{m_0}^2}{(m_0!)^2}\Delta t^{2m_0}, & m_0\le r,\\[1.2em]
\dfrac{\|\Pcal\Fv_{s_r}\|_{L^2(\mu)}^2}{(s_r!)^2} \Delta t^{2s_r}, & m_0>r .
\end{cases}
\label{eq:truncated_flowmap_scaling}
\end{equation}
The relation to the training error [\cref{eq:population_projection_error}] follows from the triangle inequality:
\begin{equation}
\left| \Dcal_{\Phi,r}^{1/2}(\Delta t) - \mathcal E^{1/2}(\Delta t) \right| \le \|\Rv_{\Delta t}^{[r]}\|_{L^2(\mu)} =\Ocal(\Delta t^{r+1}).
\label{eq:truncated_flowmap_training_bound}
\end{equation}
Consequently, if the truncation order includes the first unresolved Lie coefficient, $r\ge m_0$, then the truncated flow-map error is asymptotically equivalent to the training error:
\begin{equation}
\Dcal_{\Phi,r}(\Delta t) \simeq \mathcal E(\Delta t) .
\end{equation}
If $r<m_0$, however, the error is controlled by the Taylor terms omitted from the reference map.

\subsection{Semigroup consistency}
\label{sec_semigroup_consistency}

The exact flow map satisfies the semigroup property \cite{brunton_modern_2022}
\beq
\Phi_{s+t}=\Phi_s\circ \Phi_t .
\eeq
This can be tested for the learned flow map $\widehat{\Phi}_h$, and we show that it essentially follows the same $\Dt$-scaling as the training error.
In the population limit, 
\beq
\widehat\Phi_h(\xv):=\xv+\Pcal \Fv_h(\xv),
\eeq
the $\ell$-step semigroup defect is given by
\beq
\mathcal S_\ell(h)^2:=\left\|\widehat\Phi_{\ell h}-\widehat\Phi_h^\ell\right\|_{L^2(\mu)}^2 ,
\label{eq:semigroup_defect}
\eeq
where $\widehat\Phi_h^\ell$ denotes $\ell$-fold composition.
Since $\Phi_{\ell h}=\Phi_h^\ell$ for the exact flow, it follows that $\widehat\Phi_{\ell h}-\widehat\Phi_h^\ell = (\widehat\Phi_{\ell h}-\Phi_{\ell h}) + (\Phi_h^\ell-\widehat\Phi_h^\ell)$ and the semigroup defect can be bounded as
\beq
\mathcal S_\ell(h) \leq \sqrt{\mathcal E(\ell h)}+C_\ell \sqrt{\mathcal E(h)} 
\label{eq:semigroup_bound_training}
\eeq
in terms of the increment-prediction training error [\cref{eq:population_projection_error}], $\mathcal E(h)=\|\Qcal \Fv_h(\xv)\|_{L^2(\mu)}^2 =\|\Phi_h(\xv)-\widehat\Phi_h(\xv)\|^2$.
Here, $C_\ell$ depends on Lipschitz constants of the flow, assuming stability along intermediate trajectories. 

If the first unresolved Lie coefficient occurs at order
$m_0\geq 2$, i.e. $\qv_1=\cdots=\qv_{m_0-1}=0$, $\qv_{m_0}\neq 0$,
then, for fixed $\ell$,
\beq
\widehat\Phi_{\ell h}-\widehat\Phi_h^\ell=-\frac{\ell^{m_0}-\ell}{m_0!} h^{m_0}\qv_{m_0} + \Ocal(h^{m_0+1})
\label{eq:multistep_semigroup_leading}
\eeq
Combining this with \cref{eq:true_asymptotic_scaling}, i.e., $\mathcal E(h)=\frac{h^{2m_0}}{(m_0!)^2} \|\qv_{m_0}\|_{L^2(\mu)}^2 + \Ocal(h^{2m_0+1})$, gives
\beq
\mathcal S_\ell(h)^2=(\ell^{m_0}-\ell)^2\mathcal E(h)+\Ocal(h^{2m_0+1}).
\label{eq:semigroup_training_relation}
\eeq
Thus, whenever the vector field and the early Lie coefficients are exactly represented, the semigroup defect has the same asymptotic exponent as the training error. 
In particular, for a polynomial vector field and a polynomial library, $m_0=r^\ast+1$ with $r^\ast$ given by \cref{eq:poly_scaling_generic}.

By contrast, if the vector field is not exactly representable, $\qv_1\neq 0$, then the $\Ocal(h)$ terms cancel in $\widehat\Phi_{\ell h}-\widehat\Phi_h^\ell$, such that the semigroup defect starts at order $h^2$. Consequently, semigroup consistency is in this case at best an internal diagnostic of whether the learned map is has flow-map character.


\section{Remarks on the linear delay extrapolation stencil}
\label{app_delay_extrap}

A linear delay model is defined by 
\beq
\boldsymbol{\psi}((n+1)\Dt) = H_\mathrm{ld}\boldsymbol{\psi}(n\Dt),
\label{eq:app_delay_statespace}
\eeq
with the delay state 
\beq
\boldsymbol{\psi}(n \Dt) =
\begin{bmatrix}
\xv(n\Dt)\\
\xv((n-1)\Dt)\\
\vdots\\
\xv((n-\Gamma+1)\Dt)
\end{bmatrix}
\in \mathbb{R}^{d \Gamma}
\eeq
and the companion (delay-shift) matrix 
\beq
H_\mathrm{ld}
=
\begin{bmatrix}
\zeta_0 \Imat_d & \zeta_1 \Imat_d & \cdots & \zeta_{\Gamma-1} \Imat_d\\
\Imat_d & 0 & \cdots & 0\\
0 & \Imat_d & \cdots & 0\\
\vdots & & \ddots & \vdots\\
0 & \cdots & \Imat_d & 0
\end{bmatrix} \in \mathbb{R}^{d \Gamma\times d \Gamma}.
\label{eq:app_Hst}
\eeq
For general coefficients $\zeta_j$, the block companion matrix has characteristic polynomial
\[
\chi_{H_\mathrm{ld}}(\lambda)
=
\left(
\lambda^\Gamma-\sum_{j=0}^{\Gamma-1}\zeta_j\lambda^{\Gamma-1-j}
\right)^d .
\]
For the polynomial extrapolation stencil [\cref{eq_delay_extrap_predictor}] $\zeta_j=(-1)^j\binom{\Gamma}{j+1}$, this reduces to $\chi_{H_\mathrm{ld}}(\lambda)=(\lambda-1)^{d\Gamma}$, which implies a single eigenvalue $\lambda=1$ with algebraic multiplicity $d \Gamma$. 
A $d=1$ harmonic signal $x(t) = \sum_{k=1}^N (a_k \cos(k\omega_0 t) + b_k \sin(k\omega_0 t))$ can be modeled with $\Gamma= 2N$ time taps \cite{so_linear_2005,vaseghi_advanced_2008,pan_structure_2020}. 
The $\zeta_k$ are then given by the coefficients of the characteristic polynomial $\prod_{k=1}^N(\lambda^2-2\cos(k\omega_0\Delta t)\lambda+1)$ expanded in $\lambda$ \cite{stoica_spectral_2005,oppenheim_discretetime_2009}.



\section{Training error and forecast horizon}
\label{app_train_err}

We review here how the training error of an NG-RC-type model trained on dynamical system trajectories $x(t)$ affect the forecast horizon.
This is established theory in numerical analysis of time series \cite{kantz_nonlinear_2004,abarbanel_analysis_1993,kennel_prediction_1994}, recast here in the language of NG-RC.

\subsection{Residual error and model bias}

Consider the flow map $\Phi_{\Delta t}(x)$, acting as
\beq
x(t+\Delta t)=\Phi_{\Delta t}\bigl(x(t)\bigr),
\eeq 
and its approximation by a learned next-step map 
\beq
\hat{x}(t+\Delta t)= \hat H\,g\bigl(\hat{x}(t)\bigr),
\eeq
where $g=(x_1,\ldots,x_d,\psi_1,\ldots,\psi_m)\in \reals^n$ is a reservoir vector. 
Consider the residual
\beq
R(x) \equiv \Phi_{\Delta t}(x) - \hat H\,g(x),
\eeq
which we assume to be Lipschitz continuous with some constant $L_R$, i.e., $\|R(x)-R(y)\| \le L_R \|x-y\|$.
On the training data, we assume the uniform bound 
\beq 
\|R(x)\| \le  \kappa \varepsilon_{\mathrm{train}}, \qquad (\kappa\sim \Ocal(1))
\label{eq_trainerr_bound}\eeq
where we assume the residual is sufficiently concentrated such that it can be bounded by the training error up to a factor of $\Ocal(1)$ (which we neglect henceforth).
Denoting by $\hat x$ a point outside the training data (generated, e.g., by the autonomous forecast), it follows from the triangle inequality that
\beq
\|R(\hat x)\| \le  \|R(x)\| + L_R \|x-\hat x\| \le \varepsilon_{\mathrm{train}} + L_R \|x-\hat x\|.
\label{eq_residual_bound}\eeq
The second term on the r.h.s.\ captures the effect of model misspecification (bias), which happens if the trained model is structurally different from the true model. 

\subsection{Error recurrence in autonomous forecasting}

We study the error evolution during autonomous forecasting.
Let $x_{n+1}=\Phi_{\Delta t}(x_n)$ be the true dynamics and $\hat{x}_{n+1}=\hat H\,g(\hat{x}_n)$ be the autonomous prediction obtained from the trained model. Denoting the deviation between the two by $e_n=\|x_n-\hat{x}_n\|$, we have
\beq
e_{n+1}
=\|\Phi_{\Delta t}(x_n)-\hat H\,g(\hat{x}_n)\|
\le 
\|\Phi_{\Delta t}(x_n)-\Phi_{\Delta t}(\hat{x}_n)\|
+\|\Phi_{\Delta t}(\hat{x}_n)-\hat H\,g(\hat{x}_n)\|.
\label{eq_trainerr_evol}\eeq
Lipschitz continuity of the flow map implies 
$\|\Phi_{\Delta t}(x_n)-\Phi_{\Delta t}(\hat{x}_n)\|\le L_\Phi e_n$. In general, for small $\Delta t$, one can express the Lipschitz constant as $L_\Phi\approx 1+ \Delta t \|\Dcal f\|$ in terms of the Jacobian $\Dcal f$ of the dynamical system. 
We now use the residual bound of \cref{eq_residual_bound} in the second term in \cref{eq_trainerr_evol}, and assume the model generalizes well, such that the one-step error $\|R(x_n)\|$ on the true trajectory is also bounded by $\varepsilon_{\mathrm{train}}$.
This yields the recurrence relation
\beq
e_{n+1}\le \underbrace{(L_\Phi+L_R)}_{L} e_n+\varepsilon_{\mathrm{train}},
\label{eq_err_recur}\eeq 
where we introduced an effective Lipschitz constant $L$ that accounts for both the flow map approximation error and model misspecification.

\subsubsection{Stable dynamics}

For non-chaotic (precisely, non-expansive) systems, we take $L_\Phi \le 1$, while the error may still grow exponentially if $L_R$ is large. If the system is \emph{absolutely stable} ($L < 1$), iterating \cref{eq_err_recur} leads to the error bound $e_n \le L^n e_0 + \varepsilon_{\text{train}} (1 - L^n)/(1 - L)$. As $n \to \infty$, the error converges to a finite value $e_\infty = \varepsilon_{\text{train}} / (1 - L)$. If this asymptotic error is smaller than some given tolerance $E_{\text{tol}}$, the forecast horizon can be considered to be effectively infinite. If $e_\infty > E_{\text{tol}}$, the horizon is finite but typically determined by the initial transient rather than the accumulation of $\varepsilon_{\text{train}}$.

In the case of \emph{marginal stability} ($L = 1$, e.g., for conservative or periodic systems with $L_\Phi=1$ and $L_R=0$), the recurrence bound becomes $e_{n+1} \le e_n + \varepsilon_{\text{train}}$. Iterating this bound suggests a worst-case scenario where errors accumulate coherently, leading to linear error growth: $e_n \le e_0 + n \varepsilon_{\text{train}}$. If the forecast horizon $T\st{fch} = n \Delta t$ is defined by when $e_n$ reaches $E_{\text{tol}}$, this implies 
\beq T\st{fch} \approx \Delta t (E_{\text{tol}} - e_0) / \varepsilon_{\text{train}}.
\eeq 
By contrast, assume now that the one-step prediction errors $\delta_n = \Phi_{\Delta t}(\hat{x}_n) - \hat H g(\hat{x}_n)$ (with $\|\delta_n\| \le \varepsilon_{\text{train}}$) accumulate \emph{incoherently}, akin to steps in a random walk---a useful heuristic if $\varepsilon_{\text{train}}$ originates from observation noise that decorrelates sufficiently fast. Then the expected squared error grows linearly with time: $\mathbb{E}[\|e_n\|^2] \approx \|e_0\|^2 + n \mathbb{E}[\|\delta_k\|^2] \approx \|e_0\|^2 + n \varepsilon_{\text{train}}^2$ (interpreting $\varepsilon_{\text{train}}^2$ as the mean squared one-step error). In this scenario, setting the error magnitude to the tolerance $E_{\text{tol}}$ yields a forecast horizon 
\beq T\st{fch} \approx \Delta t (E_{\text{tol}}^2 - \|e_0\|^2) / \varepsilon_{\text{train}}^2,
\label{eq_pred_hor_incoh}\eeq 
which scales as $\varepsilon_{\text{train}}^{-2}$. The relevant scaling depends on the nature of the approximation error and how it propagates in the specific neutrally stable system.

\subsubsection{Chaotic dynamics forecast horizons}

We finally consider the case $L>1$, which not only includes \emph{chaotic} dynamics, but also exponential error growth due to large $L_R$.
For chaotic dynamics, one has $L_\Phi \simeq e^{\lambda_1 \Delta t}$ with $\lambda_1>0$ the largest Lyapunov exponent. 
Accordingly, iterating \cref{eq_err_recur} over $n$ steps gives
\beq
e_n \lesssim L^n e_0 +  \varepsilon_{\text{train}} \sum_{k=0}^{n-1} L^k,
\label{eq_errprop_sum}\eeq
where $e_0$ is introduced as the error representing the mismatch with the exact initial condition.
Evaluating the geometric series in \cref{eq_errprop_sum} as $\sum_{k=0}^{n-1} L^k = \frac{1 - L^n}{1 - L} \approx \frac{L^n}{L - 1}$ for large $n$, we obtain
\beq 
e_n \lesssim L^n \left( e_0 + \frac{\varepsilon_{\text{train}}}{L - 1} \right).
\eeq 
Both an imperfect initial condition and a nonzero training error feed into the exponential divergence between true and predicted dynamics.

\begin{figure}[tb]
    \centering
    \includegraphics[width = 0.5\linewidth]{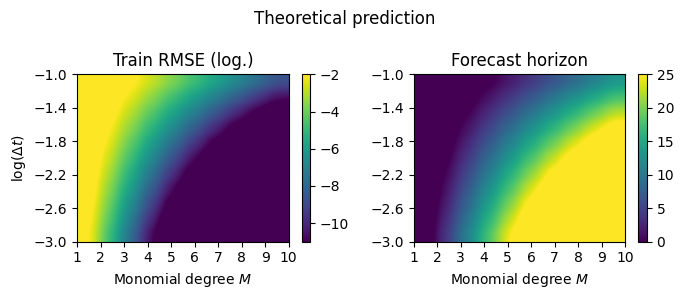}
    \caption{Theoretically predicted training error $\varepsilon\st{train}\simeq a \Dt^{r^\ast+1}$ [\cref{eq_trainerr_poly}] and forecast horizon $T\st{fch}\simeq \ln(C/\varepsilon\st{train})$ [\cref{eq_forec_hor}] as functions of $\Dt$ and $M$. We use values for the fitting parameters $a$ and $C\equiv \varepsilon\st{tol}/c$ obtained from numerical experiments.} 
    \label{fig_trainerr_fch_theory}
\end{figure}

Defining a tolerance $\varepsilon_{\mathrm{tol}}$ for the accumulated error, such that $e_n \lesssim \varepsilon_{\mathrm{tol}}$ signifies good forecasting accuracy, one obtains the prediction horizon
\beq 
T\st{fch}=n \Delta t \simeq \frac{1}{\lambda} \ln \left(\frac{\varepsilon\st{tol}}{e_0+ c\varepsilon\st{train}}\right),
\label{eq_forec_hor}\eeq
with the rate constant (effective Lyapunov exponent) $\lambda\equiv \Dt^{-1} \ln L$ and the constant $c\equiv 1/(L-1)$. 
As an example, if $L\simeq e^{\lambda_1\Dt}$ with $\lambda_1\sim\Ocal(1)$ and $\Dt\sim\Ocal(10^{-1}-10^{-3})$, then $c=(L-1)^{-1}\sim\Ocal(10-1000)$.
Taking $\varepsilon\st{tol}\sim\Ocal(10^{-1})$ and $\varepsilon\st{train}\sim\Ocal(10^{-2}-10^{-7})$, the dimensionless horizon $\lambda T\st{fch}$ ranges from values close to zero, when $c\varepsilon\st{train}\gtrsim\varepsilon\st{tol}$, up to $\Ocal(10)$ for the smallest training errors, neglecting the influence of the initial condition, i.e., $e_0\approx 0$.
\Cref{fig_trainerr_fch_theory} illustrates the relation between the training error and the forecast horizon for a well-specified (bias-free) model.

\bibliography{bibliography}

\begin{thebibliography}{112}%
\makeatletter
\providecommand \@ifxundefined [1]{%
 \@ifx{#1\undefined}
}%
\providecommand \@ifnum [1]{%
 \ifnum #1\expandafter \@firstoftwo
 \else \expandafter \@secondoftwo
 \fi
}%
\providecommand \@ifx [1]{%
 \ifx #1\expandafter \@firstoftwo
 \else \expandafter \@secondoftwo
 \fi
}%
\providecommand \natexlab [1]{#1}%
\providecommand \enquote  [1]{``#1''}%
\providecommand \bibnamefont  [1]{#1}%
\providecommand \bibfnamefont [1]{#1}%
\providecommand \citenamefont [1]{#1}%
\providecommand \href@noop [0]{\@secondoftwo}%
\providecommand \href [0]{\begingroup \@sanitize@url \@href}%
\providecommand \@href[1]{\@@startlink{#1}\@@href}%
\providecommand \@@href[1]{\endgroup#1\@@endlink}%
\providecommand \@sanitize@url [0]{\catcode `\\12\catcode `\$12\catcode `\&12\catcode `\#12\catcode `\^12\catcode `\_12\catcode `\%12\relax}%
\providecommand \@@startlink[1]{}%
\providecommand \@@endlink[0]{}%
\providecommand \url  [0]{\begingroup\@sanitize@url \@url }%
\providecommand \@url [1]{\endgroup\@href {#1}{\urlprefix }}%
\providecommand \urlprefix  [0]{URL }%
\providecommand \Eprint [0]{\href }%
\providecommand \doibase [0]{https://doi.org/}%
\providecommand \selectlanguage [0]{\@gobble}%
\providecommand \bibinfo  [0]{\@secondoftwo}%
\providecommand \bibfield  [0]{\@secondoftwo}%
\providecommand \translation [1]{[#1]}%
\providecommand \BibitemOpen [0]{}%
\providecommand \bibitemStop [0]{}%
\providecommand \bibitemNoStop [0]{.\EOS\space}%
\providecommand \EOS [0]{\spacefactor3000\relax}%
\providecommand \BibitemShut  [1]{\csname bibitem#1\endcsname}%
\let\auto@bib@innerbib\@empty
\bibitem [{\citenamefont {Luko{\v s}evi{\v c}ius}(2012)}]{lukosevicius_practical_2012}%
  \BibitemOpen
  \bibfield  {author} {\bibinfo {author} {\bibfnamefont {M.}~\bibnamefont {Luko{\v s}evi{\v c}ius}},\ }\bibfield  {title} {\bibinfo {title} {A {{Practical Guide}} to {{Applying Echo State Networks}}},\ }in\ \href {https://doi.org/10.1007/978-3-642-35289-8_36} {\emph {\bibinfo {booktitle} {Neural {{Networks}}: {{Tricks}} of the {{Trade}}}}},\ Vol.\ \bibinfo {volume} {7700},\ \bibinfo {editor} {edited by\ \bibinfo {editor} {\bibfnamefont {G.}~\bibnamefont {Montavon}}, \bibinfo {editor} {\bibfnamefont {G.~B.}\ \bibnamefont {Orr}},\ and\ \bibinfo {editor} {\bibfnamefont {K.-R.}\ \bibnamefont {M{\"u}ller}}}\ (\bibinfo  {publisher} {Springer Berlin Heidelberg},\ \bibinfo {address} {Berlin, Heidelberg},\ \bibinfo {year} {2012})\ pp.\ \bibinfo {pages} {659--686}\BibitemShut {NoStop}%
\bibitem [{\citenamefont {Gilpin}(2023)}]{gilpin_model_2023}%
  \BibitemOpen
  \bibfield  {author} {\bibinfo {author} {\bibfnamefont {W.}~\bibnamefont {Gilpin}},\ }\bibfield  {title} {\bibinfo {title} {Model scale versus domain knowledge in statistical forecasting of chaotic systems},\ }\href {https://doi.org/10.1103/PhysRevResearch.5.043252} {\bibfield  {journal} {\bibinfo  {journal} {Phys. Rev. Research}\ }\textbf {\bibinfo {volume} {5}},\ \bibinfo {pages} {043252} (\bibinfo {year} {2023})}\BibitemShut {NoStop}%
\bibitem [{\citenamefont {Yan}\ \emph {et~al.}(2024)\citenamefont {Yan}, \citenamefont {Huang}, \citenamefont {Bienstman}, \citenamefont {Tino}, \citenamefont {Lin},\ and\ \citenamefont {Sun}}]{yan_emerging_2024}%
  \BibitemOpen
  \bibfield  {author} {\bibinfo {author} {\bibfnamefont {M.}~\bibnamefont {Yan}}, \bibinfo {author} {\bibfnamefont {C.}~\bibnamefont {Huang}}, \bibinfo {author} {\bibfnamefont {P.}~\bibnamefont {Bienstman}}, \bibinfo {author} {\bibfnamefont {P.}~\bibnamefont {Tino}}, \bibinfo {author} {\bibfnamefont {W.}~\bibnamefont {Lin}},\ and\ \bibinfo {author} {\bibfnamefont {J.}~\bibnamefont {Sun}},\ }\bibfield  {title} {\bibinfo {title} {Emerging opportunities and challenges for the future of reservoir computing},\ }\href {https://doi.org/10.1038/s41467-024-45187-1} {\bibfield  {journal} {\bibinfo  {journal} {Nature Communications}\ }\textbf {\bibinfo {volume} {15}},\ \bibinfo {pages} {2056} (\bibinfo {year} {2024})}\BibitemShut {NoStop}%
\bibitem [{\citenamefont {Tanaka}\ \emph {et~al.}(2019)\citenamefont {Tanaka}, \citenamefont {Yamane}, \citenamefont {H{\'e}roux}, \citenamefont {Nakane}, \citenamefont {Kanazawa}, \citenamefont {Takeda}, \citenamefont {Numata}, \citenamefont {Nakano},\ and\ \citenamefont {Hirose}}]{tanaka_recent_2019}%
  \BibitemOpen
  \bibfield  {author} {\bibinfo {author} {\bibfnamefont {G.}~\bibnamefont {Tanaka}}, \bibinfo {author} {\bibfnamefont {T.}~\bibnamefont {Yamane}}, \bibinfo {author} {\bibfnamefont {J.~B.}\ \bibnamefont {H{\'e}roux}}, \bibinfo {author} {\bibfnamefont {R.}~\bibnamefont {Nakane}}, \bibinfo {author} {\bibfnamefont {N.}~\bibnamefont {Kanazawa}}, \bibinfo {author} {\bibfnamefont {S.}~\bibnamefont {Takeda}}, \bibinfo {author} {\bibfnamefont {H.}~\bibnamefont {Numata}}, \bibinfo {author} {\bibfnamefont {D.}~\bibnamefont {Nakano}},\ and\ \bibinfo {author} {\bibfnamefont {A.}~\bibnamefont {Hirose}},\ }\bibfield  {title} {\bibinfo {title} {Recent advances in physical reservoir computing: {{A}} review},\ }\href {https://doi.org/10.1016/j.neunet.2019.03.005} {\bibfield  {journal} {\bibinfo  {journal} {Neural Networks}\ }\textbf {\bibinfo {volume} {115}},\ \bibinfo {pages} {100} (\bibinfo {year} {2019})}\BibitemShut {NoStop}%
\bibitem [{\citenamefont {Gauthier}\ \emph {et~al.}(2021)\citenamefont {Gauthier}, \citenamefont {Bollt}, \citenamefont {Griffith},\ and\ \citenamefont {Barbosa}}]{gauthier_next_2021}%
  \BibitemOpen
  \bibfield  {author} {\bibinfo {author} {\bibfnamefont {D.~J.}\ \bibnamefont {Gauthier}}, \bibinfo {author} {\bibfnamefont {E.}~\bibnamefont {Bollt}}, \bibinfo {author} {\bibfnamefont {A.}~\bibnamefont {Griffith}},\ and\ \bibinfo {author} {\bibfnamefont {W.~A.~S.}\ \bibnamefont {Barbosa}},\ }\bibfield  {title} {\bibinfo {title} {Next generation reservoir computing},\ }\href {https://doi.org/10.1038/s41467-021-25801-2} {\bibfield  {journal} {\bibinfo  {journal} {Nature Communications}\ }\textbf {\bibinfo {volume} {12}},\ \bibinfo {pages} {5564} (\bibinfo {year} {2021})}\BibitemShut {NoStop}%
\bibitem [{\citenamefont {Shahi}\ \emph {et~al.}(2022)\citenamefont {Shahi}, \citenamefont {Fenton},\ and\ \citenamefont {Cherry}}]{shahi_prediction_2022}%
  \BibitemOpen
  \bibfield  {author} {\bibinfo {author} {\bibfnamefont {S.}~\bibnamefont {Shahi}}, \bibinfo {author} {\bibfnamefont {F.~H.}\ \bibnamefont {Fenton}},\ and\ \bibinfo {author} {\bibfnamefont {E.~M.}\ \bibnamefont {Cherry}},\ }\bibfield  {title} {\bibinfo {title} {Prediction of chaotic time series using recurrent neural networks and reservoir computing techniques: {{A}} comparative study},\ }\href {https://doi.org/10.1016/j.mlwa.2022.100300} {\bibfield  {journal} {\bibinfo  {journal} {Machine Learning with Applications}\ }\textbf {\bibinfo {volume} {8}},\ \bibinfo {pages} {100300} (\bibinfo {year} {2022})}\BibitemShut {NoStop}%
\bibitem [{\citenamefont {Barbosa}\ and\ \citenamefont {Gauthier}(2022)}]{barbosa_learning_2022}%
  \BibitemOpen
  \bibfield  {author} {\bibinfo {author} {\bibfnamefont {W.~A.~S.}\ \bibnamefont {Barbosa}}\ and\ \bibinfo {author} {\bibfnamefont {D.~J.}\ \bibnamefont {Gauthier}},\ }\bibfield  {title} {\bibinfo {title} {Learning spatiotemporal chaos using next-generation reservoir computing},\ }\href {https://doi.org/10.1063/5.0098707} {\bibfield  {journal} {\bibinfo  {journal} {Chaos: An Interdisciplinary Journal of Nonlinear Science}\ }\textbf {\bibinfo {volume} {32}},\ \bibinfo {pages} {093137} (\bibinfo {year} {2022})}\BibitemShut {NoStop}%
\bibitem [{\citenamefont {Flynn}\ \emph {et~al.}(2022)\citenamefont {Flynn}, \citenamefont {Heilmann}, \citenamefont {K{\"o}glmayr}, \citenamefont {Tsachouridis}, \citenamefont {R{\"a}th},\ and\ \citenamefont {Amann}}]{flynn_exploring_2022}%
  \BibitemOpen
  \bibfield  {author} {\bibinfo {author} {\bibfnamefont {A.}~\bibnamefont {Flynn}}, \bibinfo {author} {\bibfnamefont {O.}~\bibnamefont {Heilmann}}, \bibinfo {author} {\bibfnamefont {D.}~\bibnamefont {K{\"o}glmayr}}, \bibinfo {author} {\bibfnamefont {V.~A.}\ \bibnamefont {Tsachouridis}}, \bibinfo {author} {\bibfnamefont {C.}~\bibnamefont {R{\"a}th}},\ and\ \bibinfo {author} {\bibfnamefont {A.}~\bibnamefont {Amann}},\ }\href {https://doi.org/10.48550/arXiv.2205.11375} {\bibinfo {title} {Exploring the limits of multifunctionality across different reservoir computers}} (\bibinfo {year} {2022}),\ \Eprint {https://arxiv.org/abs/2205.11375} {arXiv:2205.11375 [cs]} \BibitemShut {NoStop}%
\bibitem [{\citenamefont {K{\"o}glmayr}\ and\ \citenamefont {R{\"a}th}(2024)}]{koglmayr_extrapolating_2024}%
  \BibitemOpen
  \bibfield  {author} {\bibinfo {author} {\bibfnamefont {D.}~\bibnamefont {K{\"o}glmayr}}\ and\ \bibinfo {author} {\bibfnamefont {C.}~\bibnamefont {R{\"a}th}},\ }\bibfield  {title} {\bibinfo {title} {Extrapolating tipping points and simulating non-stationary dynamics of complex systems using efficient machine learning},\ }\href {https://doi.org/10.1038/s41598-023-50726-9} {\bibfield  {journal} {\bibinfo  {journal} {Scientific Reports}\ }\textbf {\bibinfo {volume} {14}},\ \bibinfo {pages} {507} (\bibinfo {year} {2024})}\BibitemShut {NoStop}%
\bibitem [{\citenamefont {Brucke}\ \emph {et~al.}(2024)\citenamefont {Brucke}, \citenamefont {Schmitz}, \citenamefont {K{\"o}glmayr}, \citenamefont {Baur}, \citenamefont {R{\"a}th}, \citenamefont {Ansari},\ and\ \citenamefont {Klement}}]{brucke_benchmarking_2024}%
  \BibitemOpen
  \bibfield  {author} {\bibinfo {author} {\bibfnamefont {K.}~\bibnamefont {Brucke}}, \bibinfo {author} {\bibfnamefont {S.}~\bibnamefont {Schmitz}}, \bibinfo {author} {\bibfnamefont {D.}~\bibnamefont {K{\"o}glmayr}}, \bibinfo {author} {\bibfnamefont {S.}~\bibnamefont {Baur}}, \bibinfo {author} {\bibfnamefont {C.}~\bibnamefont {R{\"a}th}}, \bibinfo {author} {\bibfnamefont {E.}~\bibnamefont {Ansari}},\ and\ \bibinfo {author} {\bibfnamefont {P.}~\bibnamefont {Klement}},\ }\bibfield  {title} {\bibinfo {title} {Benchmarking reservoir computing for residential energy demand forecasting},\ }\href {https://doi.org/10.1016/j.enbuild.2024.114236} {\bibfield  {journal} {\bibinfo  {journal} {Energy and Buildings}\ }\textbf {\bibinfo {volume} {314}},\ \bibinfo {pages} {114236} (\bibinfo {year} {2024})}\BibitemShut {NoStop}%
\bibitem [{\citenamefont {Sch{\"o}tz}\ \emph {et~al.}(2025)\citenamefont {Sch{\"o}tz}, \citenamefont {White}, \citenamefont {Gelbrecht},\ and\ \citenamefont {Boers}}]{schotz_machine_2025}%
  \BibitemOpen
  \bibfield  {author} {\bibinfo {author} {\bibfnamefont {C.}~\bibnamefont {Sch{\"o}tz}}, \bibinfo {author} {\bibfnamefont {A.}~\bibnamefont {White}}, \bibinfo {author} {\bibfnamefont {M.}~\bibnamefont {Gelbrecht}},\ and\ \bibinfo {author} {\bibfnamefont {N.}~\bibnamefont {Boers}},\ }\href {https://doi.org/10.48550/arXiv.2407.20158} {\bibinfo {title} {Machine {{Learning}} for {{Predicting Chaotic Systems}}}} (\bibinfo {year} {2025}),\ \Eprint {https://arxiv.org/abs/2407.20158} {arXiv:2407.20158 [cs]} \BibitemShut {NoStop}%
\bibitem [{\citenamefont {Bollt}(2021)}]{bollt_explaining_2021}%
  \BibitemOpen
  \bibfield  {author} {\bibinfo {author} {\bibfnamefont {E.}~\bibnamefont {Bollt}},\ }\bibfield  {title} {\bibinfo {title} {On {{Explaining}} the {{Surprising Success}} of {{Reservoir Computing Forecaster}} of {{Chaos}}? {{The Universal Machine Learning Dynamical System}} with {{Contrasts}} to {{VAR}} and {{DMD}}},\ }\href {https://doi.org/10.1063/5.0024890} {\bibfield  {journal} {\bibinfo  {journal} {Chaos: An Interdisciplinary Journal of Nonlinear Science}\ }\textbf {\bibinfo {volume} {31}},\ \bibinfo {pages} {013108} (\bibinfo {year} {2021})},\ \Eprint {https://arxiv.org/abs/2008.06530} {arXiv:2008.06530 [physics]} \BibitemShut {NoStop}%
\bibitem [{\citenamefont {Leontaritis}\ and\ \citenamefont {Billings}(1985)}]{leontaritis_inputoutput_1985}%
  \BibitemOpen
  \bibfield  {author} {\bibinfo {author} {\bibfnamefont {I.~J.}\ \bibnamefont {Leontaritis}}\ and\ \bibinfo {author} {\bibfnamefont {S.~A.}\ \bibnamefont {Billings}},\ }\bibfield  {title} {\bibinfo {title} {Input-output parametric models for non-linear systems {{Part I}}: Deterministic non-linear systems},\ }\href {https://doi.org/10.1080/0020718508961129} {\bibfield  {journal} {\bibinfo  {journal} {International Journal of Control}\ }\textbf {\bibinfo {volume} {41}},\ \bibinfo {pages} {303} (\bibinfo {year} {1985})}\BibitemShut {NoStop}%
\bibitem [{\citenamefont {Kekatos}\ and\ \citenamefont {Giannakis}(2011)}]{kekatos_sparse_2011}%
  \BibitemOpen
  \bibfield  {author} {\bibinfo {author} {\bibfnamefont {V.}~\bibnamefont {Kekatos}}\ and\ \bibinfo {author} {\bibfnamefont {G.~B.}\ \bibnamefont {Giannakis}},\ }\bibfield  {title} {\bibinfo {title} {Sparse {{Volterra}} and {{Polynomial Regression Models}}: {{Recoverability}} and {{Estimation}}},\ }\href {https://doi.org/10.1109/TSP.2011.2165952} {\bibfield  {journal} {\bibinfo  {journal} {IEEE Transactions on Signal Processing}\ }\textbf {\bibinfo {volume} {59}},\ \bibinfo {pages} {5907} (\bibinfo {year} {2011})},\ \Eprint {https://arxiv.org/abs/1103.0769} {arXiv:1103.0769 [cs]} \BibitemShut {NoStop}%
\bibitem [{\citenamefont {Billings}(2013)}]{billings_nonlinear_2013}%
  \BibitemOpen
  \bibfield  {author} {\bibinfo {author} {\bibfnamefont {S.~A.}\ \bibnamefont {Billings}},\ }\href@noop {} {\emph {\bibinfo {title} {Nonlinear {{System Identification}}: {{NARMAX Methods}} in the {{Time}}, {{Frequency}}, and {{Spatio-Temporal Domains}}}}}\ (\bibinfo  {publisher} {Wiley},\ \bibinfo {year} {2013})\BibitemShut {NoStop}%
\bibitem [{\citenamefont {Huang}\ \emph {et~al.}(2006)\citenamefont {Huang}, \citenamefont {Zhu},\ and\ \citenamefont {Siew}}]{huang_extreme_2006}%
  \BibitemOpen
  \bibfield  {author} {\bibinfo {author} {\bibfnamefont {G.-B.}\ \bibnamefont {Huang}}, \bibinfo {author} {\bibfnamefont {Q.-Y.}\ \bibnamefont {Zhu}},\ and\ \bibinfo {author} {\bibfnamefont {C.-K.}\ \bibnamefont {Siew}},\ }\bibfield  {title} {\bibinfo {title} {Extreme learning machine: {{Theory}} and applications},\ }\href {https://doi.org/10.1016/j.neucom.2005.12.126} {\bibfield  {journal} {\bibinfo  {journal} {Neurocomputing}\ }\bibinfo {series} {Neural {{Networks}}},\ \textbf {\bibinfo {volume} {70}},\ \bibinfo {pages} {489} (\bibinfo {year} {2006})}\BibitemShut {NoStop}%
\bibitem [{\citenamefont {{van Heeswijk}}\ \emph {et~al.}(2009)\citenamefont {{van Heeswijk}}, \citenamefont {Miche}, \citenamefont {{Lindh-Knuutila}}, \citenamefont {Hilbers}, \citenamefont {Honkela}, \citenamefont {Oja},\ and\ \citenamefont {Lendasse}}]{vanheeswijk_adaptive_2009}%
  \BibitemOpen
  \bibfield  {author} {\bibinfo {author} {\bibfnamefont {M.}~\bibnamefont {{van Heeswijk}}}, \bibinfo {author} {\bibfnamefont {Y.}~\bibnamefont {Miche}}, \bibinfo {author} {\bibfnamefont {T.}~\bibnamefont {{Lindh-Knuutila}}}, \bibinfo {author} {\bibfnamefont {P.~A.~J.}\ \bibnamefont {Hilbers}}, \bibinfo {author} {\bibfnamefont {T.}~\bibnamefont {Honkela}}, \bibinfo {author} {\bibfnamefont {E.}~\bibnamefont {Oja}},\ and\ \bibinfo {author} {\bibfnamefont {A.}~\bibnamefont {Lendasse}},\ }\bibfield  {title} {\bibinfo {title} {Adaptive {{Ensemble Models}} of {{Extreme Learning Machines}} for {{Time Series Prediction}}},\ }in\ \href {https://doi.org/10.1007/978-3-642-04277-5_31} {\emph {\bibinfo {booktitle} {Artificial {{Neural Networks}} -- {{ICANN}} 2009}}},\ \bibinfo {editor} {edited by\ \bibinfo {editor} {\bibfnamefont {C.}~\bibnamefont {Alippi}}, \bibinfo {editor} {\bibfnamefont {M.}~\bibnamefont {Polycarpou}}, \bibinfo {editor} {\bibfnamefont {C.}~\bibnamefont {Panayiotou}},\ and\ \bibinfo {editor} {\bibfnamefont {G.}~\bibnamefont {Ellinas}}}\ (\bibinfo  {publisher} {Springer},\ \bibinfo {address} {Berlin, Heidelberg},\ \bibinfo {year} {2009})\ pp.\ \bibinfo {pages} {305--314}\BibitemShut {NoStop}%
\bibitem [{\citenamefont {Huang}\ \emph {et~al.}(2011)\citenamefont {Huang}, \citenamefont {Wang},\ and\ \citenamefont {Lan}}]{huang_extreme_2011}%
  \BibitemOpen
  \bibfield  {author} {\bibinfo {author} {\bibfnamefont {G.-B.}\ \bibnamefont {Huang}}, \bibinfo {author} {\bibfnamefont {D.~H.}\ \bibnamefont {Wang}},\ and\ \bibinfo {author} {\bibfnamefont {Y.}~\bibnamefont {Lan}},\ }\bibfield  {title} {\bibinfo {title} {Extreme learning machines: A survey},\ }\href {https://doi.org/10.1007/s13042-011-0019-y} {\bibfield  {journal} {\bibinfo  {journal} {International Journal of Machine Learning and Cybernetics}\ }\textbf {\bibinfo {volume} {2}},\ \bibinfo {pages} {107} (\bibinfo {year} {2011})}\BibitemShut {NoStop}%
\bibitem [{\citenamefont {Butcher}\ \emph {et~al.}(2013)\citenamefont {Butcher}, \citenamefont {Verstraeten}, \citenamefont {Schrauwen}, \citenamefont {Day},\ and\ \citenamefont {Haycock}}]{butcher_reservoir_2013}%
  \BibitemOpen
  \bibfield  {author} {\bibinfo {author} {\bibfnamefont {J.~B.}\ \bibnamefont {Butcher}}, \bibinfo {author} {\bibfnamefont {D.}~\bibnamefont {Verstraeten}}, \bibinfo {author} {\bibfnamefont {B.}~\bibnamefont {Schrauwen}}, \bibinfo {author} {\bibfnamefont {C.~R.}\ \bibnamefont {Day}},\ and\ \bibinfo {author} {\bibfnamefont {P.~W.}\ \bibnamefont {Haycock}},\ }\bibfield  {title} {\bibinfo {title} {Reservoir computing and extreme learning machines for non-linear time-series data analysis},\ }\href {https://doi.org/10.1016/j.neunet.2012.11.011} {\bibfield  {journal} {\bibinfo  {journal} {Neural Networks}\ }\textbf {\bibinfo {volume} {38}},\ \bibinfo {pages} {76} (\bibinfo {year} {2013})}\BibitemShut {NoStop}%
\bibitem [{\citenamefont {Rahimi}\ and\ \citenamefont {Recht}(2007)}]{rahimi_random_2007}%
  \BibitemOpen
  \bibfield  {author} {\bibinfo {author} {\bibfnamefont {A.}~\bibnamefont {Rahimi}}\ and\ \bibinfo {author} {\bibfnamefont {B.}~\bibnamefont {Recht}},\ }\bibfield  {title} {\bibinfo {title} {Random {{Features}} for {{Large-Scale Kernel Machines}}},\ }in\ \href@noop {} {\emph {\bibinfo {booktitle} {Advances in {{Neural Information Processing Systems}}}}},\ Vol.~\bibinfo {volume} {20}\ (\bibinfo  {publisher} {Curran Associates, Inc.},\ \bibinfo {year} {2007})\BibitemShut {NoStop}%
\bibitem [{\citenamefont {Williams}\ \emph {et~al.}(2015)\citenamefont {Williams}, \citenamefont {Kevrekidis},\ and\ \citenamefont {Rowley}}]{williams_data_2015}%
  \BibitemOpen
  \bibfield  {author} {\bibinfo {author} {\bibfnamefont {M.~O.}\ \bibnamefont {Williams}}, \bibinfo {author} {\bibfnamefont {I.~G.}\ \bibnamefont {Kevrekidis}},\ and\ \bibinfo {author} {\bibfnamefont {C.~W.}\ \bibnamefont {Rowley}},\ }\bibfield  {title} {\bibinfo {title} {A {{Data}}--{{Driven Approximation}} of the {{Koopman Operator}}: {{Extending Dynamic Mode Decomposition}}},\ }\href {https://doi.org/10.1007/s00332-015-9258-5} {\bibfield  {journal} {\bibinfo  {journal} {Journal of Nonlinear Science}\ }\textbf {\bibinfo {volume} {25}},\ \bibinfo {pages} {1307} (\bibinfo {year} {2015})}\BibitemShut {NoStop}%
\bibitem [{\citenamefont {Brunton}\ and\ \citenamefont {Kutz}(2019)}]{brunton_datadriven_2019}%
  \BibitemOpen
  \bibfield  {author} {\bibinfo {author} {\bibfnamefont {S.~L.}\ \bibnamefont {Brunton}}\ and\ \bibinfo {author} {\bibfnamefont {J.~N.}\ \bibnamefont {Kutz}},\ }\href {https://doi.org/10.1017/9781108380690} {\emph {\bibinfo {title} {Data-{{Driven Science}} and {{Engineering}}: {{Machine Learning}}, {{Dynamical Systems}}, and {{Control}}}}}\ (\bibinfo  {publisher} {Cambridge University Press},\ \bibinfo {address} {Cambridge},\ \bibinfo {year} {2019})\BibitemShut {NoStop}%
\bibitem [{\citenamefont {Tu}\ \emph {et~al.}(2014)\citenamefont {Tu}, \citenamefont {Rowley}, \citenamefont {Luchtenburg}, \citenamefont {Brunton},\ and\ \citenamefont {Kutz}}]{tu_dynamic_2014}%
  \BibitemOpen
  \bibfield  {author} {\bibinfo {author} {\bibfnamefont {J.~H.}\ \bibnamefont {Tu}}, \bibinfo {author} {\bibfnamefont {C.~W.}\ \bibnamefont {Rowley}}, \bibinfo {author} {\bibfnamefont {D.~M.}\ \bibnamefont {Luchtenburg}}, \bibinfo {author} {\bibfnamefont {S.~L.}\ \bibnamefont {Brunton}},\ and\ \bibinfo {author} {\bibfnamefont {J.~N.}\ \bibnamefont {Kutz}},\ }\bibfield  {title} {\bibinfo {title} {On {{Dynamic Mode Decomposition}}: {{Theory}} and {{Applications}}},\ }\href {https://doi.org/10.3934/jcd.2014.1.391} {\bibfield  {journal} {\bibinfo  {journal} {Journal of Computational Dynamics}\ }\textbf {\bibinfo {volume} {1}},\ \bibinfo {pages} {391} (\bibinfo {year} {2014})},\ \Eprint {https://arxiv.org/abs/1312.0041} {arXiv:1312.0041 [math]} \BibitemShut {NoStop}%
\bibitem [{\citenamefont {Chen}\ \emph {et~al.}(2012)\citenamefont {Chen}, \citenamefont {Tu},\ and\ \citenamefont {Rowley}}]{chen_variants_2012}%
  \BibitemOpen
  \bibfield  {author} {\bibinfo {author} {\bibfnamefont {K.~K.}\ \bibnamefont {Chen}}, \bibinfo {author} {\bibfnamefont {J.~H.}\ \bibnamefont {Tu}},\ and\ \bibinfo {author} {\bibfnamefont {C.~W.}\ \bibnamefont {Rowley}},\ }\bibfield  {title} {\bibinfo {title} {Variants of {{Dynamic Mode Decomposition}}: {{Boundary Condition}}, {{Koopman}}, and {{Fourier Analyses}}},\ }\href {https://doi.org/10.1007/s00332-012-9130-9} {\bibfield  {journal} {\bibinfo  {journal} {Journal of Nonlinear Science}\ }\textbf {\bibinfo {volume} {22}},\ \bibinfo {pages} {887} (\bibinfo {year} {2012})}\BibitemShut {NoStop}%
\bibitem [{\citenamefont {Kutz}\ \emph {et~al.}(2016)\citenamefont {Kutz}, \citenamefont {Brunton}, \citenamefont {Brunton},\ and\ \citenamefont {Proctor}}]{kutz_dynamic_2016}%
  \BibitemOpen
  \bibfield  {author} {\bibinfo {author} {\bibfnamefont {J.~N.}\ \bibnamefont {Kutz}}, \bibinfo {author} {\bibfnamefont {S.~L.}\ \bibnamefont {Brunton}}, \bibinfo {author} {\bibfnamefont {B.~W.}\ \bibnamefont {Brunton}},\ and\ \bibinfo {author} {\bibfnamefont {J.~L.}\ \bibnamefont {Proctor}},\ }\href@noop {} {\emph {\bibinfo {title} {Dynamic {{Mode Decomposition}}: {{Data-Driven Modeling}} of {{Complex Systems}}}}}\ (\bibinfo  {publisher} {{SIAM-Society for Industrial and Applied Mathematics}},\ \bibinfo {address} {Philadelphia, PA, USA},\ \bibinfo {year} {2016})\BibitemShut {NoStop}%
\bibitem [{\citenamefont {Korda}\ and\ \citenamefont {Mezi{\'c}}(2018)}]{korda_convergence_2018}%
  \BibitemOpen
  \bibfield  {author} {\bibinfo {author} {\bibfnamefont {M.}~\bibnamefont {Korda}}\ and\ \bibinfo {author} {\bibfnamefont {I.}~\bibnamefont {Mezi{\'c}}},\ }\bibfield  {title} {\bibinfo {title} {On {{Convergence}} of {{Extended Dynamic Mode Decomposition}} to the {{Koopman Operator}}},\ }\href {https://doi.org/10.1007/s00332-017-9423-0} {\bibfield  {journal} {\bibinfo  {journal} {Journal of Nonlinear Science}\ }\textbf {\bibinfo {volume} {28}},\ \bibinfo {pages} {687} (\bibinfo {year} {2018})}\BibitemShut {NoStop}%
\bibitem [{\citenamefont {Mezic}(2020)}]{mezic_koopman_2020}%
  \BibitemOpen
  \bibfield  {author} {\bibinfo {author} {\bibfnamefont {I.}~\bibnamefont {Mezic}},\ }\href {https://doi.org/10.48550/arXiv.2010.05377} {\bibinfo {title} {Koopman {{Operator}}, {{Geometry}}, and {{Learning}}}} (\bibinfo {year} {2020}),\ \Eprint {https://arxiv.org/abs/2010.05377} {arXiv:2010.05377 [math]} \BibitemShut {NoStop}%
\bibitem [{\citenamefont {Surana}(2020)}]{surana_koopman_2020}%
  \BibitemOpen
  \bibfield  {author} {\bibinfo {author} {\bibfnamefont {A.}~\bibnamefont {Surana}},\ }\bibfield  {title} {\bibinfo {title} {Koopman {{Operator Framework}} for {{Time Series Modeling}} and {{Analysis}}},\ }\href {https://doi.org/10.1007/s00332-017-9441-y} {\bibfield  {journal} {\bibinfo  {journal} {Journal of Nonlinear Science}\ }\textbf {\bibinfo {volume} {30}},\ \bibinfo {pages} {1973} (\bibinfo {year} {2020})}\BibitemShut {NoStop}%
\bibitem [{\citenamefont {Brunton}\ \emph {et~al.}(2022)\citenamefont {Brunton}, \citenamefont {Budi{\v s}i{\'c}}, \citenamefont {Kaiser},\ and\ \citenamefont {Kutz}}]{brunton_modern_2022}%
  \BibitemOpen
  \bibfield  {author} {\bibinfo {author} {\bibfnamefont {S.~L.}\ \bibnamefont {Brunton}}, \bibinfo {author} {\bibfnamefont {M.}~\bibnamefont {Budi{\v s}i{\'c}}}, \bibinfo {author} {\bibfnamefont {E.}~\bibnamefont {Kaiser}},\ and\ \bibinfo {author} {\bibfnamefont {J.~N.}\ \bibnamefont {Kutz}},\ }\bibfield  {title} {\bibinfo {title} {Modern {{Koopman Theory}} for {{Dynamical Systems}}},\ }\href {https://doi.org/10.1137/21M1401243} {\bibfield  {journal} {\bibinfo  {journal} {SIAM Review}\ }\textbf {\bibinfo {volume} {64}},\ \bibinfo {pages} {229} (\bibinfo {year} {2022})}\BibitemShut {NoStop}%
\bibitem [{\citenamefont {Mauroy}\ and\ \citenamefont {Goncalves}(2019)}]{mauroy_koopmanbased_2019}%
  \BibitemOpen
  \bibfield  {author} {\bibinfo {author} {\bibfnamefont {A.}~\bibnamefont {Mauroy}}\ and\ \bibinfo {author} {\bibfnamefont {J.}~\bibnamefont {Goncalves}},\ }\href {https://doi.org/10.48550/arXiv.1709.02003} {\bibinfo {title} {Koopman-based lifting techniques for nonlinear systems identification}} (\bibinfo {year} {2019}),\ \Eprint {https://arxiv.org/abs/1709.02003} {arXiv:1709.02003 [math]} \BibitemShut {NoStop}%
\bibitem [{\citenamefont {Bevanda}\ \emph {et~al.}(2021)\citenamefont {Bevanda}, \citenamefont {Sosnowski},\ and\ \citenamefont {Hirche}}]{bevanda_koopman_2021}%
  \BibitemOpen
  \bibfield  {author} {\bibinfo {author} {\bibfnamefont {P.}~\bibnamefont {Bevanda}}, \bibinfo {author} {\bibfnamefont {S.}~\bibnamefont {Sosnowski}},\ and\ \bibinfo {author} {\bibfnamefont {S.}~\bibnamefont {Hirche}},\ }\bibfield  {title} {\bibinfo {title} {Koopman {{Operator Dynamical Models}}: {{Learning}}, {{Analysis}} and {{Control}}},\ }\href {https://doi.org/10.1016/j.arcontrol.2021.09.002} {\bibfield  {journal} {\bibinfo  {journal} {Annual Reviews in Control}\ }\textbf {\bibinfo {volume} {52}},\ \bibinfo {pages} {197} (\bibinfo {year} {2021})},\ \Eprint {https://arxiv.org/abs/2102.02522} {arXiv:2102.02522 [eess]} \BibitemShut {NoStop}%
\bibitem [{\citenamefont {Otto}\ and\ \citenamefont {Rowley}(2021)}]{otto_koopman_2021}%
  \BibitemOpen
  \bibfield  {author} {\bibinfo {author} {\bibfnamefont {S.~E.}\ \bibnamefont {Otto}}\ and\ \bibinfo {author} {\bibfnamefont {C.~W.}\ \bibnamefont {Rowley}},\ }\bibfield  {title} {\bibinfo {title} {Koopman {{Operators}} for {{Estimation}} and {{Control}} of {{Dynamical Systems}}},\ }\href {https://doi.org/10.1146/annurev-control-071020-010108} {\bibfield  {journal} {\bibinfo  {journal} {Annual Review of Control, Robotics, and Autonomous Systems}\ }\textbf {\bibinfo {volume} {4}},\ \bibinfo {pages} {59} (\bibinfo {year} {2021})}\BibitemShut {NoStop}%
\bibitem [{\citenamefont {Ghosh}\ and\ \citenamefont {Mcafee}(2024)}]{ghosh_koopman_2024}%
  \BibitemOpen
  \bibfield  {author} {\bibinfo {author} {\bibfnamefont {R.}~\bibnamefont {Ghosh}}\ and\ \bibinfo {author} {\bibfnamefont {M.}~\bibnamefont {Mcafee}},\ }\bibfield  {title} {\bibinfo {title} {Koopman operator theory and dynamic mode decomposition in data-driven science and engineering: {{A}} comprehensive review},\ }\href {https://doi.org/10.53391/mmnsa.1512698} {\bibfield  {journal} {\bibinfo  {journal} {Mathematical Modelling and Numerical Simulation with Applications}\ }\textbf {\bibinfo {volume} {4}},\ \bibinfo {pages} {562} (\bibinfo {year} {2024})}\BibitemShut {NoStop}%
\bibitem [{\citenamefont {Shi}\ \emph {et~al.}(2026)\citenamefont {Shi}, \citenamefont {Haseli}, \citenamefont {Mamakoukas}, \citenamefont {Bruder}, \citenamefont {Abraham}, \citenamefont {Murphey}, \citenamefont {Cort{\'e}s},\ and\ \citenamefont {Karydis}}]{shi_koopman_2026}%
  \BibitemOpen
  \bibfield  {author} {\bibinfo {author} {\bibfnamefont {L.}~\bibnamefont {Shi}}, \bibinfo {author} {\bibfnamefont {M.}~\bibnamefont {Haseli}}, \bibinfo {author} {\bibfnamefont {G.}~\bibnamefont {Mamakoukas}}, \bibinfo {author} {\bibfnamefont {D.}~\bibnamefont {Bruder}}, \bibinfo {author} {\bibfnamefont {I.}~\bibnamefont {Abraham}}, \bibinfo {author} {\bibfnamefont {T.}~\bibnamefont {Murphey}}, \bibinfo {author} {\bibfnamefont {J.}~\bibnamefont {Cort{\'e}s}},\ and\ \bibinfo {author} {\bibfnamefont {K.}~\bibnamefont {Karydis}},\ }\bibfield  {title} {\bibinfo {title} {Koopman {{Operators}} in {{Robot Learning}}},\ }\href {https://doi.org/10.1109/TRO.2026.3654384} {\bibfield  {journal} {\bibinfo  {journal} {IEEE Transactions on Robotics}\ }\textbf {\bibinfo {volume} {42}},\ \bibinfo {pages} {1088} (\bibinfo {year} {2026})}\BibitemShut {NoStop}%
\bibitem [{\citenamefont {Kowalski}\ and\ \citenamefont {Steeb}(1991)}]{kowalski_nonlinear_1991}%
  \BibitemOpen
  \bibfield  {author} {\bibinfo {author} {\bibfnamefont {K.}~\bibnamefont {Kowalski}}\ and\ \bibinfo {author} {\bibfnamefont {W.-H.}\ \bibnamefont {Steeb}},\ }\href@noop {} {\emph {\bibinfo {title} {Nonlinear {{Dynamical Systems And Carleman Linearization}}}}}\ (\bibinfo  {publisher} {World Scientific},\ \bibinfo {year} {1991})\BibitemShut {NoStop}%
\bibitem [{\citenamefont {van Overschee}\ and\ \citenamefont {de~Moor}(1996)}]{overschee_subspace_1996}%
  \BibitemOpen
  \bibfield  {author} {\bibinfo {author} {\bibfnamefont {P.}~\bibnamefont {van Overschee}}\ and\ \bibinfo {author} {\bibfnamefont {B.~L.}\ \bibnamefont {de~Moor}},\ }\href@noop {} {\emph {\bibinfo {title} {{Subspace Identification for Linear Systems: Theory \texthorizontalbar{} Implementation \texthorizontalbar{} Applications}}}}\ (\bibinfo  {publisher} {Springer},\ \bibinfo {address} {Boston},\ \bibinfo {year} {1996})\BibitemShut {NoStop}%
\bibitem [{\citenamefont {Ljung}(1999)}]{ljung_system_1999}%
  \BibitemOpen
  \bibfield  {author} {\bibinfo {author} {\bibfnamefont {L.}~\bibnamefont {Ljung}},\ }\href@noop {} {\emph {\bibinfo {title} {{System Identification: Theory for the User}}}}\ (\bibinfo  {publisher} {Pearson},\ \bibinfo {address} {Upper Saddle River, NJ},\ \bibinfo {year} {1999})\BibitemShut {NoStop}%
\bibitem [{\citenamefont {Brunton}\ \emph {et~al.}(2016)\citenamefont {Brunton}, \citenamefont {Proctor},\ and\ \citenamefont {Kutz}}]{brunton_discovering_2016}%
  \BibitemOpen
  \bibfield  {author} {\bibinfo {author} {\bibfnamefont {S.~L.}\ \bibnamefont {Brunton}}, \bibinfo {author} {\bibfnamefont {J.~L.}\ \bibnamefont {Proctor}},\ and\ \bibinfo {author} {\bibfnamefont {J.~N.}\ \bibnamefont {Kutz}},\ }\bibfield  {title} {\bibinfo {title} {Discovering governing equations from data by sparse identification of nonlinear dynamical systems},\ }\href {https://doi.org/10.1073/pnas.1517384113} {\bibfield  {journal} {\bibinfo  {journal} {Proceedings of the National Academy of Sciences}\ }\textbf {\bibinfo {volume} {113}},\ \bibinfo {pages} {3932} (\bibinfo {year} {2016})}\BibitemShut {NoStop}%
\bibitem [{\citenamefont {Iten}\ \emph {et~al.}(2020)\citenamefont {Iten}, \citenamefont {Metger}, \citenamefont {Wilming}, \citenamefont {{del Rio}},\ and\ \citenamefont {Renner}}]{iten_discovering_2020}%
  \BibitemOpen
  \bibfield  {author} {\bibinfo {author} {\bibfnamefont {R.}~\bibnamefont {Iten}}, \bibinfo {author} {\bibfnamefont {T.}~\bibnamefont {Metger}}, \bibinfo {author} {\bibfnamefont {H.}~\bibnamefont {Wilming}}, \bibinfo {author} {\bibfnamefont {L.}~\bibnamefont {{del Rio}}},\ and\ \bibinfo {author} {\bibfnamefont {R.}~\bibnamefont {Renner}},\ }\bibfield  {title} {\bibinfo {title} {Discovering {{Physical Concepts}} with {{Neural Networks}}},\ }\href {https://doi.org/10.1103/PhysRevLett.124.010508} {\bibfield  {journal} {\bibinfo  {journal} {Phys. Rev. Lett.}\ }\textbf {\bibinfo {volume} {124}},\ \bibinfo {pages} {010508} (\bibinfo {year} {2020})}\BibitemShut {NoStop}%
\bibitem [{\citenamefont {Lai}\ \emph {et~al.}(2021)\citenamefont {Lai}, \citenamefont {Mylonas}, \citenamefont {Nagarajaiah},\ and\ \citenamefont {Chatzi}}]{lai_structural_2021}%
  \BibitemOpen
  \bibfield  {author} {\bibinfo {author} {\bibfnamefont {Z.}~\bibnamefont {Lai}}, \bibinfo {author} {\bibfnamefont {C.}~\bibnamefont {Mylonas}}, \bibinfo {author} {\bibfnamefont {S.}~\bibnamefont {Nagarajaiah}},\ and\ \bibinfo {author} {\bibfnamefont {E.}~\bibnamefont {Chatzi}},\ }\bibfield  {title} {\bibinfo {title} {Structural identification with physics-informed neural ordinary differential equations},\ }\href {https://doi.org/10.1016/j.jsv.2021.116196} {\bibfield  {journal} {\bibinfo  {journal} {Journal of Sound and Vibration}\ }\textbf {\bibinfo {volume} {508}},\ \bibinfo {pages} {116196} (\bibinfo {year} {2021})}\BibitemShut {NoStop}%
\bibitem [{\citenamefont {Fronk}\ and\ \citenamefont {Petzold}(2023)}]{fronk_interpretable_2023}%
  \BibitemOpen
  \bibfield  {author} {\bibinfo {author} {\bibfnamefont {C.}~\bibnamefont {Fronk}}\ and\ \bibinfo {author} {\bibfnamefont {L.}~\bibnamefont {Petzold}},\ }\bibfield  {title} {\bibinfo {title} {Interpretable {{Polynomial Neural Ordinary Differential Equations}}},\ }\href {https://doi.org/10.1063/5.0130803} {\bibfield  {journal} {\bibinfo  {journal} {Chaos: An Interdisciplinary Journal of Nonlinear Science}\ }\textbf {\bibinfo {volume} {33}},\ \bibinfo {pages} {043101} (\bibinfo {year} {2023})},\ \Eprint {https://arxiv.org/abs/2208.05072} {arXiv:2208.05072 [cs]} \BibitemShut {NoStop}%
\bibitem [{\citenamefont {Churchill}\ and\ \citenamefont {Xiu}(2023)}]{churchill_flow_2023}%
  \BibitemOpen
  \bibfield  {author} {\bibinfo {author} {\bibfnamefont {V.}~\bibnamefont {Churchill}}\ and\ \bibinfo {author} {\bibfnamefont {D.}~\bibnamefont {Xiu}},\ }\href {https://doi.org/10.48550/arXiv.2307.11013} {\bibinfo {title} {Flow {{Map Learning}} for {{Unknown Dynamical Systems}}: {{Overview}}, {{Implementation}}, and {{Benchmarks}}}} (\bibinfo {year} {2023}),\ \Eprint {https://arxiv.org/abs/2307.11013} {arXiv:2307.11013 [cs]} \BibitemShut {NoStop}%
\bibitem [{\citenamefont {Chen}\ and\ \citenamefont {Wu}(2023)}]{chen_deeposg_2023}%
  \BibitemOpen
  \bibfield  {author} {\bibinfo {author} {\bibfnamefont {J.}~\bibnamefont {Chen}}\ and\ \bibinfo {author} {\bibfnamefont {K.}~\bibnamefont {Wu}},\ }\href {https://doi.org/10.48550/arXiv.2302.03358} {\bibinfo {title} {Deep-{{OSG}}: {{Deep Learning}} of {{Operators}} in {{Semigroup}}}} (\bibinfo {year} {2023}),\ \Eprint {https://arxiv.org/abs/2302.03358} {arXiv:2302.03358 [cs]} \BibitemShut {NoStop}%
\bibitem [{\citenamefont {Yu}\ and\ \citenamefont {Wang}(2024)}]{yu_learning_2024}%
  \BibitemOpen
  \bibfield  {author} {\bibinfo {author} {\bibfnamefont {R.}~\bibnamefont {Yu}}\ and\ \bibinfo {author} {\bibfnamefont {R.}~\bibnamefont {Wang}},\ }\bibfield  {title} {\bibinfo {title} {Learning dynamical systems from data: {{An}} introduction to physics-guided deep learning},\ }\href {https://doi.org/10.1073/pnas.2311808121} {\bibfield  {journal} {\bibinfo  {journal} {Proceedings of the National Academy of Sciences}\ }\textbf {\bibinfo {volume} {121}},\ \bibinfo {pages} {e2311808121} (\bibinfo {year} {2024})}\BibitemShut {NoStop}%
\bibitem [{\citenamefont {Koltai}\ and\ \citenamefont {Kunde}(2024)}]{koltai_koopman_2024}%
  \BibitemOpen
  \bibfield  {author} {\bibinfo {author} {\bibfnamefont {P.}~\bibnamefont {Koltai}}\ and\ \bibinfo {author} {\bibfnamefont {P.}~\bibnamefont {Kunde}},\ }\bibfield  {title} {\bibinfo {title} {A {{Koopman}}--{{Takens Theorem}}: {{Linear Least Squares Prediction}} of {{Nonlinear Time Series}}},\ }\href {https://doi.org/10.1007/s00220-024-05004-8} {\bibfield  {journal} {\bibinfo  {journal} {Communications in Mathematical Physics}\ }\textbf {\bibinfo {volume} {405}},\ \bibinfo {pages} {120} (\bibinfo {year} {2024})}\BibitemShut {NoStop}%
\bibitem [{\citenamefont {Zhang}\ and\ \citenamefont {Zuazua}(2024)}]{zhang_quantitative_2024}%
  \BibitemOpen
  \bibfield  {author} {\bibinfo {author} {\bibfnamefont {C.}~\bibnamefont {Zhang}}\ and\ \bibinfo {author} {\bibfnamefont {E.}~\bibnamefont {Zuazua}},\ }\bibfield  {title} {\bibinfo {title} {A quantitative analysis of {{Koopman}} operator methods for system identification and predictions},\ }\href {https://doi.org/10.5802/crmeca.138} {\bibfield  {journal} {\bibinfo  {journal} {Comptes Rendus. M\'ecanique}\ }\textbf {\bibinfo {volume} {351}},\ \bibinfo {pages} {721} (\bibinfo {year} {2024})}\BibitemShut {NoStop}%
\bibitem [{\citenamefont {Wang}\ \emph {et~al.}(2024)\citenamefont {Wang}, \citenamefont {Huang}, \citenamefont {Gong}, \citenamefont {Geng}, \citenamefont {Liu}, \citenamefont {Zhang},\ and\ \citenamefont {Tao}}]{wang_identifiability_2024}%
  \BibitemOpen
  \bibfield  {author} {\bibinfo {author} {\bibfnamefont {Y.}~\bibnamefont {Wang}}, \bibinfo {author} {\bibfnamefont {W.}~\bibnamefont {Huang}}, \bibinfo {author} {\bibfnamefont {M.}~\bibnamefont {Gong}}, \bibinfo {author} {\bibfnamefont {X.}~\bibnamefont {Geng}}, \bibinfo {author} {\bibfnamefont {T.}~\bibnamefont {Liu}}, \bibinfo {author} {\bibfnamefont {K.}~\bibnamefont {Zhang}},\ and\ \bibinfo {author} {\bibfnamefont {D.}~\bibnamefont {Tao}},\ }\href {https://doi.org/10.48550/arXiv.2210.05955} {\bibinfo {title} {Identifiability and {{Asymptotics}} in {{Learning Homogeneous Linear ODE Systems}} from {{Discrete Observations}}}} (\bibinfo {year} {2024}),\ \Eprint {https://arxiv.org/abs/2210.05955} {arXiv:2210.05955 [stat]} \BibitemShut {NoStop}%
\bibitem [{\citenamefont {Sch{\"o}nlieb}\ and\ \citenamefont {Shumaylov}(2025)}]{schonlieb_datadriven_2025}%
  \BibitemOpen
  \bibfield  {author} {\bibinfo {author} {\bibfnamefont {C.-B.}\ \bibnamefont {Sch{\"o}nlieb}}\ and\ \bibinfo {author} {\bibfnamefont {Z.}~\bibnamefont {Shumaylov}},\ }\href {https://doi.org/10.48550/arXiv.2506.11732} {\bibinfo {title} {Data-driven approaches to inverse problems}} (\bibinfo {year} {2025}),\ \Eprint {https://arxiv.org/abs/2506.11732} {arXiv:2506.11732 [math]} \BibitemShut {NoStop}%
\bibitem [{\citenamefont {Shumaylov}\ \emph {et~al.}(2025)\citenamefont {Shumaylov}, \citenamefont {Zaika}, \citenamefont {Scholl}, \citenamefont {Kutyniok}, \citenamefont {Horesh},\ and\ \citenamefont {Sch{\"o}nlieb}}]{shumaylov_when_2025}%
  \BibitemOpen
  \bibfield  {author} {\bibinfo {author} {\bibfnamefont {Z.}~\bibnamefont {Shumaylov}}, \bibinfo {author} {\bibfnamefont {P.}~\bibnamefont {Zaika}}, \bibinfo {author} {\bibfnamefont {P.}~\bibnamefont {Scholl}}, \bibinfo {author} {\bibfnamefont {G.}~\bibnamefont {Kutyniok}}, \bibinfo {author} {\bibfnamefont {L.}~\bibnamefont {Horesh}},\ and\ \bibinfo {author} {\bibfnamefont {C.-B.}\ \bibnamefont {Sch{\"o}nlieb}},\ }\href {https://doi.org/10.48550/arXiv.2511.08860} {\bibinfo {title} {When is a {{System Discoverable}} from {{Data}}? {{Discovery Requires Chaos}}}} (\bibinfo {year} {2025}),\ \Eprint {https://arxiv.org/abs/2511.08860} {arXiv:2511.08860 [math]} \BibitemShut {NoStop}%
\bibitem [{\citenamefont {Parikh}(2026)}]{parikh_why_2026}%
  \BibitemOpen
  \bibfield  {author} {\bibinfo {author} {\bibfnamefont {H.}~\bibnamefont {Parikh}},\ }\href {https://doi.org/10.48550/arXiv.2601.06730} {\bibinfo {title} {Why are there many equally good models? {{An Anatomy}} of the {{Rashomon Effect}}}} (\bibinfo {year} {2026}),\ \Eprint {https://arxiv.org/abs/2601.06730} {arXiv:2601.06730 [cs]} \BibitemShut {NoStop}%
\bibitem [{\citenamefont {Roeder}\ \emph {et~al.}(2021)\citenamefont {Roeder}, \citenamefont {Metz},\ and\ \citenamefont {Kingma}}]{roeder_linear_2021}%
  \BibitemOpen
  \bibfield  {author} {\bibinfo {author} {\bibfnamefont {G.}~\bibnamefont {Roeder}}, \bibinfo {author} {\bibfnamefont {L.}~\bibnamefont {Metz}},\ and\ \bibinfo {author} {\bibfnamefont {D.}~\bibnamefont {Kingma}},\ }\bibfield  {title} {\bibinfo {title} {On {{Linear Identifiability}} of {{Learned Representations}}},\ }in\ \href@noop {} {\emph {\bibinfo {booktitle} {Proceedings of the 38th {{International Conference}} on {{Machine Learning}}}}}\ (\bibinfo  {publisher} {PMLR},\ \bibinfo {year} {2021})\ pp.\ \bibinfo {pages} {9030--9039}\BibitemShut {NoStop}%
\bibitem [{\citenamefont {Gy{\"o}rgyi}(1990)}]{gyorgyi_firstorder_1990}%
  \BibitemOpen
  \bibfield  {author} {\bibinfo {author} {\bibfnamefont {G.}~\bibnamefont {Gy{\"o}rgyi}},\ }\bibfield  {title} {\bibinfo {title} {First-order transition to perfect generalization in a neural network with binary synapses},\ }\href {https://doi.org/10.1103/PhysRevA.41.7097} {\bibfield  {journal} {\bibinfo  {journal} {Phys. Rev. A}\ }\textbf {\bibinfo {volume} {41}},\ \bibinfo {pages} {7097} (\bibinfo {year} {1990})}\BibitemShut {NoStop}%
\bibitem [{\citenamefont {Seung}(1992)}]{seung_statistical_1992}%
  \BibitemOpen
  \bibfield  {author} {\bibinfo {author} {\bibfnamefont {H.~S.}\ \bibnamefont {Seung}},\ }\bibfield  {title} {\bibinfo {title} {Statistical mechanics of learning from examples},\ }\href {https://doi.org/10.1103/PhysRevA.45.6056} {\bibfield  {journal} {\bibinfo  {journal} {Phys. Rev. A}\ }\textbf {\bibinfo {volume} {45}},\ \bibinfo {pages} {6056} (\bibinfo {year} {1992})}\BibitemShut {NoStop}%
\bibitem [{\citenamefont {Watkin}(1993)}]{watkin_statistical_1993}%
  \BibitemOpen
  \bibfield  {author} {\bibinfo {author} {\bibfnamefont {T.~L.~H.}\ \bibnamefont {Watkin}},\ }\bibfield  {title} {\bibinfo {title} {The statistical mechanics of learning a rule},\ }\href {https://doi.org/10.1103/RevModPhys.65.499} {\bibfield  {journal} {\bibinfo  {journal} {Reviews of Modern Physics}\ }\textbf {\bibinfo {volume} {65}},\ \bibinfo {pages} {499} (\bibinfo {year} {1993})}\BibitemShut {NoStop}%
\bibitem [{\citenamefont {Hess}\ \emph {et~al.}(2023)\citenamefont {Hess}, \citenamefont {Monfared}, \citenamefont {Brenner},\ and\ \citenamefont {Durstewitz}}]{hess_generalized_2023}%
  \BibitemOpen
  \bibfield  {author} {\bibinfo {author} {\bibfnamefont {F.}~\bibnamefont {Hess}}, \bibinfo {author} {\bibfnamefont {Z.}~\bibnamefont {Monfared}}, \bibinfo {author} {\bibfnamefont {M.}~\bibnamefont {Brenner}},\ and\ \bibinfo {author} {\bibfnamefont {D.}~\bibnamefont {Durstewitz}},\ }\href {https://doi.org/10.48550/arXiv.2306.04406} {\bibinfo {title} {Generalized {{Teacher Forcing}} for {{Learning Chaotic Dynamics}}}} (\bibinfo {year} {2023}),\ \Eprint {https://arxiv.org/abs/2306.04406} {arXiv:2306.04406 [cs]} \BibitemShut {NoStop}%
\bibitem [{\citenamefont {Karniadakis}\ \emph {et~al.}(2021)\citenamefont {Karniadakis}, \citenamefont {Kevrekidis}, \citenamefont {Lu}, \citenamefont {Perdikaris}, \citenamefont {Wang},\ and\ \citenamefont {Yang}}]{karniadakis_physicsinformed_2021}%
  \BibitemOpen
  \bibfield  {author} {\bibinfo {author} {\bibfnamefont {G.~E.}\ \bibnamefont {Karniadakis}}, \bibinfo {author} {\bibfnamefont {I.~G.}\ \bibnamefont {Kevrekidis}}, \bibinfo {author} {\bibfnamefont {L.}~\bibnamefont {Lu}}, \bibinfo {author} {\bibfnamefont {P.}~\bibnamefont {Perdikaris}}, \bibinfo {author} {\bibfnamefont {S.}~\bibnamefont {Wang}},\ and\ \bibinfo {author} {\bibfnamefont {L.}~\bibnamefont {Yang}},\ }\bibfield  {title} {\bibinfo {title} {Physics-informed machine learning},\ }\href {https://doi.org/10.1038/s42254-021-00314-5} {\bibfield  {journal} {\bibinfo  {journal} {Nature Reviews Physics}\ }\textbf {\bibinfo {volume} {3}},\ \bibinfo {pages} {422} (\bibinfo {year} {2021})}\BibitemShut {NoStop}%
\bibitem [{\citenamefont {Adler}\ \emph {et~al.}(2024)\citenamefont {Adler}, \citenamefont {Hocking}, \citenamefont {Hu},\ and\ \citenamefont {Islam}}]{adler_physicsinformed_2024}%
  \BibitemOpen
  \bibfield  {author} {\bibinfo {author} {\bibfnamefont {J.~H.}\ \bibnamefont {Adler}}, \bibinfo {author} {\bibfnamefont {S.}~\bibnamefont {Hocking}}, \bibinfo {author} {\bibfnamefont {X.}~\bibnamefont {Hu}},\ and\ \bibinfo {author} {\bibfnamefont {S.}~\bibnamefont {Islam}},\ }\href {https://doi.org/10.48550/arXiv.2407.18057} {\bibinfo {title} {Physics-informed nonlinear vector autoregressive models for the prediction of dynamical systems}} (\bibinfo {year} {2024}),\ \Eprint {https://arxiv.org/abs/2407.18057} {arXiv:2407.18057 [math]} \BibitemShut {NoStop}%
\bibitem [{\citenamefont {Messenger}\ and\ \citenamefont {Bortz}(2021)}]{messenger_weak_2021}%
  \BibitemOpen
  \bibfield  {author} {\bibinfo {author} {\bibfnamefont {D.~A.}\ \bibnamefont {Messenger}}\ and\ \bibinfo {author} {\bibfnamefont {D.~M.}\ \bibnamefont {Bortz}},\ }\bibfield  {title} {\bibinfo {title} {Weak {{SINDy}}: {{Galerkin-Based Data-Driven Model Selection}}},\ }\href {https://doi.org/10.1137/20M1343166} {\bibfield  {journal} {\bibinfo  {journal} {Multiscale Modeling \& Simulation}\ }\textbf {\bibinfo {volume} {19}},\ \bibinfo {pages} {1474} (\bibinfo {year} {2021})},\ \Eprint {https://arxiv.org/abs/2005.04339} {arXiv:2005.04339 [math]} \BibitemShut {NoStop}%
\bibitem [{\citenamefont {Russo}\ and\ \citenamefont {Laiu}(2024)}]{russo_convergence_2024}%
  \BibitemOpen
  \bibfield  {author} {\bibinfo {author} {\bibfnamefont {B.}~\bibnamefont {Russo}}\ and\ \bibinfo {author} {\bibfnamefont {M.~P.}\ \bibnamefont {Laiu}},\ }\href {https://doi.org/10.48550/arXiv.2209.15573} {\bibinfo {title} {Convergence of weak-{{SINDy Surrogate Models}}}} (\bibinfo {year} {2024}),\ \Eprint {https://arxiv.org/abs/2209.15573} {arXiv:2209.15573 [math]} \BibitemShut {NoStop}%
\bibitem [{\citenamefont {Pecile}\ \emph {et~al.}(2025)\citenamefont {Pecile}, \citenamefont {Demo}, \citenamefont {Tezzele}, \citenamefont {Rozza},\ and\ \citenamefont {Breda}}]{pecile_datadriven_2025}%
  \BibitemOpen
  \bibfield  {author} {\bibinfo {author} {\bibfnamefont {A.}~\bibnamefont {Pecile}}, \bibinfo {author} {\bibfnamefont {N.}~\bibnamefont {Demo}}, \bibinfo {author} {\bibfnamefont {M.}~\bibnamefont {Tezzele}}, \bibinfo {author} {\bibfnamefont {G.}~\bibnamefont {Rozza}},\ and\ \bibinfo {author} {\bibfnamefont {D.}~\bibnamefont {Breda}},\ }\bibfield  {title} {\bibinfo {title} {Data-driven discovery of delay differential equations with discrete delays},\ }\href {https://doi.org/10.1016/j.cam.2024.116439} {\bibfield  {journal} {\bibinfo  {journal} {Journal of Computational and Applied Mathematics}\ }\textbf {\bibinfo {volume} {461}},\ \bibinfo {pages} {116439} (\bibinfo {year} {2025})}\BibitemShut {NoStop}%
\bibitem [{\citenamefont {Zhang}\ and\ \citenamefont {Cornelius}(2023)}]{zhang_catch22s_2023}%
  \BibitemOpen
  \bibfield  {author} {\bibinfo {author} {\bibfnamefont {Y.}~\bibnamefont {Zhang}}\ and\ \bibinfo {author} {\bibfnamefont {S.~P.}\ \bibnamefont {Cornelius}},\ }\bibfield  {title} {\bibinfo {title} {Catch-22s of reservoir computing},\ }\href {https://doi.org/10.1103/PhysRevResearch.5.033213} {\bibfield  {journal} {\bibinfo  {journal} {Phys. Rev. Research}\ }\textbf {\bibinfo {volume} {5}},\ \bibinfo {pages} {033213} (\bibinfo {year} {2023})}\BibitemShut {NoStop}%
\bibitem [{\citenamefont {Zhang}\ \emph {et~al.}(2025)\citenamefont {Zhang}, \citenamefont {Santos},\ and\ \citenamefont {Cornelius}}]{zhang_how_2025}%
  \BibitemOpen
  \bibfield  {author} {\bibinfo {author} {\bibfnamefont {Y.}~\bibnamefont {Zhang}}, \bibinfo {author} {\bibfnamefont {E.~R.}\ \bibnamefont {Santos}},\ and\ \bibinfo {author} {\bibfnamefont {S.~P.}\ \bibnamefont {Cornelius}},\ }\href {https://doi.org/10.48550/arXiv.2407.08641} {\bibinfo {title} {How more data can hurt: {{Instability}} and regularization in next-generation reservoir computing}} (\bibinfo {year} {2025}),\ \Eprint {https://arxiv.org/abs/2407.08641} {arXiv:2407.08641 [cs]} \BibitemShut {NoStop}%
\bibitem [{\citenamefont {dos Santos}\ and\ \citenamefont {Bollt}(2025)}]{santos_emergence_2025}%
  \BibitemOpen
  \bibfield  {author} {\bibinfo {author} {\bibfnamefont {E.~R.}\ \bibnamefont {dos Santos}}\ and\ \bibinfo {author} {\bibfnamefont {E.}~\bibnamefont {Bollt}},\ }\href {https://doi.org/10.48550/arXiv.2505.00846} {\bibinfo {title} {On the emergence of numerical instabilities in {{Next Generation Reservoir Computing}}}} (\bibinfo {year} {2025}),\ \Eprint {https://arxiv.org/abs/2505.00846} {arXiv:2505.00846 [stat]} \BibitemShut {NoStop}%
\bibitem [{\citenamefont {Liu}\ \emph {et~al.}(2023)\citenamefont {Liu}, \citenamefont {Xiao}, \citenamefont {Yan},\ and\ \citenamefont {Gao}}]{liu_noise_2023}%
  \BibitemOpen
  \bibfield  {author} {\bibinfo {author} {\bibfnamefont {S.}~\bibnamefont {Liu}}, \bibinfo {author} {\bibfnamefont {J.}~\bibnamefont {Xiao}}, \bibinfo {author} {\bibfnamefont {Z.}~\bibnamefont {Yan}},\ and\ \bibinfo {author} {\bibfnamefont {J.}~\bibnamefont {Gao}},\ }\bibfield  {title} {\bibinfo {title} {Noise resistance of next-generation reservoir computing: A comparative study with high-order correlation computation},\ }\href {https://doi.org/10.1007/s11071-023-08592-7} {\bibfield  {journal} {\bibinfo  {journal} {Nonlinear Dynamics}\ }\textbf {\bibinfo {volume} {111}},\ \bibinfo {pages} {14295} (\bibinfo {year} {2023})}\BibitemShut {NoStop}%
\bibitem [{\citenamefont {Chen}\ \emph {et~al.}(2022)\citenamefont {Chen}, \citenamefont {Penny}, \citenamefont {Smith},\ and\ \citenamefont {Platt}}]{chen_next_2022}%
  \BibitemOpen
  \bibfield  {author} {\bibinfo {author} {\bibfnamefont {T.-C.}\ \bibnamefont {Chen}}, \bibinfo {author} {\bibfnamefont {S.~G.}\ \bibnamefont {Penny}}, \bibinfo {author} {\bibfnamefont {T.~A.}\ \bibnamefont {Smith}},\ and\ \bibinfo {author} {\bibfnamefont {J.~A.}\ \bibnamefont {Platt}},\ }\href {https://doi.org/10.48550/arXiv.2201.05193} {\bibinfo {title} {`{{Next Generation}}' {{Reservoir Computing}}: An {{Empirical Data-Driven Expression}} of {{Dynamical Equations}} in {{Time-Stepping Form}}}} (\bibinfo {year} {2022}),\ \Eprint {https://arxiv.org/abs/2201.05193} {arXiv:2201.05193 [cs]} \BibitemShut {NoStop}%
\bibitem [{\citenamefont {Kay}\ and\ \citenamefont {Marple}(1981)}]{kay_spectrum_1981}%
  \BibitemOpen
  \bibfield  {author} {\bibinfo {author} {\bibfnamefont {S.}~\bibnamefont {Kay}}\ and\ \bibinfo {author} {\bibfnamefont {S.}~\bibnamefont {Marple}},\ }\bibfield  {title} {\bibinfo {title} {Spectrum analysis---{{A}} modern perspective},\ }\href {https://doi.org/10.1109/PROC.1981.12184} {\bibfield  {journal} {\bibinfo  {journal} {Proceedings of the IEEE}\ }\textbf {\bibinfo {volume} {69}},\ \bibinfo {pages} {1380} (\bibinfo {year} {1981})}\BibitemShut {NoStop}%
\bibitem [{\citenamefont {R{\"a}th}\ \emph {et~al.}(2012)\citenamefont {R{\"a}th}, \citenamefont {Gliozzi}, \citenamefont {Papadakis},\ and\ \citenamefont {Brinkmann}}]{rath_revisiting_2012}%
  \BibitemOpen
  \bibfield  {author} {\bibinfo {author} {\bibfnamefont {C.}~\bibnamefont {R{\"a}th}}, \bibinfo {author} {\bibfnamefont {M.}~\bibnamefont {Gliozzi}}, \bibinfo {author} {\bibfnamefont {I.~E.}\ \bibnamefont {Papadakis}},\ and\ \bibinfo {author} {\bibfnamefont {W.}~\bibnamefont {Brinkmann}},\ }\bibfield  {title} {\bibinfo {title} {Revisiting {{Algorithms}} for {{Generating Surrogate Time Series}}},\ }\href {https://doi.org/10.1103/PhysRevLett.109.144101} {\bibfield  {journal} {\bibinfo  {journal} {Phys. Rev. Lett.}\ }\textbf {\bibinfo {volume} {109}},\ \bibinfo {pages} {144101} (\bibinfo {year} {2012})}\BibitemShut {NoStop}%
\bibitem [{\citenamefont {So}\ \emph {et~al.}(2005)\citenamefont {So}, \citenamefont {Chan}, \citenamefont {Chan},\ and\ \citenamefont {Ho}}]{so_linear_2005}%
  \BibitemOpen
  \bibfield  {author} {\bibinfo {author} {\bibfnamefont {H.}~\bibnamefont {So}}, \bibinfo {author} {\bibfnamefont {K.~W.}\ \bibnamefont {Chan}}, \bibinfo {author} {\bibfnamefont {Y.}~\bibnamefont {Chan}},\ and\ \bibinfo {author} {\bibfnamefont {K.}~\bibnamefont {Ho}},\ }\bibfield  {title} {\bibinfo {title} {Linear prediction approach for efficient frequency estimation of multiple real sinusoids: Algorithms and analyses},\ }\href {https://doi.org/10.1109/TSP.2005.849154} {\bibfield  {journal} {\bibinfo  {journal} {IEEE Transactions on Signal Processing}\ }\textbf {\bibinfo {volume} {53}},\ \bibinfo {pages} {2290} (\bibinfo {year} {2005})}\BibitemShut {NoStop}%
\bibitem [{\citenamefont {Vaseghi}(2008)}]{vaseghi_advanced_2008}%
  \BibitemOpen
  \bibfield  {author} {\bibinfo {author} {\bibfnamefont {S.~V.}\ \bibnamefont {Vaseghi}},\ }\href@noop {} {\emph {\bibinfo {title} {Advanced {{Digital Signal Processing}} and {{Noise Reduction}}}}}\ (\bibinfo  {publisher} {Wiley},\ \bibinfo {address} {Chichester, U.K},\ \bibinfo {year} {2008})\BibitemShut {NoStop}%
\bibitem [{\citenamefont {Pan}\ and\ \citenamefont {Duraisamy}(2020)}]{pan_structure_2020}%
  \BibitemOpen
  \bibfield  {author} {\bibinfo {author} {\bibfnamefont {S.}~\bibnamefont {Pan}}\ and\ \bibinfo {author} {\bibfnamefont {K.}~\bibnamefont {Duraisamy}},\ }\bibfield  {title} {\bibinfo {title} {On the {{Structure}} of {{Time-delay Embedding}} in {{Linear Models}} of {{Non-linear Dynamical Systems}}},\ }\href {https://doi.org/10.1063/5.0010886} {\bibfield  {journal} {\bibinfo  {journal} {Chaos: An Interdisciplinary Journal of Nonlinear Science}\ }\textbf {\bibinfo {volume} {30}},\ \bibinfo {pages} {073135} (\bibinfo {year} {2020})},\ \Eprint {https://arxiv.org/abs/1902.05198} {arXiv:1902.05198 [math]} \BibitemShut {NoStop}%
\bibitem [{\citenamefont {Piroddi}(2008)}]{piroddi_simulation_2008}%
  \BibitemOpen
  \bibfield  {author} {\bibinfo {author} {\bibfnamefont {L.}~\bibnamefont {Piroddi}},\ }\bibfield  {title} {\bibinfo {title} {Simulation error minimisation methods for {{NARX}} model identification},\ }\href {https://doi.org/10.1504/IJMIC.2008.020548} {\bibfield  {journal} {\bibinfo  {journal} {International Journal of Modelling, Identification and Control}\ }\textbf {\bibinfo {volume} {3}},\ \bibinfo {pages} {392} (\bibinfo {year} {2008})}\BibitemShut {NoStop}%
\bibitem [{\citenamefont {Aguirre}\ \emph {et~al.}(2010)\citenamefont {Aguirre}, \citenamefont {Barbosa},\ and\ \citenamefont {Braga}}]{aguirre_prediction_2010}%
  \BibitemOpen
  \bibfield  {author} {\bibinfo {author} {\bibfnamefont {L.~A.}\ \bibnamefont {Aguirre}}, \bibinfo {author} {\bibfnamefont {B.~H.~G.}\ \bibnamefont {Barbosa}},\ and\ \bibinfo {author} {\bibfnamefont {A.~P.}\ \bibnamefont {Braga}},\ }\bibfield  {title} {\bibinfo {title} {Prediction and simulation errors in parameter estimation for nonlinear systems},\ }\href {https://doi.org/10.1016/j.ymssp.2010.05.003} {\bibfield  {journal} {\bibinfo  {journal} {Mechanical Systems and Signal Processing}\ }\textbf {\bibinfo {volume} {24}},\ \bibinfo {pages} {2855} (\bibinfo {year} {2010})}\BibitemShut {NoStop}%
\bibitem [{\citenamefont {Sch{\"a}r}\ \emph {et~al.}(2025)\citenamefont {Sch{\"a}r}, \citenamefont {Marelli},\ and\ \citenamefont {Sudret}}]{schar_surrogate_2025}%
  \BibitemOpen
  \bibfield  {author} {\bibinfo {author} {\bibfnamefont {S.}~\bibnamefont {Sch{\"a}r}}, \bibinfo {author} {\bibfnamefont {S.}~\bibnamefont {Marelli}},\ and\ \bibinfo {author} {\bibfnamefont {B.}~\bibnamefont {Sudret}},\ }\bibfield  {title} {\bibinfo {title} {Surrogate modeling with functional nonlinear autoregressive models ({{F-NARX}})},\ }\href {https://doi.org/10.1016/j.ress.2025.111276} {\bibfield  {journal} {\bibinfo  {journal} {Reliability Engineering \& System Safety}\ }\textbf {\bibinfo {volume} {264}},\ \bibinfo {pages} {111276} (\bibinfo {year} {2025})},\ \Eprint {https://arxiv.org/abs/2410.07293} {arXiv:2410.07293 [stat]} \BibitemShut {NoStop}%
\bibitem [{\citenamefont {Bengio}\ \emph {et~al.}(2015)\citenamefont {Bengio}, \citenamefont {Vinyals}, \citenamefont {Jaitly},\ and\ \citenamefont {Shazeer}}]{bengio_scheduled_2015}%
  \BibitemOpen
  \bibfield  {author} {\bibinfo {author} {\bibfnamefont {S.}~\bibnamefont {Bengio}}, \bibinfo {author} {\bibfnamefont {O.}~\bibnamefont {Vinyals}}, \bibinfo {author} {\bibfnamefont {N.}~\bibnamefont {Jaitly}},\ and\ \bibinfo {author} {\bibfnamefont {N.}~\bibnamefont {Shazeer}},\ }\href {https://doi.org/10.48550/arXiv.1506.03099} {\bibinfo {title} {Scheduled {{Sampling}} for {{Sequence Prediction}} with {{Recurrent Neural Networks}}}} (\bibinfo {year} {2015}),\ \Eprint {https://arxiv.org/abs/1506.03099} {arXiv:1506.03099 [cs]} \BibitemShut {NoStop}%
\bibitem [{\citenamefont {Husz{\'a}r}(2015)}]{huszar_how_2015}%
  \BibitemOpen
  \bibfield  {author} {\bibinfo {author} {\bibfnamefont {F.}~\bibnamefont {Husz{\'a}r}},\ }\href {https://doi.org/10.48550/arXiv.1511.05101} {\bibinfo {title} {How (not) to {{Train}} your {{Generative Model}}: {{Scheduled Sampling}}, {{Likelihood}}, {{Adversary}}?}} (\bibinfo {year} {2015}),\ \Eprint {https://arxiv.org/abs/1511.05101} {arXiv:1511.05101 [stat]} \BibitemShut {NoStop}%
\bibitem [{\citenamefont {Schmidt}(2019)}]{schmidt_generalization_2019}%
  \BibitemOpen
  \bibfield  {author} {\bibinfo {author} {\bibfnamefont {F.}~\bibnamefont {Schmidt}},\ }\href {https://doi.org/10.48550/arXiv.1910.00292} {\bibinfo {title} {Generalization in {{Generation}}: {{A}} closer look at {{Exposure Bias}}}} (\bibinfo {year} {2019}),\ \Eprint {https://arxiv.org/abs/1910.00292} {arXiv:1910.00292 [cs]} \BibitemShut {NoStop}%
\bibitem [{\citenamefont {Somalwar}\ \emph {et~al.}(2025)\citenamefont {Somalwar}, \citenamefont {Lee}, \citenamefont {Pappas},\ and\ \citenamefont {Matni}}]{somalwar_learning_2025}%
  \BibitemOpen
  \bibfield  {author} {\bibinfo {author} {\bibfnamefont {A.}~\bibnamefont {Somalwar}}, \bibinfo {author} {\bibfnamefont {B.~D.}\ \bibnamefont {Lee}}, \bibinfo {author} {\bibfnamefont {G.~J.}\ \bibnamefont {Pappas}},\ and\ \bibinfo {author} {\bibfnamefont {N.}~\bibnamefont {Matni}},\ }\href {https://doi.org/10.48550/arXiv.2504.01766} {\bibinfo {title} {Learning with {{Imperfect Models}}: {{When Multi-step Prediction Mitigates Compounding Error}}}} (\bibinfo {year} {2025}),\ \Eprint {https://arxiv.org/abs/2504.01766} {arXiv:2504.01766 [eess]} \BibitemShut {NoStop}%
\bibitem [{\citenamefont {Haseli}\ and\ \citenamefont {Cort{\'e}s}(2022)}]{haseli_temporal_2022}%
  \BibitemOpen
  \bibfield  {author} {\bibinfo {author} {\bibfnamefont {M.}~\bibnamefont {Haseli}}\ and\ \bibinfo {author} {\bibfnamefont {J.}~\bibnamefont {Cort{\'e}s}},\ }\href {https://doi.org/10.48550/arXiv.2207.07719} {\bibinfo {title} {Temporal {{Forward-Backward Consistency}}, {{Not Residual Error}}, {{Measures}} the {{Prediction Accuracy}} of {{Extended Dynamic Mode Decomposition}}}} (\bibinfo {year} {2022}),\ \Eprint {https://arxiv.org/abs/2207.07719} {arXiv:2207.07719 [eess]} \BibitemShut {NoStop}%
\bibitem [{\citenamefont {Haseli}\ and\ \citenamefont {Cort{\'e}s}(2023)}]{haseli_generalizing_2023}%
  \BibitemOpen
  \bibfield  {author} {\bibinfo {author} {\bibfnamefont {M.}~\bibnamefont {Haseli}}\ and\ \bibinfo {author} {\bibfnamefont {J.}~\bibnamefont {Cort{\'e}s}},\ }\bibfield  {title} {\bibinfo {title} {Generalizing dynamic mode decomposition: {{Balancing}} accuracy and expressiveness in {{Koopman}} approximations},\ }\href {https://doi.org/10.1016/j.automatica.2023.111001} {\bibfield  {journal} {\bibinfo  {journal} {Automatica}\ }\textbf {\bibinfo {volume} {153}},\ \bibinfo {pages} {111001} (\bibinfo {year} {2023})}\BibitemShut {NoStop}%
\bibitem [{\citenamefont {Haseli}\ and\ \citenamefont {Cort{\'e}s}(2025)}]{haseli_invariance_2025}%
  \BibitemOpen
  \bibfield  {author} {\bibinfo {author} {\bibfnamefont {M.}~\bibnamefont {Haseli}}\ and\ \bibinfo {author} {\bibfnamefont {J.}~\bibnamefont {Cort{\'e}s}},\ }\href {https://doi.org/10.48550/arXiv.2311.13033} {\bibinfo {title} {Invariance {{Proximity}}: {{Closed-Form Error Bounds}} for {{Finite-Dimensional Koopman-Based Models}}}} (\bibinfo {year} {2025}),\ \Eprint {https://arxiv.org/abs/2311.13033} {arXiv:2311.13033 [math]} \BibitemShut {NoStop}%
\bibitem [{\citenamefont {Forets}\ and\ \citenamefont {Pouly}(2017)}]{forets_explicit_2017}%
  \BibitemOpen
  \bibfield  {author} {\bibinfo {author} {\bibfnamefont {M.}~\bibnamefont {Forets}}\ and\ \bibinfo {author} {\bibfnamefont {A.}~\bibnamefont {Pouly}},\ }\href {https://doi.org/10.48550/arXiv.1711.02552} {\bibinfo {title} {Explicit {{Error Bounds}} for {{Carleman Linearization}}}} (\bibinfo {year} {2017}),\ \Eprint {https://arxiv.org/abs/1711.02552} {arXiv:1711.02552 [math]} \BibitemShut {NoStop}%
\bibitem [{\citenamefont {Klus}\ \emph {et~al.}(2020)\citenamefont {Klus}, \citenamefont {N{\"u}ske}, \citenamefont {Peitz}, \citenamefont {Niemann}, \citenamefont {Clementi},\ and\ \citenamefont {Sch{\"u}tte}}]{klus_datadriven_2020}%
  \BibitemOpen
  \bibfield  {author} {\bibinfo {author} {\bibfnamefont {S.}~\bibnamefont {Klus}}, \bibinfo {author} {\bibfnamefont {F.}~\bibnamefont {N{\"u}ske}}, \bibinfo {author} {\bibfnamefont {S.}~\bibnamefont {Peitz}}, \bibinfo {author} {\bibfnamefont {J.-H.}\ \bibnamefont {Niemann}}, \bibinfo {author} {\bibfnamefont {C.}~\bibnamefont {Clementi}},\ and\ \bibinfo {author} {\bibfnamefont {C.}~\bibnamefont {Sch{\"u}tte}},\ }\bibfield  {title} {\bibinfo {title} {Data-driven approximation of the {{Koopman}} generator: {{Model}} reduction, system identification, and control},\ }\href {https://doi.org/10.1016/j.physd.2020.132416} {\bibfield  {journal} {\bibinfo  {journal} {Physica D: Nonlinear Phenomena}\ }\textbf {\bibinfo {volume} {406}},\ \bibinfo {pages} {132416} (\bibinfo {year} {2020})}\BibitemShut {NoStop}%
\bibitem [{\citenamefont {Kostic}\ \emph {et~al.}(2022)\citenamefont {Kostic}, \citenamefont {Novelli}, \citenamefont {Maurer}, \citenamefont {Ciliberto}, \citenamefont {Rosasco},\ and\ \citenamefont {Pontil}}]{kostic_learning_2022}%
  \BibitemOpen
  \bibfield  {author} {\bibinfo {author} {\bibfnamefont {V.}~\bibnamefont {Kostic}}, \bibinfo {author} {\bibfnamefont {P.}~\bibnamefont {Novelli}}, \bibinfo {author} {\bibfnamefont {A.}~\bibnamefont {Maurer}}, \bibinfo {author} {\bibfnamefont {C.}~\bibnamefont {Ciliberto}}, \bibinfo {author} {\bibfnamefont {L.}~\bibnamefont {Rosasco}},\ and\ \bibinfo {author} {\bibfnamefont {M.}~\bibnamefont {Pontil}},\ }\href {https://doi.org/10.48550/arXiv.2205.14027} {\bibinfo {title} {Learning {{Dynamical Systems}} via {{Koopman Operator Regression}} in {{Reproducing Kernel Hilbert Spaces}}}} (\bibinfo {year} {2022}),\ \Eprint {https://arxiv.org/abs/2205.14027} {arXiv:2205.14027 [cs]} \BibitemShut {NoStop}%
\bibitem [{\citenamefont {Amini}\ \emph {et~al.}(2022)\citenamefont {Amini}, \citenamefont {Zheng}, \citenamefont {Sun},\ and\ \citenamefont {Motee}}]{amini_carleman_2022}%
  \BibitemOpen
  \bibfield  {author} {\bibinfo {author} {\bibfnamefont {A.}~\bibnamefont {Amini}}, \bibinfo {author} {\bibfnamefont {C.}~\bibnamefont {Zheng}}, \bibinfo {author} {\bibfnamefont {Q.}~\bibnamefont {Sun}},\ and\ \bibinfo {author} {\bibfnamefont {N.}~\bibnamefont {Motee}},\ }\href {https://doi.org/10.48550/arXiv.2207.07755} {\bibinfo {title} {Carleman {{Linearization}} of {{Nonlinear Systems}} and {{Its Finite-Section Approximations}}}} (\bibinfo {year} {2022}),\ \Eprint {https://arxiv.org/abs/2207.07755} {arXiv:2207.07755 [math]} \BibitemShut {NoStop}%
\bibitem [{\citenamefont {N{\"u}ske}\ \emph {et~al.}(2022)\citenamefont {N{\"u}ske}, \citenamefont {Peitz}, \citenamefont {Philipp}, \citenamefont {Schaller},\ and\ \citenamefont {Worthmann}}]{nuske_finitedata_2022}%
  \BibitemOpen
  \bibfield  {author} {\bibinfo {author} {\bibfnamefont {F.}~\bibnamefont {N{\"u}ske}}, \bibinfo {author} {\bibfnamefont {S.}~\bibnamefont {Peitz}}, \bibinfo {author} {\bibfnamefont {F.}~\bibnamefont {Philipp}}, \bibinfo {author} {\bibfnamefont {M.}~\bibnamefont {Schaller}},\ and\ \bibinfo {author} {\bibfnamefont {K.}~\bibnamefont {Worthmann}},\ }\bibfield  {title} {\bibinfo {title} {Finite-{{Data Error Bounds}} for {{Koopman-Based Prediction}} and {{Control}}},\ }\href {https://doi.org/10.1007/s00332-022-09862-1} {\bibfield  {journal} {\bibinfo  {journal} {Journal of Nonlinear Science}\ }\textbf {\bibinfo {volume} {33}},\ \bibinfo {pages} {14} (\bibinfo {year} {2022})}\BibitemShut {NoStop}%
\bibitem [{\citenamefont {Philipp}\ \emph {et~al.}(2024)\citenamefont {Philipp}, \citenamefont {Schaller}, \citenamefont {Boshoff}, \citenamefont {Peitz}, \citenamefont {N{\"u}ske},\ and\ \citenamefont {Worthmann}}]{philipp_variance_2024}%
  \BibitemOpen
  \bibfield  {author} {\bibinfo {author} {\bibfnamefont {F.~M.}\ \bibnamefont {Philipp}}, \bibinfo {author} {\bibfnamefont {M.}~\bibnamefont {Schaller}}, \bibinfo {author} {\bibfnamefont {S.}~\bibnamefont {Boshoff}}, \bibinfo {author} {\bibfnamefont {S.}~\bibnamefont {Peitz}}, \bibinfo {author} {\bibfnamefont {F.}~\bibnamefont {N{\"u}ske}},\ and\ \bibinfo {author} {\bibfnamefont {K.}~\bibnamefont {Worthmann}},\ }\href {https://doi.org/10.48550/arXiv.2402.02494} {\bibinfo {title} {Variance representations and convergence rates for data-driven approximations of {{Koopman}} operators}} (\bibinfo {year} {2024}),\ \Eprint {https://arxiv.org/abs/2402.02494} {arXiv:2402.02494 [math]} \BibitemShut {NoStop}%
\bibitem [{\citenamefont {Kamb}\ \emph {et~al.}(2020)\citenamefont {Kamb}, \citenamefont {Kaiser}, \citenamefont {Brunton},\ and\ \citenamefont {Kutz}}]{kamb_timedelay_2020}%
  \BibitemOpen
  \bibfield  {author} {\bibinfo {author} {\bibfnamefont {M.}~\bibnamefont {Kamb}}, \bibinfo {author} {\bibfnamefont {E.}~\bibnamefont {Kaiser}}, \bibinfo {author} {\bibfnamefont {S.~L.}\ \bibnamefont {Brunton}},\ and\ \bibinfo {author} {\bibfnamefont {J.~N.}\ \bibnamefont {Kutz}},\ }\href {https://doi.org/10.48550/arXiv.1810.01479} {\bibinfo {title} {Time-{{Delay Observables}} for {{Koopman}}: {{Theory}} and {{Applications}}}} (\bibinfo {year} {2020}),\ \Eprint {https://arxiv.org/abs/1810.01479} {arXiv:1810.01479 [math]} \BibitemShut {NoStop}%
\bibitem [{\citenamefont {Iacob}\ \emph {et~al.}(2023)\citenamefont {Iacob}, \citenamefont {Schoukens},\ and\ \citenamefont {T{\'o}th}}]{iacob_finite_2023}%
  \BibitemOpen
  \bibfield  {author} {\bibinfo {author} {\bibfnamefont {L.~C.}\ \bibnamefont {Iacob}}, \bibinfo {author} {\bibfnamefont {M.}~\bibnamefont {Schoukens}},\ and\ \bibinfo {author} {\bibfnamefont {R.}~\bibnamefont {T{\'o}th}},\ }\href {https://doi.org/10.1016/j.ifacol.2023.10.849} {\bibinfo {title} {Finite {{Dimensional Koopman Form}} of {{Polynomial Nonlinear Systems}}}} (\bibinfo {year} {2023})\BibitemShut {NoStop}%
\bibitem [{\citenamefont {Raissi}\ \emph {et~al.}(2018)\citenamefont {Raissi}, \citenamefont {Perdikaris},\ and\ \citenamefont {Karniadakis}}]{raissi_multistep_2018}%
  \BibitemOpen
  \bibfield  {author} {\bibinfo {author} {\bibfnamefont {M.}~\bibnamefont {Raissi}}, \bibinfo {author} {\bibfnamefont {P.}~\bibnamefont {Perdikaris}},\ and\ \bibinfo {author} {\bibfnamefont {G.~E.}\ \bibnamefont {Karniadakis}},\ }\href {https://doi.org/10.48550/arXiv.1801.01236} {\bibinfo {title} {Multistep {{Neural Networks}} for {{Data-driven Discovery}} of {{Nonlinear Dynamical Systems}}}} (\bibinfo {year} {2018}),\ \Eprint {https://arxiv.org/abs/1801.01236} {arXiv:1801.01236 [math]} \BibitemShut {NoStop}%
\bibitem [{\citenamefont {Du}\ \emph {et~al.}(2022)\citenamefont {Du}, \citenamefont {Gu}, \citenamefont {Yang},\ and\ \citenamefont {Zhou}}]{du_discovery_2022}%
  \BibitemOpen
  \bibfield  {author} {\bibinfo {author} {\bibfnamefont {Q.}~\bibnamefont {Du}}, \bibinfo {author} {\bibfnamefont {Y.}~\bibnamefont {Gu}}, \bibinfo {author} {\bibfnamefont {H.}~\bibnamefont {Yang}},\ and\ \bibinfo {author} {\bibfnamefont {C.}~\bibnamefont {Zhou}},\ }\href {https://doi.org/10.48550/arXiv.2103.11488} {\bibinfo {title} {The {{Discovery}} of {{Dynamics}} via {{Linear Multistep Methods}} and {{Deep Learning}}: {{Error Estimation}}}} (\bibinfo {year} {2022}),\ \Eprint {https://arxiv.org/abs/2103.11488} {arXiv:2103.11488 [math]} \BibitemShut {NoStop}%
\bibitem [{Note1()}]{Note1}%
  \BibitemOpen
  \bibinfo {note} {The equality for $M'$ holds for the generic case, disregarding possible cancellations or highly structured flow maps. Moreover, the flow map of a polynomial vector field is in general not polynomial, but analytic. For example, for $\protect \dot x = x^2$, the exact flow map is $\Phi _{\Delta t}(x)=\protect \frac {x}{1-\Delta t x} = x + \Delta t x^2 + \Delta t^2 x^3 + \protect \cdots $.}\BibitemShut {Stop}%
\bibitem [{\citenamefont {Langtangen}\ and\ \citenamefont {Linge}(2017)}]{langtangen_finite_2017}%
  \BibitemOpen
  \bibfield  {author} {\bibinfo {author} {\bibfnamefont {H.~P.}\ \bibnamefont {Langtangen}}\ and\ \bibinfo {author} {\bibfnamefont {S.}~\bibnamefont {Linge}},\ }\href {https://doi.org/10.1007/978-3-319-55456-3} {\emph {\bibinfo {title} {Finite {{Difference Computing}} with {{PDEs}}: {{A Modern Software Approach}}}}},\ \bibinfo {series} {Texts in {{Computational Science}} and {{Engineering}}}, Vol.~\bibinfo {volume} {16}\ (\bibinfo  {publisher} {Springer International Publishing},\ \bibinfo {address} {Cham},\ \bibinfo {year} {2017})\BibitemShut {NoStop}%
\bibitem [{\citenamefont {Lange}\ \emph {et~al.}(2020)\citenamefont {Lange}, \citenamefont {Brunton},\ and\ \citenamefont {Kutz}}]{lange_fourier_2020}%
  \BibitemOpen
  \bibfield  {author} {\bibinfo {author} {\bibfnamefont {H.}~\bibnamefont {Lange}}, \bibinfo {author} {\bibfnamefont {S.~L.}\ \bibnamefont {Brunton}},\ and\ \bibinfo {author} {\bibfnamefont {N.}~\bibnamefont {Kutz}},\ }\href {https://doi.org/10.48550/arXiv.2004.00574} {\bibinfo {title} {From {{Fourier}} to {{Koopman}}: {{Spectral Methods}} for {{Long-term Time Series Prediction}}}} (\bibinfo {year} {2020}),\ \Eprint {https://arxiv.org/abs/2004.00574} {arXiv:2004.00574 [cs]} \BibitemShut {NoStop}%
\bibitem [{Note2()}]{Note2}%
  \BibitemOpen
  \bibinfo {note} {If the data-generating process obeys recurrence relations, other forms of linear predictors may yield lower training errors. For example, for $x(t)=\sin (\omega t)$, a vanishing training error is achieved for $\protect \tilde x(t+\Delta t) = 2\cos (\omega \Delta t) x(t)-x(t-\Delta t)$.}\BibitemShut {Stop}%
\bibitem [{\citenamefont {Sch{\"o}tz}\ and\ \citenamefont {Boers}(2025)}]{schotz_machineprecision_2025}%
  \BibitemOpen
  \bibfield  {author} {\bibinfo {author} {\bibfnamefont {C.}~\bibnamefont {Sch{\"o}tz}}\ and\ \bibinfo {author} {\bibfnamefont {N.}~\bibnamefont {Boers}},\ }\href {https://doi.org/10.48550/arXiv.2507.09652} {\bibinfo {title} {Machine-{{Precision Prediction}} of {{Low-Dimensional Chaotic Systems}}}} (\bibinfo {year} {2025}),\ \Eprint {https://arxiv.org/abs/2507.09652} {arXiv:2507.09652 [nlin]} \BibitemShut {NoStop}%
\bibitem [{\citenamefont {Sprott}(2010)}]{sprott_elegant_2010}%
  \BibitemOpen
  \bibfield  {author} {\bibinfo {author} {\bibfnamefont {J.~C.}\ \bibnamefont {Sprott}},\ }\href@noop {} {\emph {\bibinfo {title} {Elegant {{Chaos}}: {{Algebraically Simple Chaotic Flows}}}}}\ (\bibinfo  {publisher} {World Scientific},\ \bibinfo {year} {2010})\BibitemShut {NoStop}%
\bibitem [{\citenamefont {Ma}\ \emph {et~al.}(2023)\citenamefont {Ma}, \citenamefont {Prosperino}, \citenamefont {Haluszczynski},\ and\ \citenamefont {R{\"a}th}}]{ma_efficient_2023}%
  \BibitemOpen
  \bibfield  {author} {\bibinfo {author} {\bibfnamefont {H.}~\bibnamefont {Ma}}, \bibinfo {author} {\bibfnamefont {D.}~\bibnamefont {Prosperino}}, \bibinfo {author} {\bibfnamefont {A.}~\bibnamefont {Haluszczynski}},\ and\ \bibinfo {author} {\bibfnamefont {C.}~\bibnamefont {R{\"a}th}},\ }\bibfield  {title} {\bibinfo {title} {Efficient forecasting of chaotic systems with block-diagonal and binary reservoir computing},\ }\href {https://doi.org/10.1063/5.0151290} {\bibfield  {journal} {\bibinfo  {journal} {Chaos}\ }\textbf {\bibinfo {volume} {33}},\ \bibinfo {pages} {063130} (\bibinfo {year} {2023})}\BibitemShut {NoStop}%
\bibitem [{\citenamefont {Vaidyanathan}\ and\ \citenamefont {Azar}(2016)}]{vaidyanathan_adaptive_2016}%
  \BibitemOpen
  \bibfield  {author} {\bibinfo {author} {\bibfnamefont {S.}~\bibnamefont {Vaidyanathan}}\ and\ \bibinfo {author} {\bibfnamefont {A.~T.}\ \bibnamefont {Azar}},\ }\bibfield  {title} {\bibinfo {title} {Adaptive {{Control}} and {{Synchronization}} of~{{Halvorsen Circulant Chaotic Systems}}},\ }in\ \href {https://doi.org/10.1007/978-3-319-30340-6_10} {\emph {\bibinfo {booktitle} {Advances in {{Chaos Theory}} and {{Intelligent Control}}}}},\ \bibinfo {editor} {edited by\ \bibinfo {editor} {\bibfnamefont {A.~T.}\ \bibnamefont {Azar}}\ and\ \bibinfo {editor} {\bibfnamefont {S.}~\bibnamefont {Vaidyanathan}}}\ (\bibinfo  {publisher} {Springer International Publishing},\ \bibinfo {address} {Cham},\ \bibinfo {year} {2016})\ pp.\ \bibinfo {pages} {225--247}\BibitemShut {NoStop}%
\bibitem [{\citenamefont {Ratas}\ and\ \citenamefont {Pyragas}(2024)}]{ratas_application_2024}%
  \BibitemOpen
  \bibfield  {author} {\bibinfo {author} {\bibfnamefont {I.}~\bibnamefont {Ratas}}\ and\ \bibinfo {author} {\bibfnamefont {K.}~\bibnamefont {Pyragas}},\ }\bibfield  {title} {\bibinfo {title} {Application of next-generation reservoir computing for predicting chaotic systems from partial observations},\ }\href {https://doi.org/10.1103/PhysRevE.109.064215} {\bibfield  {journal} {\bibinfo  {journal} {Phys. Rev. E}\ }\textbf {\bibinfo {volume} {109}},\ \bibinfo {pages} {064215} (\bibinfo {year} {2024})}\BibitemShut {NoStop}%
\bibitem [{\citenamefont {Sprott}(1997)}]{sprott_simple_1997}%
  \BibitemOpen
  \bibfield  {author} {\bibinfo {author} {\bibfnamefont {J.~C.}\ \bibnamefont {Sprott}},\ }\bibfield  {title} {\bibinfo {title} {Some simple chaotic jerk functions},\ }\href {https://doi.org/10.1119/1.18585} {\bibfield  {journal} {\bibinfo  {journal} {American Journal of Physics}\ }\textbf {\bibinfo {volume} {65}},\ \bibinfo {pages} {537} (\bibinfo {year} {1997})}\BibitemShut {NoStop}%
\bibitem [{\citenamefont {Vaidyanathan}\ \emph {et~al.}(2019)\citenamefont {Vaidyanathan}, \citenamefont {Sambas}, \citenamefont {Zhang}, \citenamefont {{Mujiarto}}, \citenamefont {Mamat},\ and\ \citenamefont {{Subiyanto}}}]{vaidyanathan_chaotic_2019}%
  \BibitemOpen
  \bibfield  {author} {\bibinfo {author} {\bibfnamefont {S.}~\bibnamefont {Vaidyanathan}}, \bibinfo {author} {\bibfnamefont {A.}~\bibnamefont {Sambas}}, \bibinfo {author} {\bibfnamefont {S.}~\bibnamefont {Zhang}}, \bibinfo {author} {\bibnamefont {{Mujiarto}}}, \bibinfo {author} {\bibfnamefont {M.}~\bibnamefont {Mamat}},\ and\ \bibinfo {author} {\bibnamefont {{Subiyanto}}},\ }\bibfield  {title} {\bibinfo {title} {A {{Chaotic Jerk System}} with {{Three Cubic Nonlinearities}}, {{Dynamical Analysis}}, {{Adaptive Chaos Synchronization}} and {{Circuit Simulation}}},\ }\href {https://doi.org/10.1088/1742-6596/1179/1/012083} {\bibfield  {journal} {\bibinfo  {journal} {Journal of Physics: Conference Series}\ }\textbf {\bibinfo {volume} {1179}},\ \bibinfo {pages} {012083} (\bibinfo {year} {2019})}\BibitemShut {NoStop}%
\bibitem [{\citenamefont {Sprott}(2003)}]{sprott_chaos_2003}%
  \BibitemOpen
  \bibfield  {author} {\bibinfo {author} {\bibfnamefont {J.~C.}\ \bibnamefont {Sprott}},\ }\href {https://doi.org/10.1093/oso/9780198508397.001.0001} {\emph {\bibinfo {title} {Chaos and {{Time-Series Analysis}}}}}\ (\bibinfo  {publisher} {Oxford University Press},\ \bibinfo {year} {2003})\BibitemShut {NoStop}%
\bibitem [{Note3()}]{Note3}%
  \BibitemOpen
  \bibinfo {note} {We used the \protect \texttt {solve\protect \_ivp} function from the Python \protect \texttt {scipy} library with accuracy $10^{-13}$.}\BibitemShut {Stop}%
\bibitem [{\citenamefont {Schuld}\ and\ \citenamefont {Petruccione}(2021)}]{schuld_machine_2021}%
  \BibitemOpen
  \bibfield  {author} {\bibinfo {author} {\bibfnamefont {M.}~\bibnamefont {Schuld}}\ and\ \bibinfo {author} {\bibfnamefont {F.}~\bibnamefont {Petruccione}},\ }\href@noop {} {\emph {\bibinfo {title} {{Machine Learning with Quantum Computers}}}}\ (\bibinfo  {publisher} {Springer},\ \bibinfo {address} {Cham},\ \bibinfo {year} {2021})\BibitemShut {NoStop}%
\bibitem [{\citenamefont {Fujii}\ and\ \citenamefont {Nakajima}(2017)}]{fujii_harnessing_2017}%
  \BibitemOpen
  \bibfield  {author} {\bibinfo {author} {\bibfnamefont {K.}~\bibnamefont {Fujii}}\ and\ \bibinfo {author} {\bibfnamefont {K.}~\bibnamefont {Nakajima}},\ }\bibfield  {title} {\bibinfo {title} {Harnessing {{Disordered-Ensemble Quantum Dynamics}} for {{Machine Learning}}},\ }\href {https://doi.org/10.1103/PhysRevApplied.8.024030} {\bibfield  {journal} {\bibinfo  {journal} {Phys. Rev. Applied}\ }\textbf {\bibinfo {volume} {8}},\ \bibinfo {pages} {024030} (\bibinfo {year} {2017})}\BibitemShut {NoStop}%
\bibitem [{\citenamefont {Gross}\ and\ \citenamefont {Rieser}(2026)}]{gross_theory_2026}%
  \BibitemOpen
  \bibfield  {author} {\bibinfo {author} {\bibfnamefont {M.}~\bibnamefont {Gross}}\ and\ \bibinfo {author} {\bibfnamefont {H.-M.}\ \bibnamefont {Rieser}},\ }\href {https://doi.org/10.48550/arXiv.2602.18377} {\bibinfo {title} {Theory and interpretability of {{Quantum Extreme Learning Machines}}: A {{Pauli-transfer}} matrix approach}} (\bibinfo {year} {2026}),\ \Eprint {https://arxiv.org/abs/2602.18377} {arXiv:2602.18377 [quant-ph]} \BibitemShut {NoStop}%
\bibitem [{\citenamefont {Schuld}\ \emph {et~al.}(2021)\citenamefont {Schuld}, \citenamefont {Sweke},\ and\ \citenamefont {Meyer}}]{schuld_effect_2021}%
  \BibitemOpen
  \bibfield  {author} {\bibinfo {author} {\bibfnamefont {M.}~\bibnamefont {Schuld}}, \bibinfo {author} {\bibfnamefont {R.}~\bibnamefont {Sweke}},\ and\ \bibinfo {author} {\bibfnamefont {J.~J.}\ \bibnamefont {Meyer}},\ }\bibfield  {title} {\bibinfo {title} {Effect of data encoding on the expressive power of variational quantum-machine-learning models},\ }\href {https://doi.org/10.1103/PhysRevA.103.032430} {\bibfield  {journal} {\bibinfo  {journal} {Phys. Rev. A}\ }\textbf {\bibinfo {volume} {103}},\ \bibinfo {pages} {032430} (\bibinfo {year} {2021})}\BibitemShut {NoStop}%
\bibitem [{\citenamefont {Stoica}\ and\ \citenamefont {Moses}(2005)}]{stoica_spectral_2005}%
  \BibitemOpen
  \bibfield  {author} {\bibinfo {author} {\bibfnamefont {P.}~\bibnamefont {Stoica}}\ and\ \bibinfo {author} {\bibfnamefont {R.}~\bibnamefont {Moses}},\ }\href@noop {} {\emph {\bibinfo {title} {{Spectral Analysis Of Signals}}}}\ (\bibinfo  {publisher} {Pearson},\ \bibinfo {address} {Upper Saddle River, NJ},\ \bibinfo {year} {2005})\BibitemShut {NoStop}%
\bibitem [{\citenamefont {Oppenheim}\ and\ \citenamefont {Schafer}(2009)}]{oppenheim_discretetime_2009}%
  \BibitemOpen
  \bibfield  {author} {\bibinfo {author} {\bibfnamefont {A.~V.}\ \bibnamefont {Oppenheim}}\ and\ \bibinfo {author} {\bibfnamefont {R.~W.}\ \bibnamefont {Schafer}},\ }\href@noop {} {\emph {\bibinfo {title} {{Discrete-Time Signal Processing}}}}\ (\bibinfo  {publisher} {Prentice Hall},\ \bibinfo {address} {Upper Saddle River Munich},\ \bibinfo {year} {2009})\BibitemShut {NoStop}%
\bibitem [{\citenamefont {Kantz}\ and\ \citenamefont {Schreiber}(2004)}]{kantz_nonlinear_2004}%
  \BibitemOpen
  \bibfield  {author} {\bibinfo {author} {\bibfnamefont {H.}~\bibnamefont {Kantz}}\ and\ \bibinfo {author} {\bibfnamefont {T.}~\bibnamefont {Schreiber}},\ }\href@noop {} {\emph {\bibinfo {title} {{Nonlinear Time Series Analysis}}}}\ (\bibinfo  {publisher} {Cambridge University Press},\ \bibinfo {address} {Cambridge, UK ; New York},\ \bibinfo {year} {2004})\BibitemShut {NoStop}%
\bibitem [{\citenamefont {Abarbanel}\ \emph {et~al.}(1993)\citenamefont {Abarbanel}, \citenamefont {Brown}, \citenamefont {Sidorowich},\ and\ \citenamefont {Tsimring}}]{abarbanel_analysis_1993}%
  \BibitemOpen
  \bibfield  {author} {\bibinfo {author} {\bibfnamefont {H.~D.~I.}\ \bibnamefont {Abarbanel}}, \bibinfo {author} {\bibfnamefont {R.}~\bibnamefont {Brown}}, \bibinfo {author} {\bibfnamefont {J.~J.}\ \bibnamefont {Sidorowich}},\ and\ \bibinfo {author} {\bibfnamefont {L.~S.}\ \bibnamefont {Tsimring}},\ }\bibfield  {title} {\bibinfo {title} {The analysis of observed chaotic data in physical systems},\ }\href {https://doi.org/10.1103/RevModPhys.65.1331} {\bibfield  {journal} {\bibinfo  {journal} {Reviews of Modern Physics}\ }\textbf {\bibinfo {volume} {65}},\ \bibinfo {pages} {1331} (\bibinfo {year} {1993})}\BibitemShut {NoStop}%
\bibitem [{\citenamefont {Kennel}\ \emph {et~al.}(1994)\citenamefont {Kennel}, \citenamefont {Abarbanel},\ and\ \citenamefont {Sidorowich}}]{kennel_prediction_1994}%
  \BibitemOpen
  \bibfield  {author} {\bibinfo {author} {\bibfnamefont {M.~B.}\ \bibnamefont {Kennel}}, \bibinfo {author} {\bibfnamefont {H.~D.~I.}\ \bibnamefont {Abarbanel}},\ and\ \bibinfo {author} {\bibfnamefont {J.~J.~S.}\ \bibnamefont {Sidorowich}},\ }\href {https://doi.org/10.48550/arXiv.chao-dyn/9403001} {\bibinfo {title} {Prediction {{Errors}} and {{Local Lyapunov Exponents}}}} (\bibinfo {year} {1994}),\ \Eprint {https://arxiv.org/abs/chao-dyn/9403001} {arXiv:chao-dyn/9403001} \BibitemShut {NoStop}%
\end{thebibliography}%

\end{document}